\documentclass[11pt,review]{paper}
\usepackage{ae,aecompl}
\usepackage[T1]{fontenc}
\usepackage[latin9]{inputenc}
\usepackage{geometry}
\usepackage{float}
\usepackage{textcomp}
\usepackage{amsbsy}
\usepackage{graphicx}
\usepackage[numbers]{natbib}
\usepackage[unicode=true,
 bookmarks=false,
 breaklinks=false,pdfborder={0 0 1},backref=section,colorlinks=false]
 {hyperref}

\makeatletter

\providecommand{\tabularnewline}{\\}
\floatstyle{ruled}
\newfloat{algorithm}{tbp}{loa}
\providecommand{\algorithmname}{Algorithm}
\floatname{algorithm}{\protect\algorithmname}

\newcommand{\lyxaddress}[1]{
\par {\raggedright #1
\vspace{1.4em}
\noindent\par}
}

\usepackage{url}

\makeatother

\begin{document}

\title{A Category Space Approach to Supervised Dimensionality Reduction}

\author{\textonesuperior Anthony O. Smith and \texttwosuperior Anand Rangarajan}
\maketitle

\lyxaddress{\textonesuperior Dept. of Electrical and Computer Engineering, Florida
Institute of Technology, 150 W. University Blvd., Melbourne, FL 32901,
USA}

\lyxaddress{\texttwosuperior Dept. of Computer and Information Science and Engineering,
University of Florida, P. O. Box 116120, Gainesville, FL, 32611-6120,
USA}
\begin{abstract}
Supervised dimensionality reduction has emerged as an important theme
in the last decade. Despite the plethora of models and formulations,
there is a lack of a simple model which aims to project the set of
patterns into a space defined by the classes (or categories). To this
end, we set up a model in which each class is represented as a 1D
subspace of the vector space formed by the features. Assuming the
set of classes does not exceed the cardinality of the features, the
model results in multi-class supervised learning in which the features
of each class are projected into the class subspace. Class discrimination
is automatically guaranteed via the imposition of orthogonality of
the 1D class sub-spaces. The resulting optimization problem\textemdash formulated
as the minimization of a sum of quadratic functions on a Stiefel manifold\textemdash while
being non-convex (due to the constraints), nevertheless has a structure
for which we can identify when we have reached a global minimum. After
formulating a version with standard inner products, we extend the
formulation to reproducing kernel Hilbert spaces in a straightforward
manner. The optimization approach also extends in a similar fashion
to the kernel version. Results and comparisons with the multi-class
Fisher linear (and kernel) discriminants and principal component analysis
(linear and kernel) showcase the relative merits of this approach
to dimensionality reduction.
\end{abstract}

\section{Introduction\label{Introduction} }

Dimensionality reduction and supervised learning have long been active
tropes in machine learning. For example, principal component analysis
(PCA) and the support vector machine (SVM) are standard bearers for
dimensionality reduction and supervised learning respectively. Even
now, machine learning researchers are accustomed to performing PCA
when seeking a simple dimensionality reduction technique despite the
fact that it is an unsupervised learning approach. In the past decade,
there has been considerable interest to include supervision (expert
label information) into dimensionality reduction techniques. Beginning
with the well known EigenFaces versus FisherFaces debate \cite{BelhumeurHespanhaKriegman1997},
there has been considerable activity centered around using Fisher
linear discriminants (FLD) and other supervised learning approaches
in dimensionality reduction. Since the Fisher linear discriminant
has a multi-class extension, it is natural to begin there. However,
it is also natural to ask the question if this is the only possible
approach. In this work, we design a category space approach with the
fundamental goal of using multi-class information to aid in dimensionality
reduction. The motivation for the approach and the main thrust of
this work are our focus, next.

The venerable Fisher discriminant is a supervised dimensionality reduction
technique, wherein, a maximally discriminative one dimensional subspace
is estimated from the data. The criterion used for discrimination
is the ratio between the squared distance of the projected class means
and a weighted sum of the projected variances. This criterion has
a closed form solution yielding the best 1D subspace. The Fisher discriminant
also has an extension to the multi-class case. Here the criterion
used is more complex and highly unusual: it is the ratio between a
squared distance between each class projected mean and the total projected
mean and the sum of the projected variances. This too results in a
closed form solution but with the subspace dimension cardinality being
one less than the number of classes.

The above description of the multi-class FLD sets the stage for our
approach. We begin with the assumption that the set of categories
(classes) is a subspace of the original feature space (similar to
FLD). However, we add the restriction that the category bases are
mutually orthogonal with the origin of the vector space belonging
to no category. Given this restriction, the criterion for multi-class
category space dimensionality reduction is quite straightforward:
we simply maximize the square of the inner product between each pattern
and its own category axis with the aim of discovering the category
space via this process. (Setting the origin is a highly technical
issue and therefore not described here.) The result is a sum of quadratic
objective functions on a Stiefel manifold\textemdash the category
space of orthonormal basis vectors. This is a very interesting objective
function which has coincidentally received quite a bit of treatment
recently \cite{Rapcsak2002,BollaMichaletzkyTusnadyEtAl1998}. Furthermore,
there is no need to restrict ourselves to sums of quadratic objective
functions provided we are willing to forego useful analysis of this
base case. The unusual aspect of the objective function comprising
sums of quadratic objective functions is that we can formulate a criterion
which guarantees that we have reached a global minimum if the achieved
solution satisfies it. Unfortunately, there is no algorithm at the
present time that can \emph{a priori} guarantee satisfaction of this
criterion and hence we can only check on a case by case basis. Despite
this, our experimental results show that we get efficient solutions,
competitive with those obtained from other dimensionality reduction
algorithms. Extensive comparisons are conducted against principal
component analysis (PCA) and multi-class Fisher using support vector
machine (SVM) classifiers on the reduced set of features. 

It should be clear that the contribution of this paper is to a very
old problem in pattern recognition. While numerous alternatives exist
to the FLD (such as canonical correlation analysis \cite{Hardoon2004})
and while there are many nonlinear unsupervised dimensionality reduction
techniques (such as local linear embedding \cite{Roweis2000}, ISOMAP
\cite{Tenenbaum2000} and Laplacian Eigenmaps \cite{Belkin2003}),
we have not encountered a simple dimensionality reduction technique
which is based on projecting the data into a space spanned by the
categories. Obviously, numerous extensions and more abstract formulations
of the base case in this paper can be considered, but to reiterate,
we have not seen any previous work perform supervised dimensionality
reduction in the manner suggested here.

\section{Related Work\label{Related-Work}}

Traditional dimensionality reduction techniques like principal component
analysis (PCA) \cite{Jolliffe1986}, and supervised algorithms such
as Fisher linear discriminant analysis \cite{fisher1936use} seek
to retain significant features while removing insignificant, redundant,
or noisy features. These algorithms are frequently utilized as preprocessing
steps before the application of a classification algorithm and have
been been successful in solving many real-world problems. A limitation
in the vast majority of methods is that there is no specific connection
between the dimensionality reduction technique and the supervised
learning-driven classifier. Dimensionality reduction techniques such
as canonical correlation analysis (CCA) \cite{Hotelling1933}, and
partial least squares (PLS) \cite{Arenas-GarciaPetersenHansen2007}
on the one hand and classification algorithms such as support vector
machines (SVM) \cite{Vapnik1998} on the other seek to optimize different
criteria. In contrast, in this paper, we analyze dimensionality reduction
from the perspective of multi-class classification. The use of a category
vector space (with dimension equal to class cardinality) is an integral
aspect of this approach.

In supervised learning, it is customary for classification methodologies
to regard classes as nominal labels without having any internal structure.
This remains true regardless of whether a discriminant or classifier
is sought. Discriminants are designed by attempting to separate patterns
into oppositional classes \cite{Bishop1996,DudaHart1973,hastie1996discriminant}.
When generalization to a multi-class classifier is required, many
oppositional discriminants are combined with the final classifier
being a winner-take-all (or voting-based) decision w.r.t. the set
of nominal labels. Convex objective functions based on misclassification
error minimization (or approximation) are not that different either.
Least-squares or logistic regression methods set up convex objective
functions with nominal labels converted to binary outputs \cite{Ye2007,bishop2006pattern}.
When extensions to multi-class are sought, the binary labels are extended
to a one of $K$ encoding with $K$ being the number of classes. Support
vector machines (SVM's) were inherently designed for two class discrimination
and all formulations of multi-class SVM's extend this oppositional
framework using one-versus-one or one-versus-all schemes. Below, we
begin by describing the different approaches to the multi-class problem.
This is not meant to be exhaustive, but provides an overview of some
of the popular methods and approaches that have been researched in
classification and dimensionality reduction. Folley and Sammon \cite{Sammon1970},
\cite{FoleySammon1975} studied the two class problem and feature
selection and focused on criteria with greatest potential to discriminate.
The goal of feature selection is to find a set of features with the
best discrimination properties. To identify the best feature vectors
they chose the generalized Fisher optimality criterion proposed by
\cite{AndersonBahadur1962}. The selected directions maximize the
Fisher criterion which has attractive properties of discrimination.
Principal components analysis (PCA) permits the reduction of dimensions
of high dimensional data without losing significant information \cite{Hotelling1933,Jolliffe1986,scholkopf1999advances}.
Principal components are a way of identifying patterns or significant
features without taking into account discriminative considerations
\cite{rao1964use}. Supervised PCA (SPCA), derived from PCA is a method
for obtaining useful sub-spaces when the labels are taken into account.
This technique was first described in \cite{bair2004semi} under the
title ``supervised clustering.'' The idea behind SPCA is to perform
selective dimensionality reduction using carefully chosen subsets
of labeled samples. This is used to build a prediction model \cite{bair2012prediction}.
While we have addressed the most popular techniques in dimensionality
reduction and multi-class classification, this is not an exhaustive
study of the literature. Our focus so far is primarily on discriminative
dimensionality reduction methods that assist in better multi-class
classification performance. The closest we have seen in relation to
our work on category spaces is the work in \cite{widdows2004geometry}
and \cite{widdows2003orthogonal}. Here, they mention the importance
and usefulness of modeling categories as vector spaces for document
retrieval and explain how unrelated items should have an orthogonal
relationship. This is to say that they should have no features in
common. The structured SVM in \cite{tsochantaridis2004support} is
another effort at going beyond nominal classes. Here, classes are
allowed to have internal structure in the form of strings, trees etc.
However, an explicit modeling of classes as vector spaces is not carried
out. 

From the above, the modest goal of the present work should be clear.
We seek to project the input feature vectors to a category space\textemdash a
subspace formed by category basis vectors. The multi-class FLD falls
short of this goal since the number of projected dimensions is one
less than the number of classes. The multi-class (and more recently
multi-label) SVM \cite{ji2009linear} literature is fragmented due
to lack of agreement regarding the core issue of multi-class discrimination.
The varieties of supervised PCA do not begin by clearly formulating
a criterion for category space projection. Variants such as CCA \cite{johnson2002applied,sun2011canonical},
PLS \cite{sun2013multi} and structured SVM's \cite{tsochantaridis2004support}
while attempting to add structure to the categories do not go as far
as the present work in attempting to fit a category subspace. Kernel
variants of the above also do not touch the basic issue addressed
in the present work. Nonlinear (and manifold learning-based) dimensionality
reduction techniques \cite{Roweis2000,Tenenbaum2000,Belkin2003} are
unsupervised and therefore do not qualify.

\section{Dimensionality Reduction using a Category Space Formulation\label{Category-Vector-Space}}

\subsection{Maximizing the square of the inner product}

The principal goal of this paper is a new form of supervised dimensionality
reduction. Specifically, when we seek to marry principal component
analysis with supervised learning, by far the simplest synthesis is
category space dimensionality reduction with orthogonal class vectors.
Assume the existence of a feature space with each feature vector $x_{i}\in\mathbf{R}^{D}$.
Our goal is to perform supervised dimensionality reduction by reducing
the number of feature dimensions from $D$ to $K$ where $K\leq D$.
Here $K$ is the number of classes and the first simplifying assumption
made in this work is that we will represent the category space using
$K$ \emph{orthonormal} basis vectors $\left\{ w_{k}\right\} $ together
with an \emph{origin} $x_{0}\in\mathbf{R}^{D}$. The second assumption
we make is that each feature vector $x_{i}$ should have a large magnitude
inner product with its assigned class. From the orthonormality constraint
above, this automatically implies a small magnitude inner product
with all other weight vectors. A \emph{candidate objective function}
and constraints following the above considerations is

\begin{equation}
E(W)=-\frac{1}{2}\sum_{k=1}^{K}\sum_{i_{k}\in C_{k}}\left[w_{k}^{T}\left(x_{i_{k}}-x_{0}\right)\right]^{2}\label{eq:innerprodmax}
\end{equation}
and 

\begin{equation}
w_{k}^{T}w_{l}=\left\{ \begin{array}{cc}
1, & k=l\\
0, & k\neq l
\end{array}\right.\label{eq:wkwlinner}
\end{equation}
respectively. In (\ref{eq:innerprodmax}), $W=\left[w_{1},w_{2},\ldots,w_{K}\right]$.
Note that we have referred to this as a candidate objective function
for two reasons. First, the origin $x_{0}$ is still unspecified and
we cannot obviously minimize (\ref{eq:innerprodmax}) w.r.t. $x_{0}$
as the minimum value is not bounded from below. Second, it is not
clear why we cannot use the absolute value or other symmetric functions
of the inner product. Both these issues are addressed later in this
work. At present, we resolve the origin issue by setting $x_{0}$
to the centroid of all the feature vectors (with this choice getting
a principled justification below).

The objective function in (\ref{eq:innerprodmax}) is the negative
of a quadratic function. Since the function $-x^{2}$ is concave,
it admits a Legendre transform-based majorization \cite{yuille2003concave}
using the tangent of the function. That is, we propose to replace
objective functions of the form $-\frac{1}{2}x^{2}$ with $\min_{y}-xy+\frac{1}{2}y^{2}$
which can quickly checked to be valid for an unconstrained auxiliary
variable $y$. Note that this transformation yields a linear objective
function w.r.t. $x$ which is to be expected from the geometric interpretation
of a tangent. 

Consider the following Legendre transformation of the objective function
in (\ref{eq:innerprodmax}). The new objective function is 

\begin{equation}
E_{\mathrm{quad}}(W,Z)=\sum_{k=1}^{K}\sum_{i_{k}\in C_{k}}\left[z_{ki_{k}}\left(-w_{k}^{T}x_{i_{k}}+w_{k}^{T}x_{0}\right)+\frac{1}{2}z_{ki_{k}}^{2}\right]\label{eq:EWZ}
\end{equation}
where $Z=\left\{ z_{ki_{k}}|k\in\left\{ 1,\ldots,K\right\} ,i_{k}\in\left\{ 1,\ldots,|C_{k}|\right\} \right\} $.
Now consider this to be an objective function over $x_{0}$ as well.
In order to avoid minima at negative infinity, we require additional
constraints. One such constraint (and perhaps not the only one) is
of the form $\sum_{i_{k}\in C_{k}}z_{ki_{k}}=0,\forall k$. When this
constraint is imposed, we obtain a new objective function 

\begin{equation}
E_{\mathrm{quad}}(W,Z)=\sum_{k=1}^{K}\sum_{i_{k}\in C_{k}}\left[-z_{ki_{k}}w_{k}^{T}x_{i_{k}}+\frac{1}{2}z_{ki_{k}}^{2}\right]\label{eq:EWZ2}
\end{equation}
to be minimized subject to the constraints 

\begin{equation}
\sum_{i_{k}\in C_{k}}z_{ki_{k}}=0,\forall k\label{eq:sumzequalszero}
\end{equation}
in addition to the orthonormal constraints in (\ref{eq:wkwlinner}).
This objective function yields a $Z$ which removes the class-specific
centroid of $C_{k}$ for all classes.

\subsection{Maximizing the absolute value of the inner product}

We have justified our choice of centroid removal mentioned above indirectly
obtained via constraints imposed on Legendre transform auxiliary variables.
The above objective function can be suitably modified when we use
different forms (absolute inner product etc.). To see this, consider
the following objective function which minimizes the negative of the
magnitude of the inner product:
\begin{equation}
E(W)=-\sum_{k=1}^{K}\sum_{i_{k}\in C_{k}}|w_{k}^{T}(x_{i_{k}}-x_{0})|.\label{eq:absvalobj}
\end{equation}
Since $-|x|$ is also a concave function, it too can be majorized.
Consider first replacing the non-differentiable objective function
$-|x|$ with $-\sqrt{x^{2}+\epsilon}$ (also concave) where $\epsilon$
can be chosen to be a suitably small value. Now consider replacing
$-\sqrt{x^{2}+\epsilon}$ with $\min_{y}-xy-\epsilon\sqrt{1-y^{2}}$
which can again quickly checked to be valid for a constrained auxiliary
variable $y\in\left[-1,1\right]$. The constraint is somewhat less
relevant since the minimum w.r.t. $y$ occurs at $y=\frac{x}{\sqrt{x^{2}+\epsilon^{2}}}$
which lies within the constraint interval. Note that this transformation
also yields a linear objective function w.r.t. $x$. As before, we
introduce a new objective function
\begin{equation}
E_{\mathrm{abs}}(W,Z)=\sum_{k=1}^{K}\sum_{i_{k}\in C_{k}}\left[-z_{ki_{k}}w_{k}^{T}x_{i_{k}}-\epsilon\sqrt{1-z_{ki_{k}}^{2}}\right]\label{eq:EWZabs}
\end{equation}
to be minimized subject to the constraints $\sum_{i_{k}\in C_{k}}z_{ki_{k}}=0,\,\forall k$
and $z_{ki_{k}}\in\left[-1,1\right]$ which are the same as in (\ref{eq:sumzequalszero})
in addition to the orthonormal constraints in (\ref{eq:wkwlinner}). 

\subsection{Extension to RKHS kernels}

The generalization to RKHS kernels is surprisingly straightforward.
First, we follow standard kernel PCA and write the weight vector in
terms of the RKHS projected patterns $\phi(x_{l})$ to get 
\begin{equation}
w_{k}=\sum_{i=1}^{N}\alpha_{ki}\phi(x_{i}).\label{eq:wkRKHS}
\end{equation}
Note that the expansion of the weight vector is over all patterns
rather than just the class-specific ones. This assumes that the weight
vector for each class lives in the subspace (potentially infinite
dimensional) spanned by the RKHS projected patterns\textemdash the
same assumption as in standard kernel PCA. The orthogonality constraint
between weight vectors becomes 
\begin{equation}
\begin{array}{ccc}
\langle w_{k},w_{l}\rangle & = & \langle\sum_{i=1}^{N}\alpha_{ki}\phi(x_{i}),\sum_{i=1}^{N}\alpha_{li}\phi(x_{i})\rangle\\
 & = & \sum_{i=1}^{N}\sum_{j=1}^{N}\alpha_{ki}\alpha_{kj}\langle\phi(x_{i}),\phi(x_{j})\rangle\\
 & = & \sum_{i=1}^{N}\sum_{j=1}^{N}\alpha_{ki}\alpha_{kj}K(x_{i},x_{j})
\end{array}\label{eq:wkwlRKHS}
\end{equation}
which is equal to one if $k=l$ and zero otherwise. In matrix form,
the orthonormality constraints become
\begin{equation}
AGA^{T}=I_{K}\label{eq:orthoRKHS}
\end{equation}
where $\left[A\right]_{kl}\equiv\alpha_{ki}$ and $\left[G\right]_{ij}=K(x_{i},x_{j})$
is the well-known Gram matrix of pairwise RKHS inner products between
the patterns. 

The corresponding squared inner product and absolute value of inner
product objective functions are 
\begin{equation}
E_{\mathrm{Kquad}}(A,Z)=\sum_{k=1}^{K}\sum_{i_{k}\in C_{k}}\left[-\sum_{j=1}^{N}z_{ki_{k}}\alpha_{kj}K(x_{j},x_{i_{k}})+\frac{1}{2}z_{ki_{k}}^{2}\right]\label{eq:Kquadobj}
\end{equation}
and
\begin{equation}
E_{\mathrm{Kabs}}(A,Z)=\sum_{k=1}^{K}\sum_{i_{k}\in C_{k}}\left[-\sum_{j=1}^{N}z_{ki_{k}}\alpha_{kj}K(x_{j},x_{i_{k}})-\epsilon\sqrt{1-z_{ki_{k}}^{2}}\right]\label{eq:Kabsobj}
\end{equation}
respectively. These have to be minimized w.r.t. the orthonormal constraints
in (\ref{eq:orthoRKHS}) and the origin constraints in (\ref{eq:sumzequalszero}).
Note that the objective functions are identical w.r.t. the matrix
$A$. The parameter $\epsilon$ can be set to a very small but positive
value.

\section{An algorithm for supervised dimensionality reduction}

We now return to the objective functions and constraints in (\ref{eq:EWZ2})
and (\ref{eq:EWZabs}) prior to tackling the corresponding kernel
versions in (\ref{eq:Kquadobj}) and (\ref{eq:Kabsobj}) respectively.
It turns out that the approach for minimizing the former can be readily
generalized to the latter with the former being easier to analyze.
Note that the objective functions in (\ref{eq:EWZ2}) and (\ref{eq:EWZabs})
are identical w.r.t. $W$. Consequently, we dispense with the optimization
problems w.r.t. $Z$ which are straightforward and focus on the optimization
problem w.r.t. $W$. 

\subsection{Weight matrix estimation with orthogonality constraints}

The objective function and constraints on $W$ can be written as 

\begin{equation}
E_{\mathrm{equiv}}(W)=-\sum_{k=1}^{K}\sum_{i_{k}\in C_{k}}z_{ki_{k}}w_{k}^{T}x_{i_{k}}\label{eq:Eequiv}
\end{equation}
and

\begin{equation}
w_{k}^{T}w_{l}=\left\{ \begin{array}{cc}
1, & k=l\\
0, & k\neq l
\end{array}.\right.\label{eq:wkwlconstraintsfinal}
\end{equation}
Note that the set $Z$ is not included in this objective function
despite its presence in the larger objective functions of (\ref{eq:EWZ2})
and (\ref{eq:EWZabs}). The orthonormal constraints can be expressed
using a Lagrange parameter matrix to obtain the following Lagrangian:

\begin{equation}
L(W,\Lambda)=-\sum_{k=1}^{K}\sum_{i_{k}\in C_{k}}z_{ki_{k}}w_{k}^{T}x_{i_{k}}+\mbox{trace}\left\{ \Lambda\left(W^{T}W-I_{K}\right)\right\} .\label{eq:Lagrangianortho}
\end{equation}
Setting the gradient of $L$ w.r.t. $W$ to zero, we obtain
\begin{equation}
\nabla_{W}L\left(W,\Lambda\right)=-Y+W\left(\Lambda+\Lambda^{T}\right)=0\label{eq:gradLagrangian}
\end{equation}
where the matrix $Y$ of size $D\times K$ is defined as 

\begin{equation}
Y\equiv\left[\sum_{i_{1}\in C_{1}}z_{1i_{1}}x_{i_{1}},\ldots,\sum_{i_{k}\in C_{k}}z_{ki_{k}}x_{i_{k}}\right]\label{eq:Ydef}
\end{equation}
Using the constraint $W^{T}W=I_{K}$ , we get 
\begin{equation}
\left(\Lambda+\Lambda^{T}\right)=W^{T}Y.\label{eq:lambdasol}
\end{equation}
Since $\left(\Lambda+\Lambda^{T}\right)$ is symmetric, this immediately
implies that $W^{T}Y$ is symmetric. From (\ref{eq:gradLagrangian}),
we also get 

\begin{equation}
\left(\Lambda+\Lambda^{T}\right)W^{T}W\left(\Lambda+\Lambda^{T}\right)=\left(\Lambda+\Lambda^{T}\right)^{2}=Y^{T}Y.\label{eq:lambdasol2}
\end{equation}
Expanding $Y$ using its singular value decomposition (SVD) as $Y=U\Sigma V^{T}$,
the above relations can be simplified to 

\begin{equation}
Y=U\Sigma V^{T}=UV^{T}(V\Sigma V^{T})=W\left(\Lambda+\Lambda^{T}\right)\label{eq:Wsoltmp}
\end{equation}
giving 

\begin{equation}
\left(\Lambda+\Lambda^{T}\right)=V\Sigma V^{T}\label{eq:lambdasolfinal}
\end{equation}
and 

\begin{equation}
W=UV^{T}.\label{eq:Wsolfinal}
\end{equation}
We have shown that the optimal solution for $W$ is the polar decomposition
of $Y$, namely $W=UV^{T}$. Since $Z$ has been held fixed during
the estimation of $W$, in the subsequent step we can hold $W$ fixed
and solve for $Z$ and repeat. We thereby obtain an alternating algorithm
which iterates between estimating $W$ and $Z$ until a convergence
criterion is met. 

\subsection{Estimation of the auxiliary variable $Z$}

The objective function and constraints on $Z$ depend on whether we
use objective functions based on the square or absolute value of the
inner product. We separately consider the two cases.

The inner product squared effective objective function 
\begin{equation}
E_{\mathrm{quadeff}}(Z)=\sum_{k=1}^{K}\sum_{i_{k}\in C_{k}}\left[-z_{ki_{k}}w_{k}^{T}x_{i_{k}}+\frac{1}{2}z_{ki_{k}}^{2}\right]\label{eq:Equadeff}
\end{equation}
is minimized w.r.t. $Z$ subject to the constraints $\sum_{i_{k}\in C_{k}}z_{ki_{k}}=0,\forall k$.
The straightforward solution obtained via standard minimization is 

\begin{equation}
\begin{array}{ccc}
z_{ki_{k}} & = & w_{k}^{T}x_{i_{k}}-\frac{1}{|C_{k}|}\sum_{i_{k}\in C_{k}}w_{k}^{T}x_{i_{k}}\\
 & = & w_{k}^{T}\left(x_{i_{k}}-\frac{1}{|C_{k}|}\sum_{i_{k}\in C_{k}}x_{i_{k}}\right).
\end{array}\label{eq:Zsolquad}
\end{equation}
The absolute value effective objective function 

\begin{equation}
E_{\mathrm{abseff}}(Z)=\sum_{k=1}^{K}\sum_{i_{k}\in C_{k}}\left[-z_{ki_{k}}w_{k}^{T}x_{i_{k}}-\epsilon\sqrt{1-z_{ki_{k}}^{2}}\right]\label{eq:Eabseff}
\end{equation}
is also minimized w.r.t. $Z$ subject to the constraints $\sum_{i_{k}\in C_{k}}z_{ki_{k}}=0,\forall k$.
A heuristic solution obtained (eschewing standard minimization) is 

\begin{equation}
z_{ki_{k}}=\frac{w_{k}^{T}x_{i_{k}}}{\sqrt{\left(w_{k}^{T}x_{i_{k}}\right)^{2}+\epsilon^{2}}}-\frac{1}{|C_{k}|}\sum_{i_{k}\in C_{k}}\frac{w_{k}^{T}x_{i_{k}}}{\sqrt{\left(w_{k}^{T}x_{i_{k}}\right)^{2}+\epsilon^{2}}}\label{eq:Zsolabs}
\end{equation}
which has to be checked to be valid. The heuristic solution acts as
an initial condition for constraint satisfaction (which can be efficiently
obtained via 1D line minimization). The first order Karush-Kuhn-Tucker
(KKT) conditions obtained from the Lagrangian

\begin{equation}
L_{\mathrm{abseff}}(Z,M)=\sum_{k=1}^{K}\sum_{i_{k}\in C_{k}}\left[-z_{ki_{k}}w_{k}^{T}x_{i_{k}}-\epsilon\sqrt{1-z_{ki_{k}}^{2}}\right]-\sum_{k=1}^{K}\mu_{k}\sum_{i_{k}\in C_{k}}z_{ki_{k}}\label{eq:Labseff}
\end{equation}
are 

\begin{equation}
-w_{k}^{T}x_{i_{k}}+\epsilon\frac{z_{ki_{k}}}{\sqrt{1-z_{ki_{k}}^{2}}}-\mu_{k}=0,\,\forall k\label{eq:LabszeffKKT}
\end{equation}
from which we obtain

\begin{equation}
z_{ki_{k}}=\frac{w_{k}^{T}x_{i_{k}}+\mu_{k}}{\sqrt{\left(w_{k}^{T}x_{i_{k}}+\mu_{k}\right)^{2}+\epsilon^{2}}}.\label{eq:zkiksol2}
\end{equation}
We see that the the constraint $z_{ki_{k}}\in[-1,1]$ is also satisfied.
For each category $C_{k}$, there exists a solution to the Lagrange
parameter $\mu_{k}$ such that $\sum_{i_{k}}z_{ki_{k}}=0$. This can
be obtained via any efficient 1D search procedure like golden section
\cite{Kiefer1953}.

\subsection{Extension to the kernel setting}

The objective function and constraints on the weight matrix $A$ in
the kernel setting are 

\begin{equation}
E_{\mathrm{Kequiv}}(A)=-\sum_{k=1}^{K}\sum_{i_{k}\in C_{k}}\sum_{j=1}^{N}z_{ki_{k}}\alpha_{kj}K(x_{j},x_{i_{k}})\label{eq:EKequivA}
\end{equation}
with the constraints 

\begin{equation}
AGA^{T}=I_{K}\label{eq:AGAeqIK}
\end{equation}
where $\left[A\right]_{ki}=\alpha_{ki}$ and $\left[G\right]_{ij}=K(x_{i},x_{j})$
is the $N\times N$ kernel Gram matrix. The constraints can be expressed
using a Lagrange parameter matrix to obtain the following Lagrangian:

\begin{eqnarray}
L_{\mathrm{ker}}(A,\Lambda) & = & -\sum_{k=1}^{K}\sum_{i_{k}\in C_{k}}\sum_{j=1}^{N}z_{ki_{k}}\alpha_{kj}K(x_{j},x_{i_{k}})\nonumber \\
 &  & +\mbox{trace}\left\{ \Lambda_{\mathrm{ker}}\left(AGA^{T}-I_{K}\right)\right\} .\label{eq:Lagrangian_ker}
\end{eqnarray}
Setting the gradient of $L_{\mathrm{ker}}$ w.r.t. $A$ to zero, we
obtain

\begin{equation}
-Y_{\mathrm{ker}}+(\Lambda_{\mathrm{ker}}+\Lambda_{\mathrm{ker}}^{T})AG=0\label{eq:Lagrangian_ker_grad}
\end{equation}
where the matrix $Y_{\mathrm{ker}}$ of size $K\times N$ is defined
as 

\begin{equation}
\left[Y_{\mathrm{ker}}\right]_{kj}\equiv\sum_{i_{k}\in C_{k}}z_{ki_{k}}K(x_{j},x_{i_{k}}).\label{eq:Ykerdef}
\end{equation}
Using the constraint $AGA^{T}=I_{K}$, we obtain 

\begin{equation}
(\Lambda_{\mathrm{ker}}+\Lambda_{\mathrm{ker}}^{T})AGA^{T}(\Lambda_{\mathrm{ker}}+\Lambda_{\mathrm{ker}}^{T})=(\Lambda_{\mathrm{ker}}+\Lambda_{\mathrm{ker}}^{T})^{2}=Y_{\mathrm{ker}}G^{-1}Y_{\mathrm{ker}}^{T}.\label{eq:lambdakersol}
\end{equation}
Expanding $Y_{\mathrm{ker}}G^{-\frac{1}{2}}$ using its singular value
decomposition as $Y_{\mathrm{ker}}G^{-\frac{1}{2}}=U_{\mathrm{ker}}S_{\mathrm{ker}}V_{\mathrm{ker}}^{T}$
, the above relations can be simplified to 

\begin{equation}
(\Lambda_{\mathrm{ker}}+\Lambda_{\mathrm{ker}}^{T})=U_{\mathrm{ker}}S_{\mathrm{ker}}U_{\mathrm{ker}}^{T}\label{eq:lambdakerfinal}
\end{equation}
and 

\begin{equation}
AG^{\frac{1}{2}}=U_{\mathrm{ker}}V_{\mathrm{ker}}^{T}\Rightarrow A=U_{\mathrm{ker}}V_{\mathrm{ker}}^{T}G^{-\frac{1}{2}}.\label{eq:Asolfinal}
\end{equation}
We have shown that the optimal solution for $A$ is related to the
polar decomposition of $Y_{\mathrm{ker}}G^{-\frac{1}{2}}$, namely
$A=U_{\mathrm{ker}}V_{\mathrm{ker}}^{T}G^{-\frac{1}{2}}$. Since $Z$
has been held fixed during the estimation of $A$, in the subsequent
step we can hold $A$ fixed and solve for $Z$ and repeat. We thereby
obtain an alternating algorithm which iterates between estimating
$A$ and $Z$ until a convergence criterion is met. This is analogous
to the non-kernel version above.

The solutions for $Z$ in this setting are very straightforward to
obtain. We eschew the derivation and merely state that
\noindent \begin{flushleft}
\begin{equation}
\begin{array}{ccc}
z_{ki_{k}} & = & \sum_{j=1}^{N}\alpha_{kj}K(x_{j},x_{i_{k}})-\frac{1}{|C_{k}|}\sum_{i_{k}\in C_{k}}\sum_{j=1}^{N}\alpha_{kj}K(x_{j},x_{i_{k}})\\
 & = & \sum_{j=1}^{N}\alpha_{kj}\left(K(x_{j},x_{i_{k}})-\frac{1}{|C_{k}|}\sum_{i_{k}\in C_{k}}K(x_{j},x_{i_{k}})\right)
\end{array}\label{eq:zksqrkersol}
\end{equation}
for the squared inner product kernel objective and 
\par\end{flushleft}

\begin{equation}
z_{ki_{k}}=\frac{\sum_{j=1}^{N}\alpha_{kj}K(x_{j},x_{i_{k}})}{\sqrt{\left(\sum_{j=1}^{N}\alpha_{kj}K(x_{j},x_{i_{k}})\right)^{2}+\epsilon^{2}}}-\frac{1}{|C_{k}|}\sum_{i_{k}\in C_{k}}\frac{\sum_{j=1}^{N}\alpha_{kj}K(x_{j},x_{i_{k}})}{\sqrt{\left(\sum_{j=1}^{N}\alpha_{kj}K(x_{j},x_{i_{k}})\right)^{2}+\epsilon^{2}}}\label{eq:zkabskersol}
\end{equation}
for the absolute valued kernel objective. This heuristic solution
acts as an initial condition for constraint satisfaction (which can
be efficiently be obtained via 1D line minimization). Following the
line of equations (\ref{eq:Labseff})-(\ref{eq:zkiksol2}) above,
the solution can be written as 

\begin{equation}
z_{ki_{k}}=\frac{\sum_{j=1}^{N}\alpha_{kj}K(x_{j},x_{i_{k}})+\mu_{k}}{\sqrt{\left(\sum_{j=1}^{N}\alpha_{kj}K(x_{j},x_{i_{k}})+\mu_{k}\right)^{2}+\epsilon^{2}}}.\label{eq:zkikKabseffsol2}
\end{equation}
For each category $C_{k}$, as before, there exists a solution to
the Lagrange parameter $\mu_{k}$ such that $\sum_{i_{k}}z_{ki_{k}}=0$.
Once again, this can be obtained via any efficient 1D search procedure
like golden section \cite{Kiefer1953}.

\subsection{Analysis}

\subsubsection{Euclidean setting\label{subsec:Euclidean-setting}}

The simplest objective function in the above sequence which has been
analyzed in the literature is the one based on the squared inner product.
Below, we summarize this work by closely following the treatment in
\cite{rapcsak2001minimization,Rapcsak2002}. First, in order to bring
our work in sync with the literature, we eliminate the auxiliary variable
$Z$ from the squared inner product objective function (treated as
a function of both $W$ and $Z$ here): 

\begin{equation}
E_{\mathrm{quadeff}}(W,Z)=\sum_{k=1}^{K}\sum_{i_{k}\in C_{k}}\left[-z_{ki_{k}}w_{k}^{T}x_{i_{k}}+\frac{1}{2}z_{ki_{k}}^{2}\right]\label{eq:Equadeff2}
\end{equation}
Setting $z_{ki_{k}}=w_{k}^{T}\left(x_{i_{k}}-\frac{1}{|C_{k}|}\sum_{i_{k}\in C_{k}}x_{i_{k}}\right)$
which is the optimum solution for $Z$, we get

\begin{equation}
E_{\mathrm{quad}}(W)=-\frac{1}{2}\sum_{k=1}^{K}w_{k}^{T}R_{k}w_{k}\equiv-\frac{1}{2}\sum_{k=1}^{K}\sum_{i_{k}\in C_{k}}\left[w_{k}^{T}\left(x_{i_{k}}-\frac{1}{|C_{k}|}\sum_{i\in C_{k}}x_{i}\right)\right]^{2}\label{eq:EquadW2}
\end{equation}
where $R_{k}$ is the class-specific covariance matrix:

\begin{equation}
R_{k}\equiv\sum_{i_{k}\in C_{k}}\left(x_{i_{k}}-\frac{1}{|C_{k}|}\sum_{i\in C_{k}}x_{i}\right)\left(x_{i_{k}}-\frac{1}{|C_{k}|}\sum_{i\in C_{k}}x_{i}\right)^{T}.\label{eq:Rkdef}
\end{equation}
We seek to minimize (\ref{eq:EquadW2}) w.r.t. $W$ under the orthonormality
constraints $W^{T}W=I_{K}$. 

A set of $K$ orthonormal vectors $\left\{ w_{k}\in\mathbf{R}^{D},\,k\in\left\{ 1,\ldots,K\right\} \right\} $
in a $D$-dimensional Euclidean space is a point on the well known
Stiefel manifold, denoted here by $M_{D,K}$ with $K\leq D$. The
problem in (\ref{eq:EquadW2}) is equivalent to the maximization of
the sum of heterogeneous quadratic functions on a Stiefel manifold.
The functions are heterogeneous in our case since the class-specific
covariance matrices $R_{k}$ are not identical in general. The Lagrangian
corresponding to this problem (with $Z$ removed via direct minimization)
is

\begin{equation}
L_{\mathrm{quad}}(W,\Lambda)=-\frac{1}{2}\sum_{k=1}^{K}w_{k}^{T}R_{k}w_{k}+\mathrm{trace}\left[\Lambda^{T}\left(W^{T}W-I_{K}\right)\right].\label{eq:Lagrangian_quad}
\end{equation}
Setting the gradient of the above Lagrangian w.r.t. $W$ to zero,
we obtain 

\begin{equation}
\left[R_{1}w_{1},R_{2}w_{2},\ldots,R_{K}w_{K}\right]=W(\Lambda+\Lambda^{T}).\label{eq:RW=00003DSWW1}
\end{equation}
Noting that $\Lambda+\Lambda^{T}$ is symmetric and using the Stiefel
orthonormality constraint $W^{T}W=I_{K}$, we get 

\begin{equation}
(\Lambda+\Lambda^{T})=W^{T}\left[R_{1}w_{1},R_{2}w_{2},\ldots,R_{K}w_{K}\right].
\end{equation}

The above can be considerably simplified. First we introduce a new
vector $\mathbf{w}\in M_{D,K}$ defined as $\mathbf{w}\equiv\left[w_{1}^{T},w_{2}^{T},\ldots,w_{K}^{T}\right]^{T}$
and then rewrite (\ref{eq:RW=00003DSWW1}) in vector form to get 

\begin{equation}
R\mathbf{w}=S(\mathbf{w})\mathbf{w}
\end{equation}
where 

\begin{equation}
R\equiv\left[\begin{array}{cccc}
R_{1} & 0_{K} & \cdots & 0_{K}\\
0_{K} & R_{2} & \cdots & 0_{K}\\
0_{K} & \cdots & \ddots & 0_{K}\\
0_{K} & \cdots & \cdots & R_{K}
\end{array}\right]\label{eq:Rdef}
\end{equation}
is a $KD\times KD$ matrix and 

\begin{equation}
S(\mathbf{w})\equiv\left[\begin{array}{ccc}
w_{1}^{T}R_{1}w_{1}I_{K} & \cdots & \frac{1}{2}\left(w_{1}^{T}R_{1}w_{K}+w_{K}^{T}R_{K}w_{1}\right)I_{K}\\
\frac{1}{2}\left(w_{1}^{T}R_{1}w_{2}+w_{2}^{T}R_{2}w_{1}\right)I_{K} & \cdots & \frac{1}{2}\left(w_{2}^{T}R_{2}w_{K}+w_{K}^{T}R_{K}w_{2}\right)I_{K}\\
\vdots & \ddots & \vdots\\
\frac{1}{2}\left(w_{1}^{T}R_{1}w_{K}+w_{K}^{T}R_{K}w_{1}\right)I_{K} & \cdots & w_{K}^{T}R_{K}w_{K}I_{K}
\end{array}\right]\label{eq:S(w)}
\end{equation}
a $KD\times KD$ \emph{symmetric} matrix. The reason $S(\mathbf{w})$
can be made symmetric is because it's closely related to the solution
to $(\Lambda+\Lambda)^{T}$\textemdash which has to be symmetric.
The first and second order necessary conditions for a vector $\mathbf{w}_{0}\in M_{D,K}$
to be a local minimum (feasible point) for the problem in (\ref{eq:EquadW2})
are as follows:

\begin{equation}
R\mathbf{w}_{0}=S(\mathbf{w}_{0})\mathbf{w}_{0}\label{eq:1storderneccond}
\end{equation}
and 

\begin{equation}
\left(R-S(\mathbf{w}_{0})\right)|_{TM(\mathbf{w}_{0})}\label{eq:2ndordneccond}
\end{equation}
is negative semi-definite. In (\ref{eq:2ndordneccond}), $TM(\mathbf{w}_{0})$
is the tangent space of the Stiefel manifold at $\mathbf{w}_{0}$.
In a \emph{tour de force} proof, Rapcs\'{a}k further shows in \cite{Rapcsak2002}
that if the matrix $\left(R-S(\mathbf{w}_{0})\right)$ is negative
semi-definite, then a feasible point $\mathbf{w}_{0}$ is a \emph{global
minimum}. This is an important result since it adds a sufficient condition
for a global minimum for the problem of minimizing a heterogeneous
sum of quadratic forms on a Stiefel manifold.\footnote{Note that this problem is fundamentally different from and cannot
be reduced to the minimization of $\mathrm{trace}\left(AW^{T}BW\right)$
subject to $W^{T}W=I_{K}$ which has a closed form solution.}

\subsubsection{The RKHS setting}

We can readily extend the above analysis to the kernel version of
the squared inner product. The complete objective function w.r.t.
both the coefficients $A$ and the auxiliary variable $Z$ is 

\begin{equation}
E_{\mathrm{Kequiv}}(A,Z)=\sum_{k=1}^{K}\sum_{i_{k}\in C_{k}}\left[-\sum_{j=1}^{N}z_{ki_{k}}\alpha_{kj}K(x_{j},x_{i_{k}})+\frac{1}{2}z_{ki_{k}}^{2}\right].
\end{equation}
Setting $z_{ki_{k}}=\sum_{j=1}^{N}\alpha_{kj}K(x_{j},x_{i_{k}})$
which is the optimum solution for $Z$, we get

\begin{eqnarray}
E_{\mathrm{Kquad}}(A) & = & -\frac{1}{2}\sum_{k=1}^{K}\sum_{i_{k}\in C_{k}}\left[\sum_{j=1}^{N}\alpha_{kj}\left(K(x_{j},x_{i_{k}})-\frac{1}{|C_{k}|}\sum_{i_{k}\in C_{k}}K(x_{j},x_{i_{k}})\right)\right]^{2}\nonumber \\
 & = & -\frac{1}{2}\sum_{k=1}^{K}\boldsymbol{\alpha}_{k}^{T}G_{k}\boldsymbol{\alpha}_{k}\label{eq:EKquadalpha}
\end{eqnarray}
where $\left[\boldsymbol{\alpha}_{k}\right]_{j}=\alpha_{kj}$, $A=\left[\boldsymbol{\alpha}_{1},\boldsymbol{\alpha}_{2},\ldots,\boldsymbol{\alpha}_{K}\right]^{T}$
and

\begin{eqnarray}
\left[G_{k}\right]_{jm} & \equiv & \sum_{i_{k}\in C_{k}}\left(K(x_{j},x_{i_{k}})-\frac{1}{|C_{k}|}\sum_{i\in C_{k}}K(x_{j},x_{i})\right)\nonumber \\
 &  & \,\,\,\,\,\cdot\left(K(x_{m},x_{i_{k}})-\frac{1}{|C_{k}|}\sum_{i\in C_{k}}K(x_{m},x_{i})\right)\label{eq:Gkjm}
\end{eqnarray}
The constraints on $A$ can be written as 

\begin{equation}
AGA^{T}=I_{K}\Rightarrow\left(G^{\frac{1}{2}}A^{T}\right)^{T}\left(G^{\frac{1}{2}}A^{T}\right)=I_{K}.
\end{equation}
Introducing a new variable $B=G^{\frac{1}{2}}A^{T}$, we may rewrite
the kernel objective function and constraints as

\begin{equation}
E_{\mathrm{Kquadnew}}(B)=-\frac{1}{2}\sum_{k=1}^{K}\boldsymbol{\beta}_{k}^{T}H\boldsymbol{\beta}_{k}\equiv-\frac{1}{2}\sum_{k=1}^{K}\boldsymbol{\beta}_{k}^{T}G^{-\frac{1}{2}}G_{k}G^{-\frac{1}{2}}\boldsymbol{\beta}_{k}\label{eq:EKquadnew}
\end{equation}
(where $B\equiv\left[\boldsymbol{\beta}_{1},\boldsymbol{\beta}_{2},\ldots,\boldsymbol{\beta}_{K}\right]$)
and

\begin{equation}
B^{T}B=I_{K}\label{eq:Bconstraints}
\end{equation}
respectively. This is now in the same form as the objective function
and constraints in Section~\ref{subsec:Euclidean-setting} and therefore
the Rapcs\'{a}k analysis of that section can be directly applied
here. The above change of variables is predicated on the positive
definiteness of $G$. If this is invalid, principal component analysis
has to be applied to $G$ resulting in a positive definite matrix
in a reduced space after which the above approach can be applied. 

In addition to providing necessary conditions for global minima, the
authors in \cite{BollaMichaletzkyTusnadyEtAl1998} developed an iterative
procedure as a method for a solution. We have adapted this to suit
our purposes. A block coordinate descent algorithm which successively
updates $W$ and $Z$ is presented in Algorithm~\ref{alg:Iterative-Process}

\begin{algorithm}[H]
\begin{itemize}
\item \textbf{Input}: A set of labeled patterns $\left\{ x_{i_{k}}\right\} _{1}^{|C_{k}|},\forall k\in\left\{ 1,\ldots,K\right\} $.
\item \textbf{Initialize}:
\begin{itemize}
\item Convergence threshold~~$\epsilon$.
\item Arbitrary orthonormal system $W^{\left(0\right)}$.
\end{itemize}
\item \textbf{Repeat}
\begin{itemize}
\item Calculate the sequence $\left[\begin{array}{cccc}
W^{\left(1\right)}, & W^{\left(2\right)}, & \ldots & ,W^{\left(m\right)}\end{array}\right]$ . Assume $W^{\left(m\right)}$ is constructed for $m=0,1,2,\ldots$
\item Update the auxiliary variable $Z^{(m+1)}$~, under the constraint
~$\sum_{i_{k}\in C_{k}}z_{ki_{k}}=0,\forall k$,
\begin{itemize}
\item $z_{ki_{k}}^{(m+1)}=\left(w^{(m)}\right)_{k}^{T}x_{i_{k}}-\frac{1}{|C_{k}|}\sum_{i_{k}\in C_{k}}\left(w^{(m)}\right)_{k}^{T}x_{i_{k}}$for
the sum of squares of inner products objective function.
\end{itemize}
\item Perform the SVD decomposition on $\left[\sum_{i_{1}\in C_{1}}z_{1i_{1}}^{(m+1)}x_{i_{1}},\ldots,\sum_{i_{k}\in C_{k}}z_{ki_{k}}^{(m+1)}x_{i_{k}}\right]$
to get $U^{(m+1)}S^{(m+1)}\left(V^{(m+1)}\right)^{T}$ where $S^{(m+1)}$
is $K\times K$. 
\item $W^{(m+1)}=U^{(m+1)}\left(V^{(m+1)}\right)^{T}$, the polar decomposition.
\end{itemize}
\item \textbf{Loop until} $\|W^{(m+1)}-W^{(m)}\|_{F}\leq\epsilon$.
\item \textbf{Output}: $W$
\end{itemize}
\caption{Iterative process for minimization of the sum of squares of inner
products objective function. \label{alg:Iterative-Process}}
\end{algorithm}

\section{Experimental Results}

\subsection{Quantitative results for linear and kernel dimensionality reduction}

In practice, dimensionality reduction is used in conjunction with
a classification algorithm. By definition the purpose of dimensionality
reduction as it relates to classification, is to reduce the complexity
of the data while retaining discriminating information. Thus we utilize
a popular classification algorithm in order to analyze the performance
of our proposed dimensionality reduction technique. In this section,
we report the results of several experiments with dimensionality reduction
combined with SVM classification. In the multi-class setting, we compare
against other state-of-the-art algorithms that perform dimensionality
reduction and then evaluate the performance using the multi-class
one-vs-all linear SVM scheme. The classification technique uses the
traditional training and testing phases, outputting the class it considers
the best prediction for a given test sample. We measure the accuracy
of these predictions averaged over all test sets. In Table~(\ref{tab:List-of-test-results}),
we demonstrate the effectiveness of both the sum of quadratic and
absolute value functions, denoted as category quadratic space (CQS)
and category absolute value space (CAS) respectively. Then, we benchmark
their overall classification accuracy against several classical dimensionality
reduction techniques, namely, least squares linear discriminant analysis
(LS-LDA) \cite{Ye2007}, Fisher linear discriminant (MC-FLD) \cite{fisher1936use},
principal component analysis (PCA) \cite{rao1964use} and their multi-class
and kernel counterparts (when applicable). In each experiment, we
choose two thirds of the data for training and the remaining third
of the samples were used for testing. The results are shown in Table~(\ref{tab:List-of-test-results}).

\textbf{Databases}: To illustrate the performance of the methods proposed
in Section~\ref{Category-Vector-Space}, we conducted experiments
using different publicly available data sets taken from the UCI machine
learning data repository \cite{Lichman:2013}. We have chosen a variety
of data sets that vary in terms of class cardinality ($K$), samples
($N$) and number of features ($D$) to demonstrate the versatility
of our approach. For a direct comparison of results, we chose the
same data sets:; Vehicle, Wine, Iris, Seeds, Thyroid, Satellite, Segmentation,
and Vertebral Silhouettes recognition databases. More details about
the individual sets are available at the respective repository sites.

We divide the results into the linear and kernel groups (as is normal
practice). The obtained results for linear dimensionality reduction
with SVM linear classification are shown in Table~(\ref{tab:List-of-test-results})
. All dimensionality reduction algorithms were implemented and configured
for optimal classification results (via cross-validation) with a linear
SVM classifier. It can be seen that the category space projection
scheme consistently provides a good projection for a standard classification
algorithms to be executed. Several of the data sets are comprise only
three classes and it can be seen that the proposed method is competitive
in performance and in some instances performs slightly better. 

\begin{table}[H]
\medskip{}

\caption{Linear dimensionality reduction w/ SVM classification..\label{tab:List-of-test-results}}

\centering{}%
\begin{tabular}{lccccc}
\hline 
\textbf{Name (\# Classes)} & \textbf{CQS} & \textbf{CAS} & \textbf{LS-LDA} & \textbf{PCA} & \textbf{MC-FLD}\tabularnewline
\hline 
Vehicle (4) & 53.91 & 53.05 & 76.56 & 55.36 & 76.82\tabularnewline
Wine (3) & 96.07 & 96.82 & 95.51 & 77.19 & 97.28\tabularnewline
Iris (3) & 97.55 & 96.88 & 96.11 & 96.77 & 96.77\tabularnewline
Seeds (3) & 90.39 & 90.79 & 95.15 & 92.53 & 95.79\tabularnewline
thyroid (3) & 94.02 & 94.08 & 94.02 & 92.57 & 93.92\tabularnewline
Satellite (6) & 85.30 & 85.20 & 86.38 & 85.45 & 86.52\tabularnewline
Segmentation (7) & 93.14 & 93.44 & 94.62 & 94.40 & 94.43\tabularnewline
Vertebral (3) & 84.13 & 82.79 & 81.45 & 80.05 & 81.18\tabularnewline
\hline 
\end{tabular}
\end{table}

Also for comparison, Table~(\ref{tab:List-of-test-results-kernel-norm})
reports the performance of the proposed kernel formulations followed
by a linear SVM classifier. These proposed methods also achieve accuracy
rates similar to their kernel counterparts. 

\begin{table}[H]
\medskip{}

\caption{Kernel dimensionality reduction w/ SVM classification. \label{tab:List-of-test-results-kernel-norm}}

\centering{}%
\begin{tabular}{lcccc}
\hline 
\textbf{Name (\# Classes)} & \textbf{K-CQS} & \textbf{K-CAS} & \textbf{K-PCA} & \textbf{K-MC-FLD}\tabularnewline
\hline 
Vehicle (4) & 40.27 & 40.92 & 44.81 & 74.35\tabularnewline
Wine (3) & 92.95 & 95.63 & 95.95 & 96.88\tabularnewline
Iris (3) & 95.55 & 93.33 & 95.55 & 94.44\tabularnewline
Seeds (3) & 90.21 & 90.47 & 91.53 & 93.65\tabularnewline
thyroid (3) & 41.97 & 40.24 & 43.08 & 72.34\tabularnewline
Satellite (6) & 81.54 & 86.23 & 89.69 & 90.61\tabularnewline
Segmentation (7) & 72.96 & 77.24 & 83.01 & 92.43\tabularnewline
Vertebral (3) & 70.96 & 69.53 & 70.96 & 82.25\tabularnewline
\hline 
\end{tabular}
\end{table}

\begin{table}[H]
\medskip{}

\caption{Kernel dimensionality reduction w/ angle classification. \label{tab:List-of-test-results-kernel-zscore-angle}}

\centering{}%
\begin{tabular}{lcc}
\hline 
\textbf{Name (\# Classes)} & \textbf{K-CQS-A} & \textbf{K-CAS-A}\tabularnewline
\hline 
Vehicle (4) & 67.96 & 68.24\tabularnewline
Wine (3) & 95.32 & 95.32\tabularnewline
Iris (3) & 95.55 & 95.18\tabularnewline
Seeds (3) & 91.79 & 91.79\tabularnewline
thyroid (3) & 67.90 & 66.79\tabularnewline
Satellite (6) & 83.33 & 76.29\tabularnewline
Segmentation (7) & 50.21 & 48.94\tabularnewline
Vertebral (3) & 77.59 & 77.77\tabularnewline
\hline 
\end{tabular}
\end{table}

The iterative approach in Algorithm (\ref{alg:Iterative-Process})
was applied to obtain an optimal orthonormal basis $W$ (which is
$D\times K$) for the category space, where $D$ dimensional input
patterns can be projected to the smaller $K$ dimensional category
space if $D>K$. We start with a set of $N$ labeled, input vectors
$x_{i}\in\mathbf{R}^{D}$ drawn randomly from multiple classes $C_{k},\,k\in\{1,\ldots,K\}$.
The optimization technique searches over Steifel manifold elements
as explained above. The algorithm is terminated when the Frobenius
norm difference between iterations, $\|W^{(m-1)}-W^{(m)}\|_{F}\le\delta$
(with $\delta=10^{-8}$). Once we have determined the optimal $W$,
the patterns are mapped to the category space by the transformation
$y_{i}=W^{T}x_{i}$ , to obtain the corresponding set of $N$ samples
$y_{i}\in\mathbf{R}^{K}$ , where $K$ is the reduced dimensional
space. 

The results above show that our proposed methods lead to classification
rates that can be compared to classical approaches. But, the main
focus of this work is to provide an algorithm that retains important
classification information while introducing a geometry (category
vector subspace) which has attractive semantic and visualization properties.
The results suggest that our classification results are competitive
with other techniques while learning a category space.

\subsection{Visualization of kernel dimensionality reduction}

Another valuable aspect of this research can be seen in the kernel
formulation which demonstrates warping of the projected patterns towards
their respective category axes. This suggests a geometric approach
to classification, i.e. we could consider the angle of deviation of
a test set pattern from each category axis as a measure of class membership.
Within the category space, a base category is represented by the bases
(axes) that define the category space. Class membership is therefore
inversely proportional to the angle between the pattern and the respective
category axis. Figures~(\ref{fig:category-space-reduced}) through
(\ref{fig:category-space-reduced-lg-sigma}) illustrate the warped
space for various three class problems, for a variation in the width
parameter ($\sigma)$ of a Gaussian radial basis function kernel in
the range $\sigma=[0.1,\,0.8]$. Note the improved visualization semantics
of the category space approach when compared to the other dimensionality
reduction techniques.
\begin{center}
\begin{figure}[H]
\begin{centering}
\begin{tabular}{|c|c|c|}
\hline 
\includegraphics[scale=0.15]{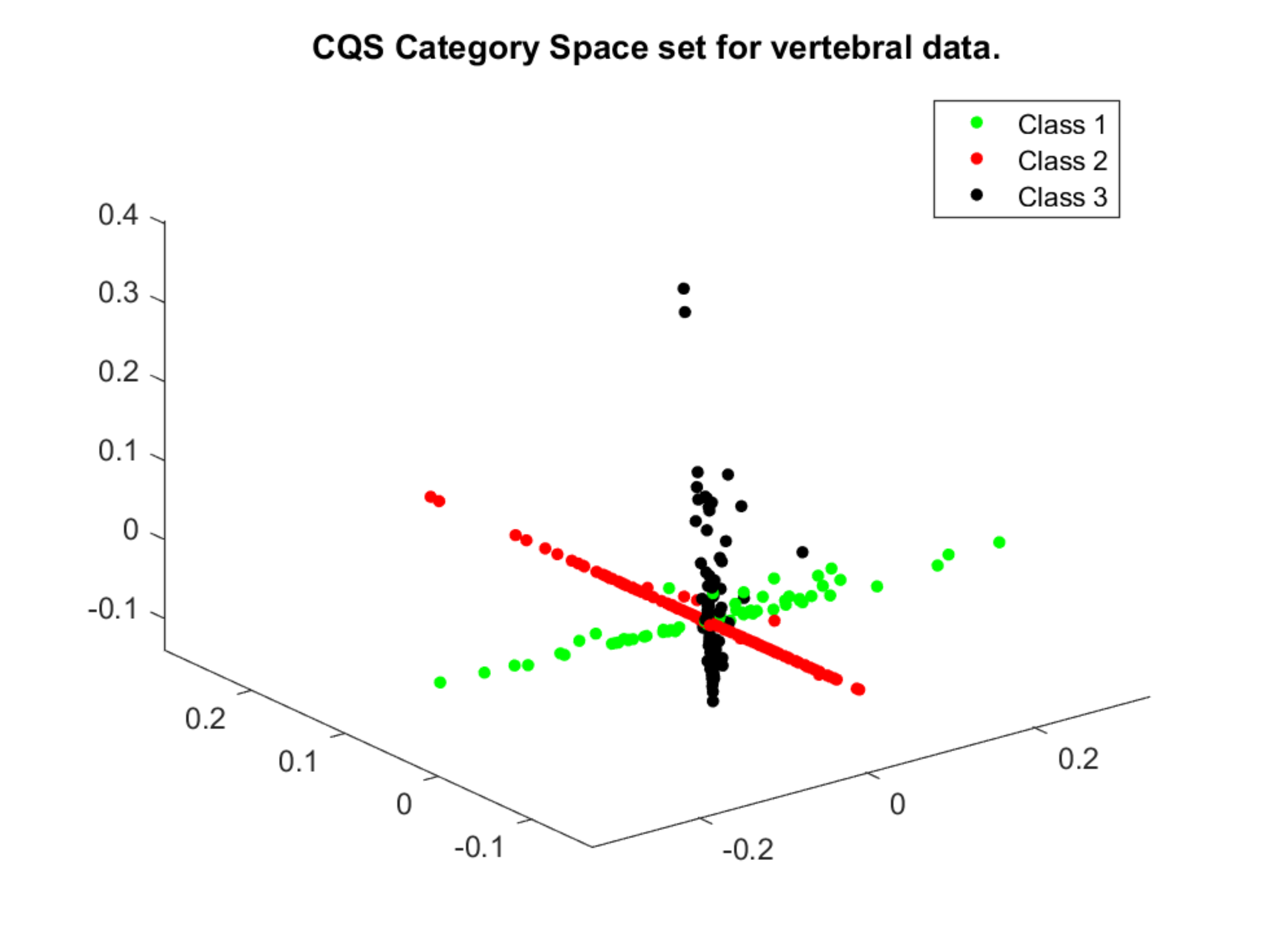} & \includegraphics[scale=0.15]{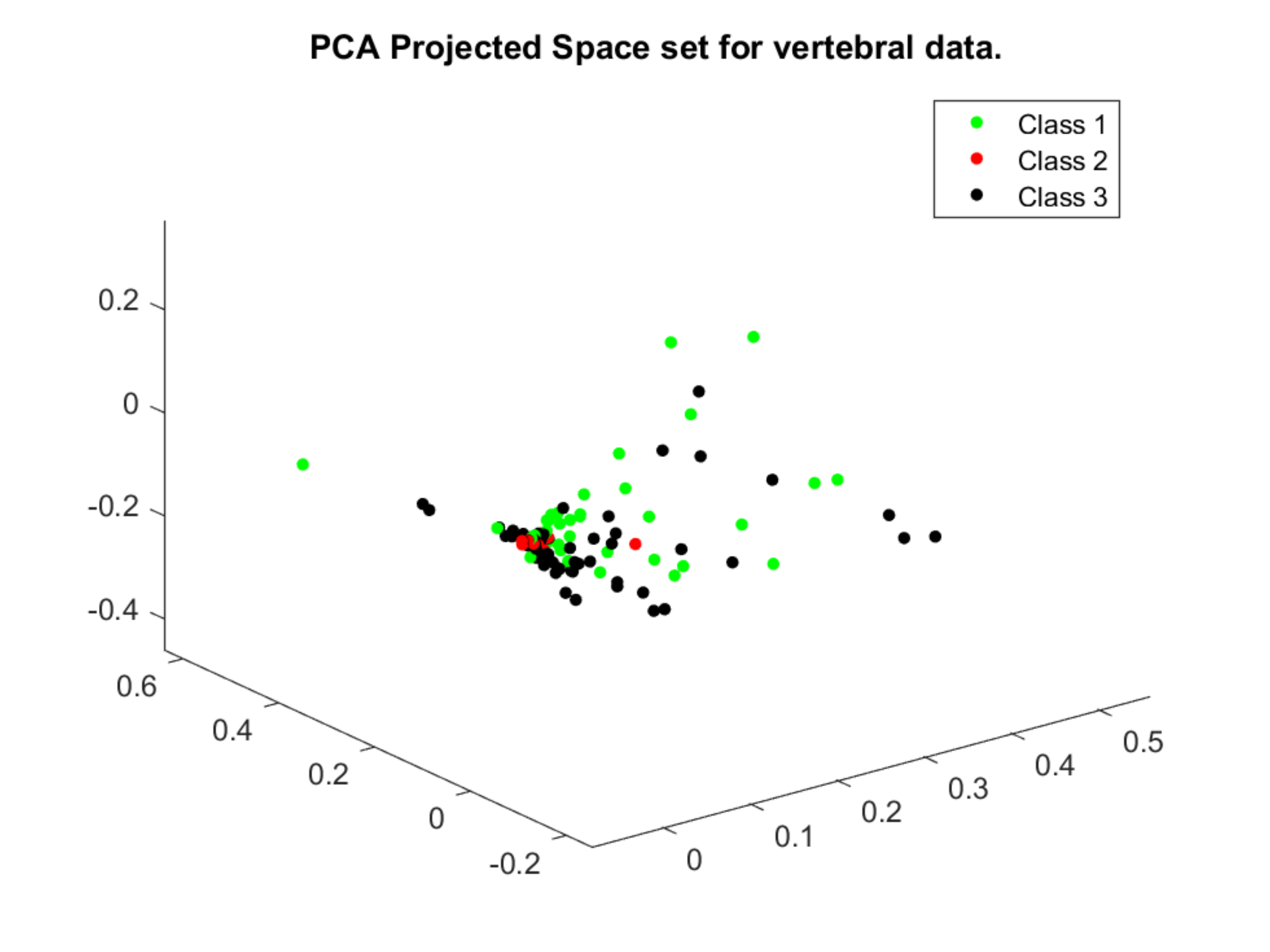} & \includegraphics[scale=0.15]{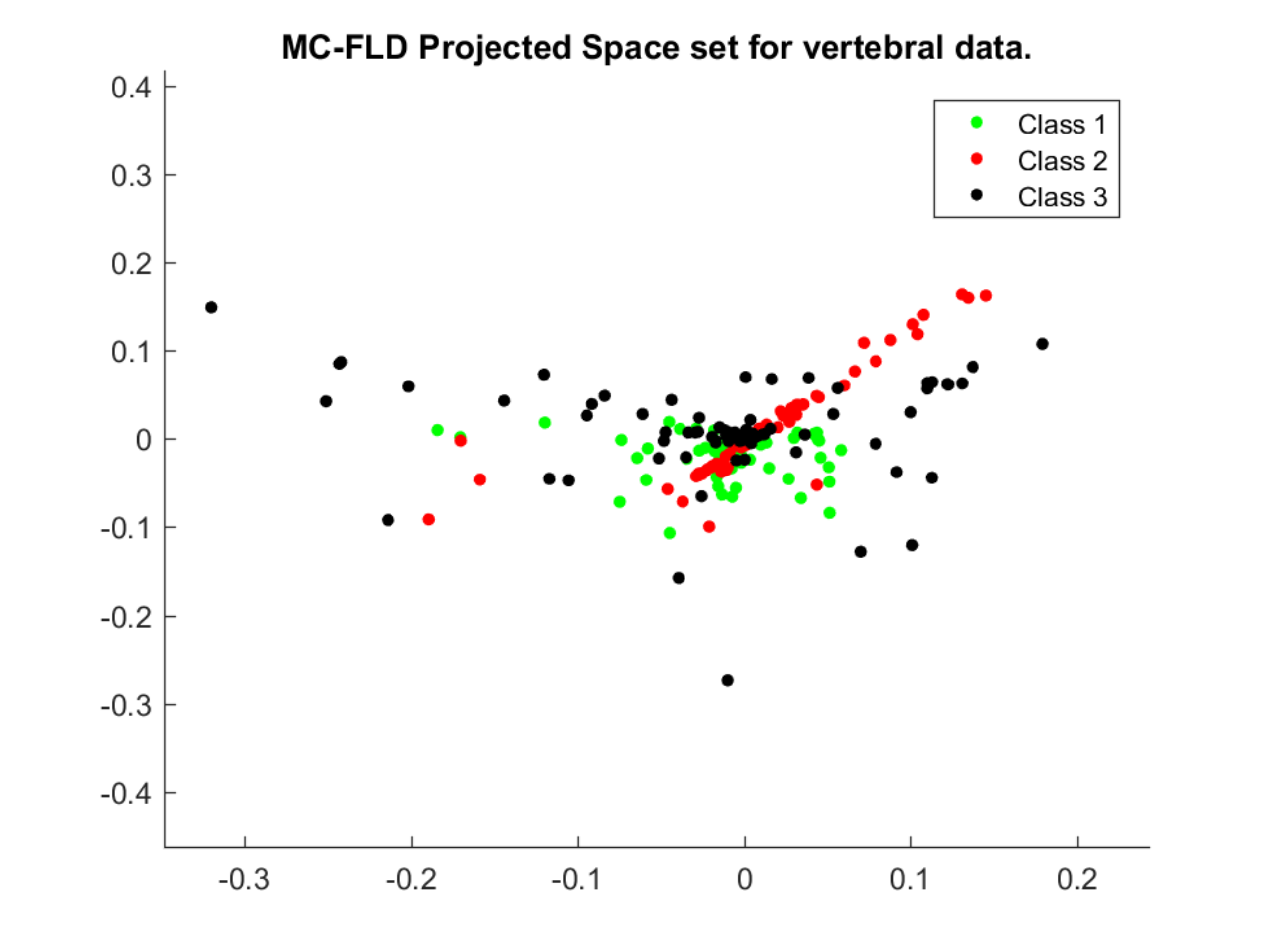}\tabularnewline
\hline 
\includegraphics[scale=0.15]{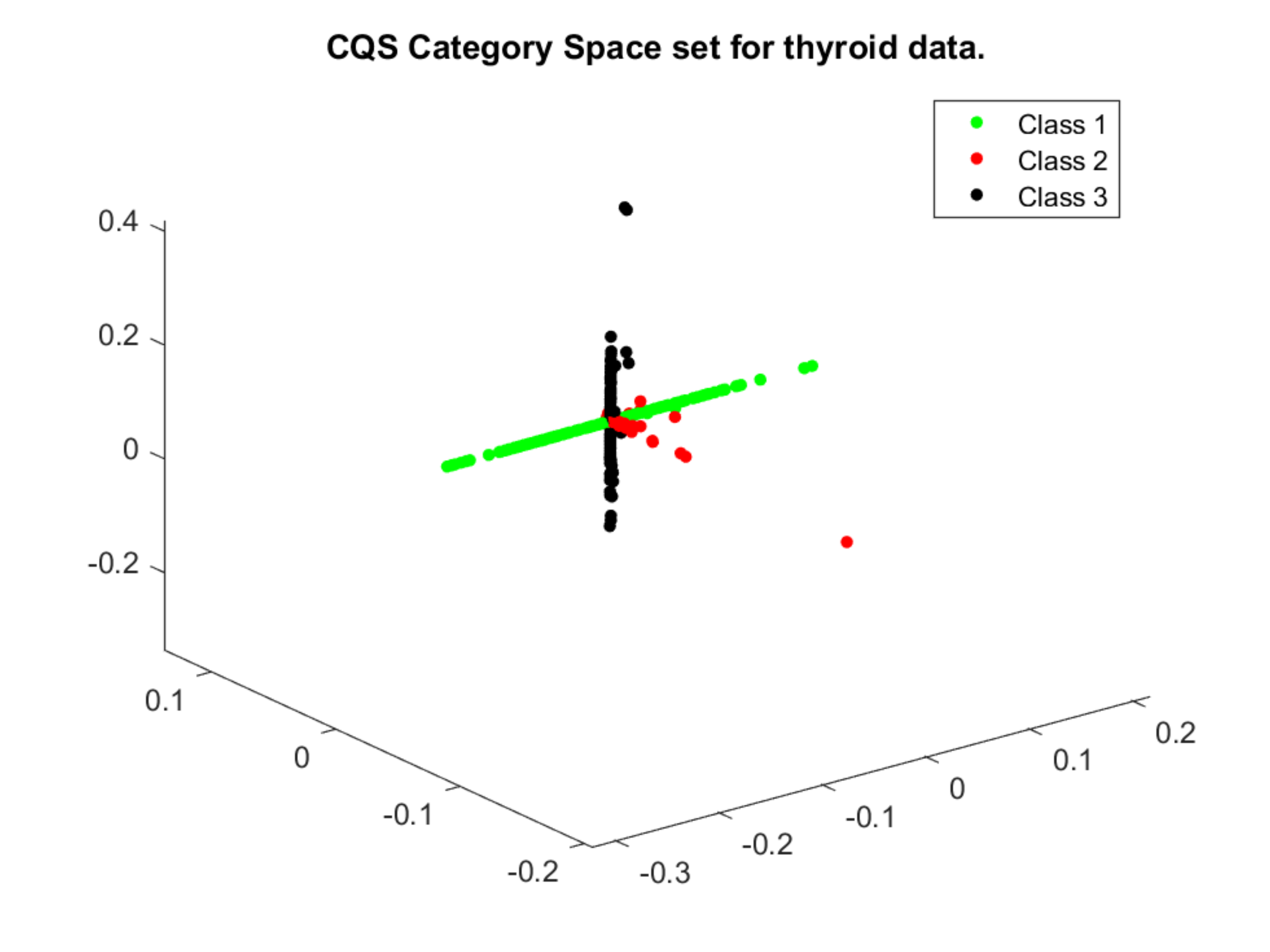} & \includegraphics[scale=0.15]{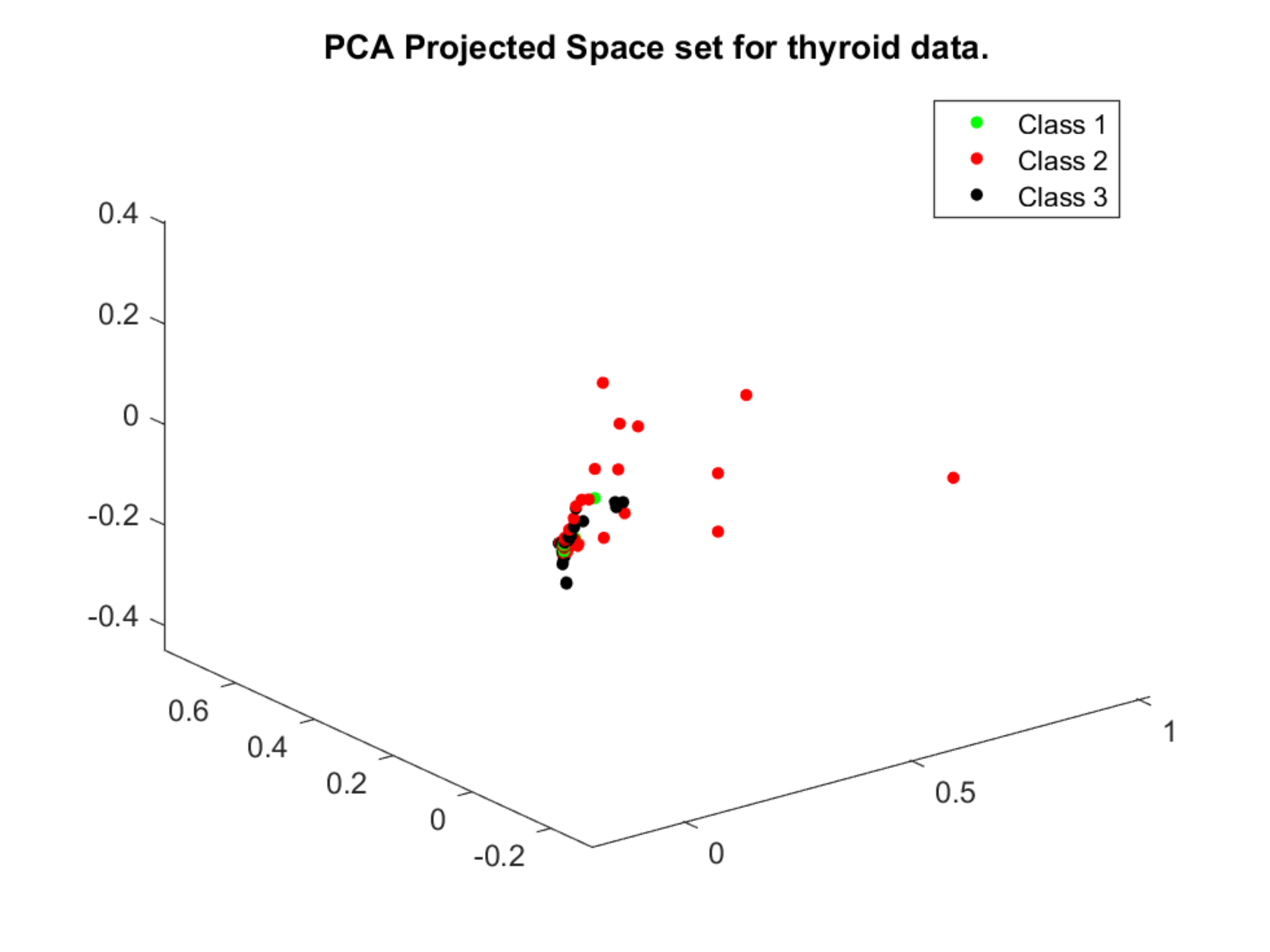} & \includegraphics[scale=0.15]{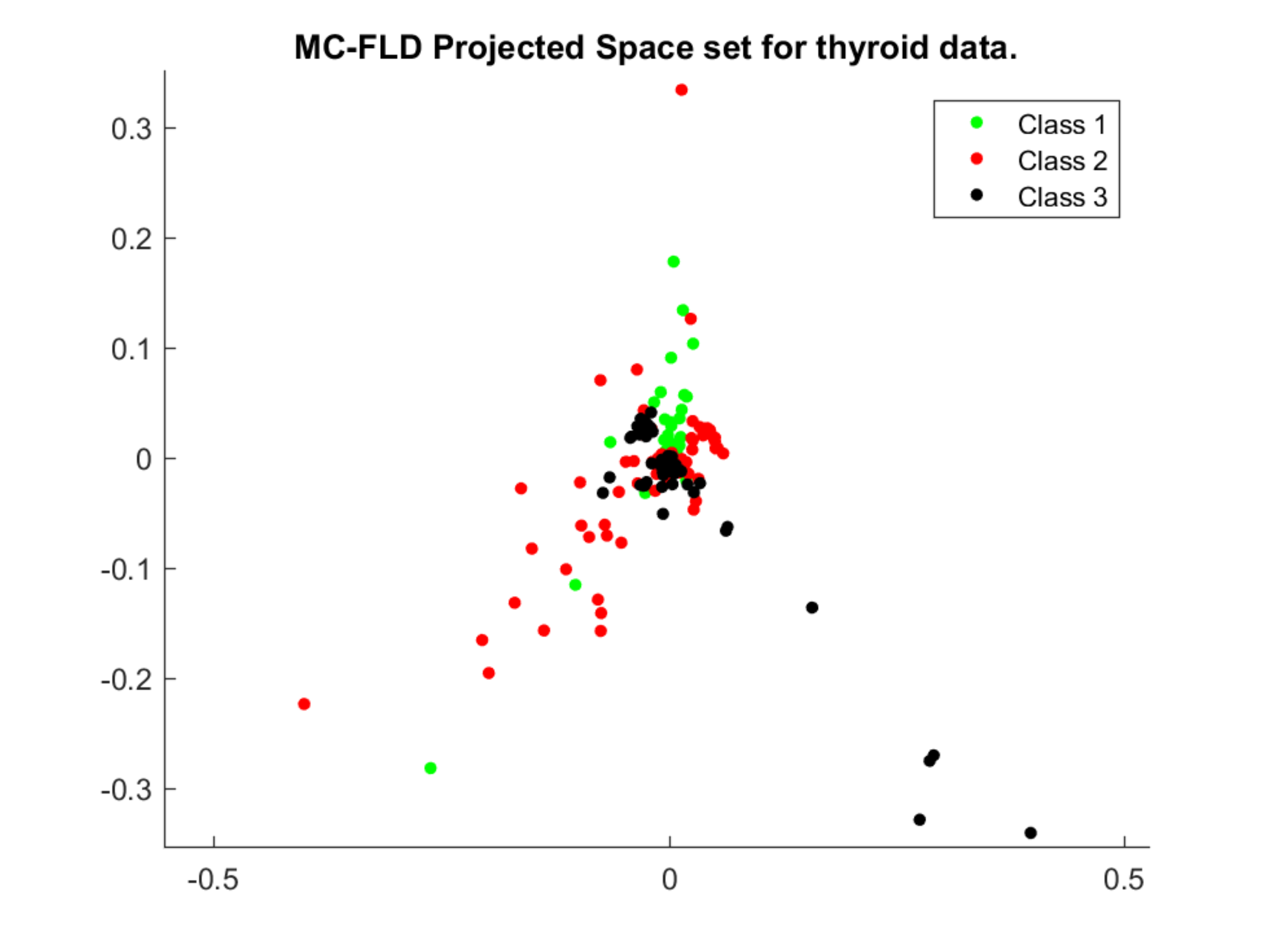}\tabularnewline
\hline 
\includegraphics[scale=0.15]{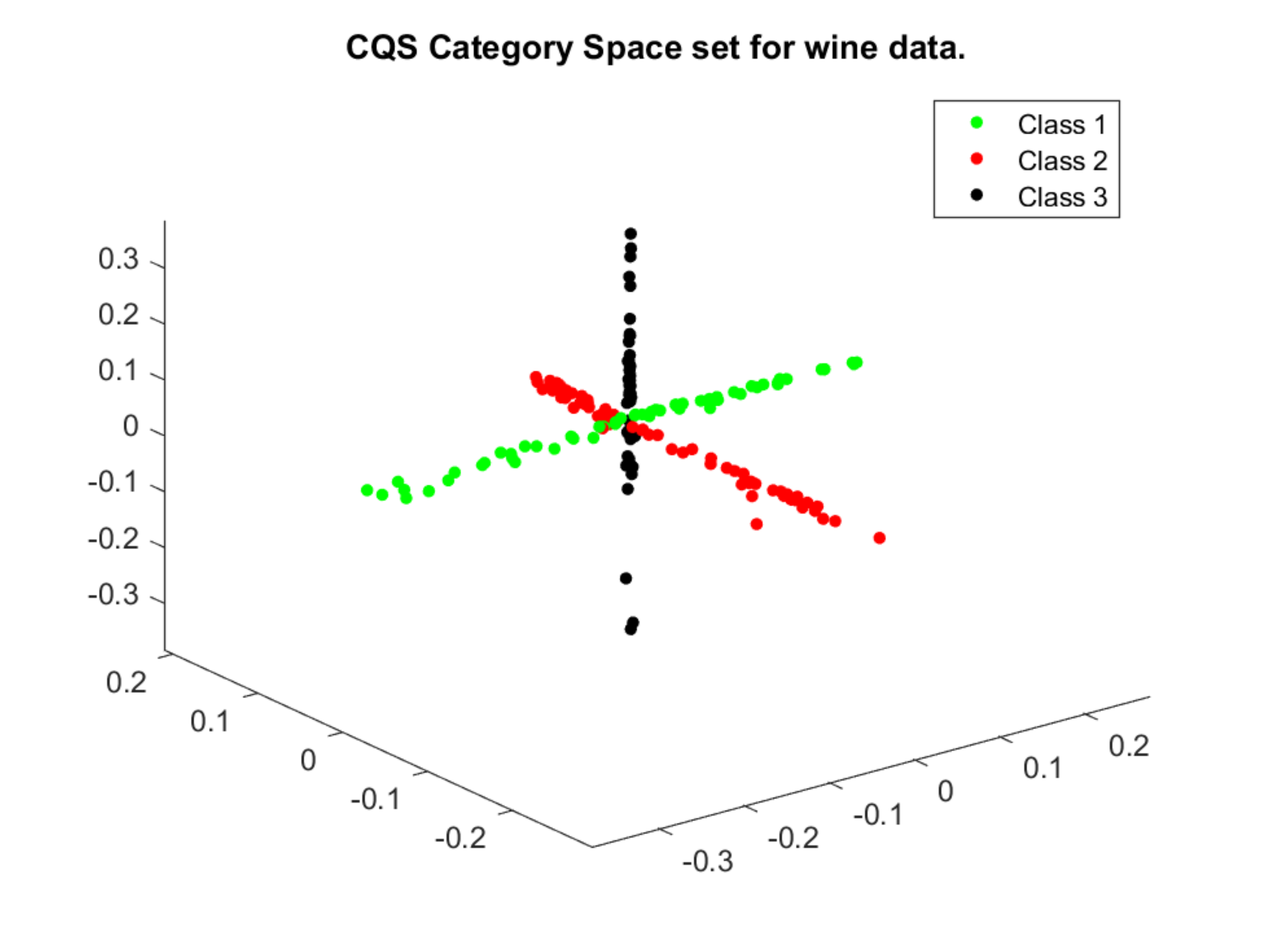} & \includegraphics[scale=0.15]{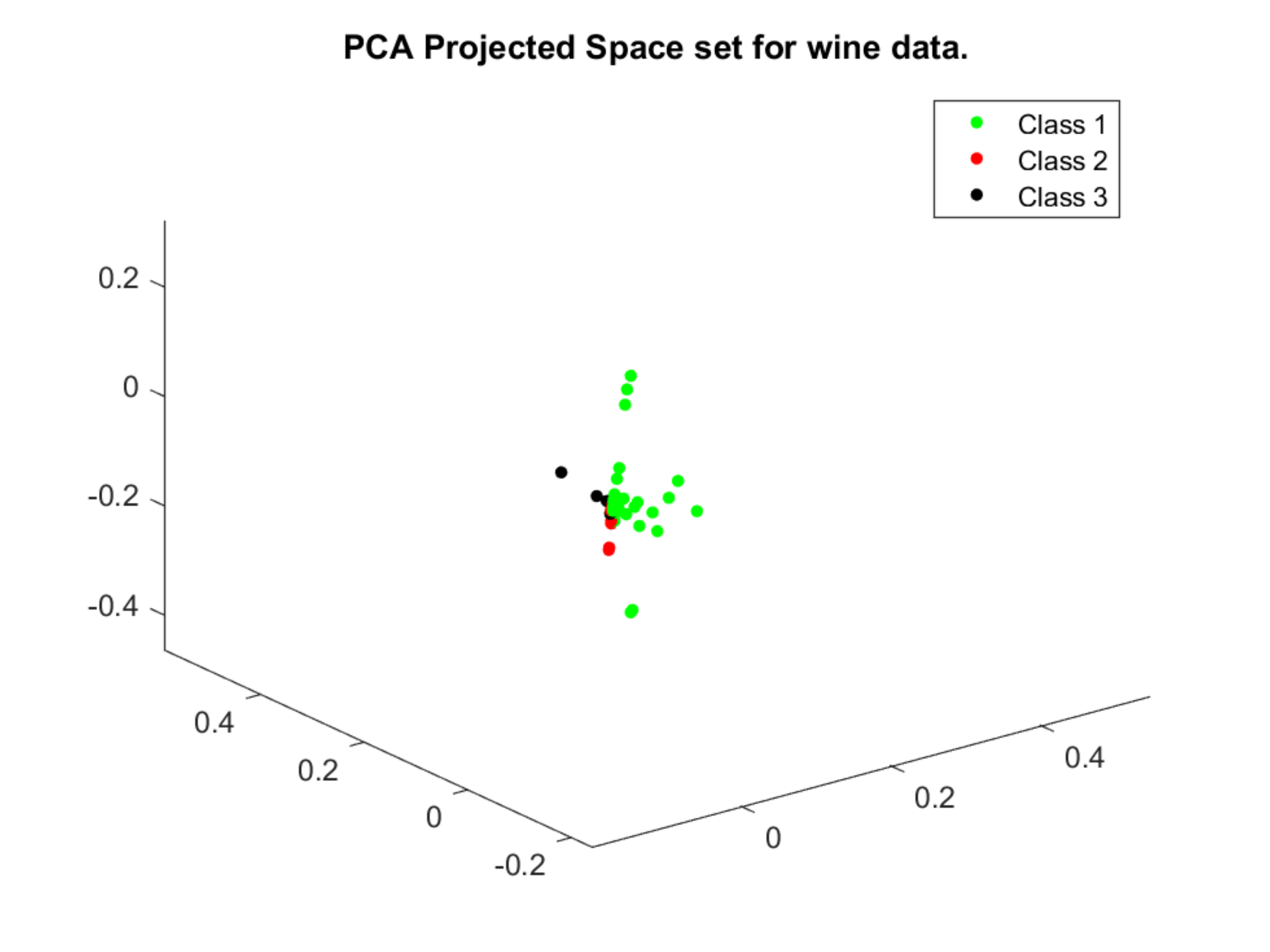} & \includegraphics[scale=0.15]{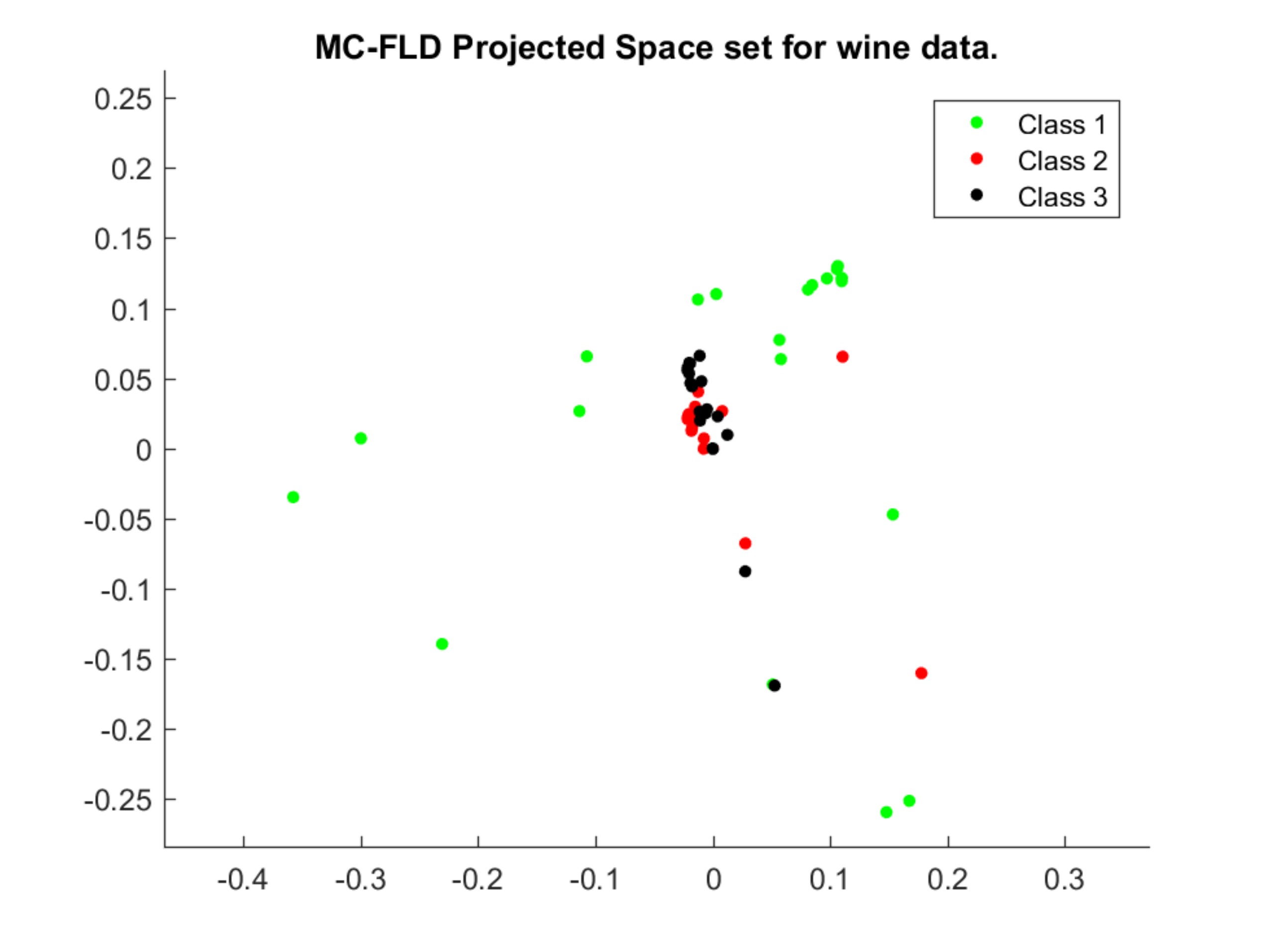}\tabularnewline
\hline 
\includegraphics[scale=0.15]{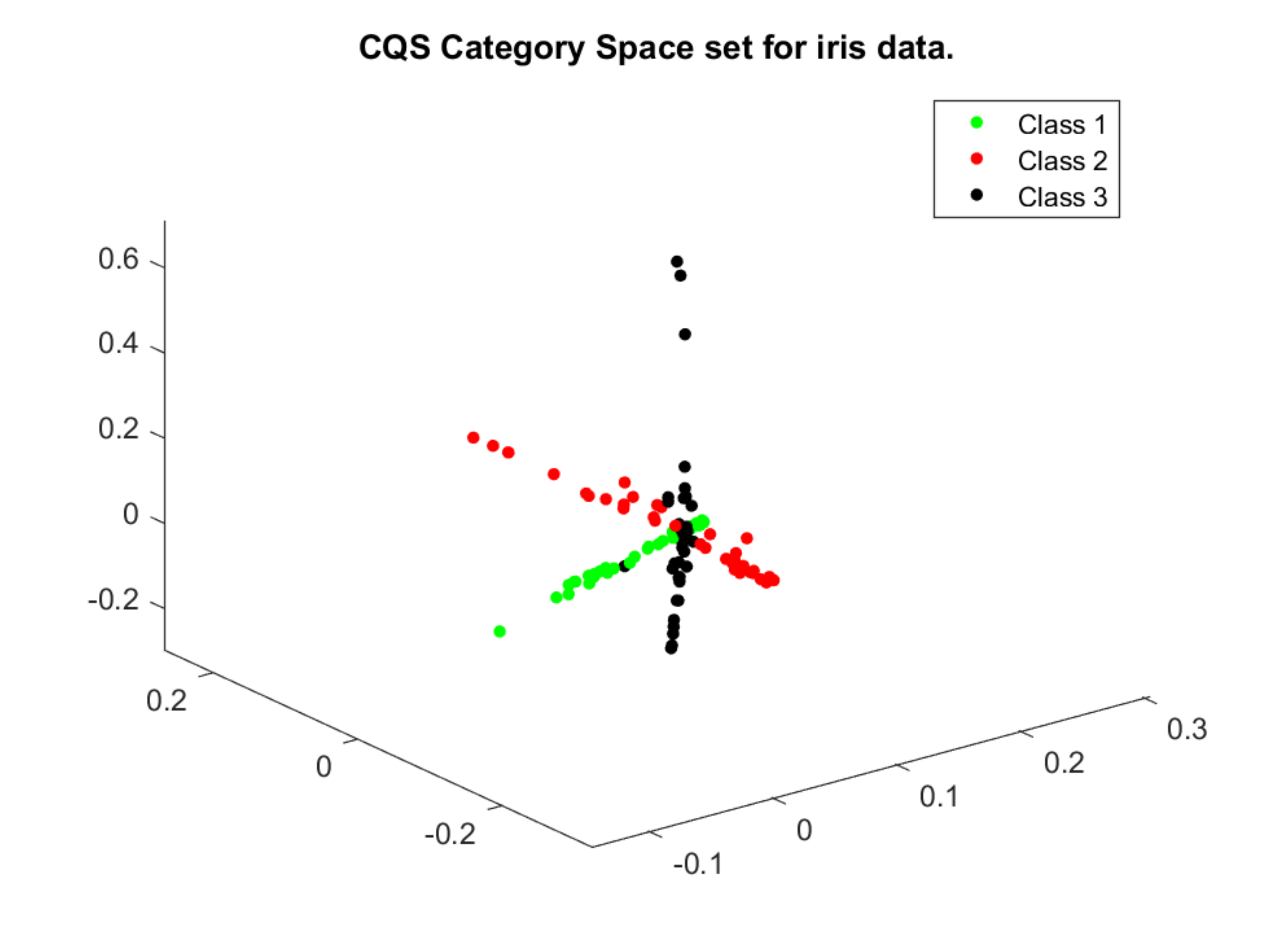} & \includegraphics[scale=0.15]{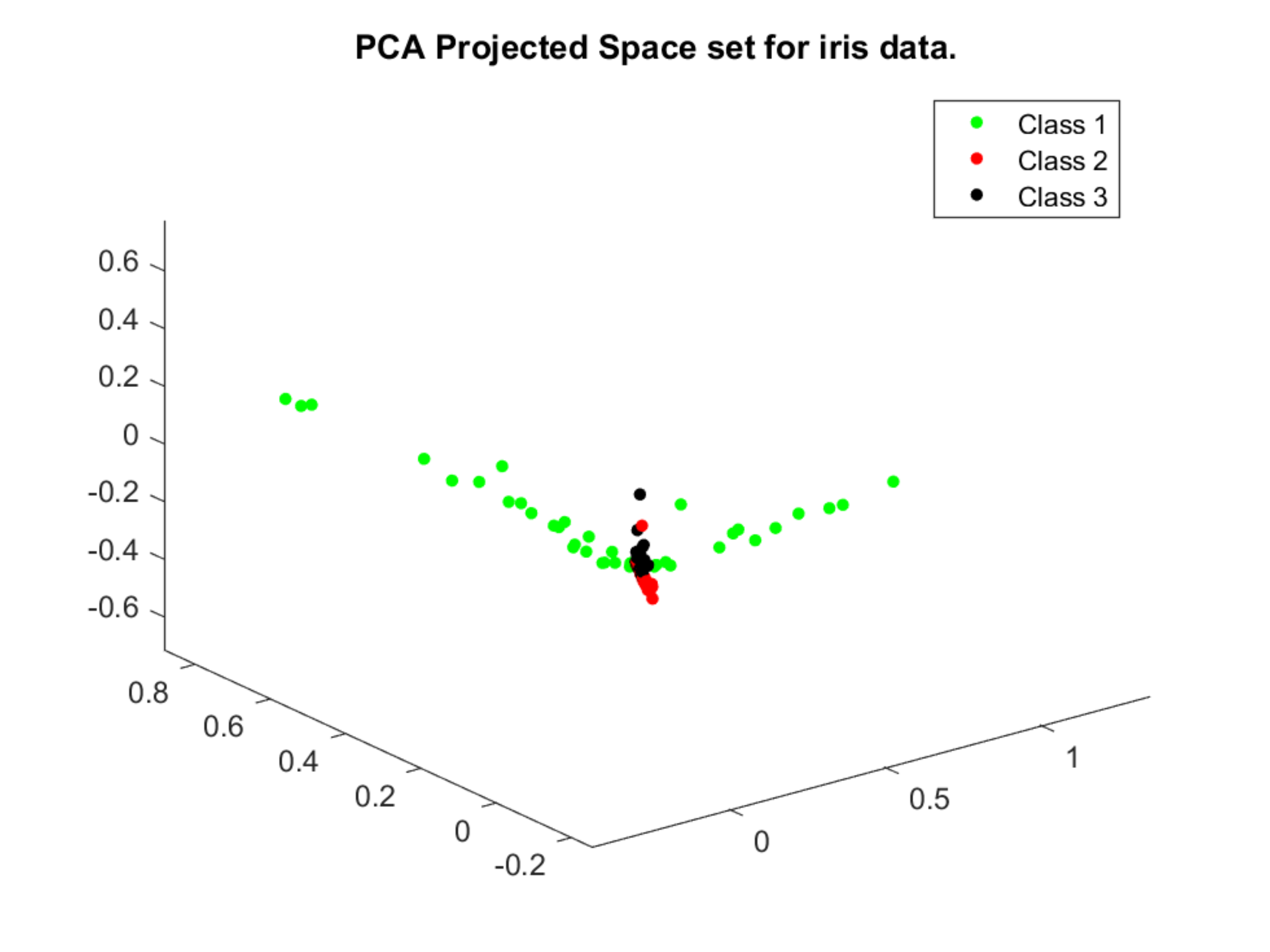} & \includegraphics[scale=0.15]{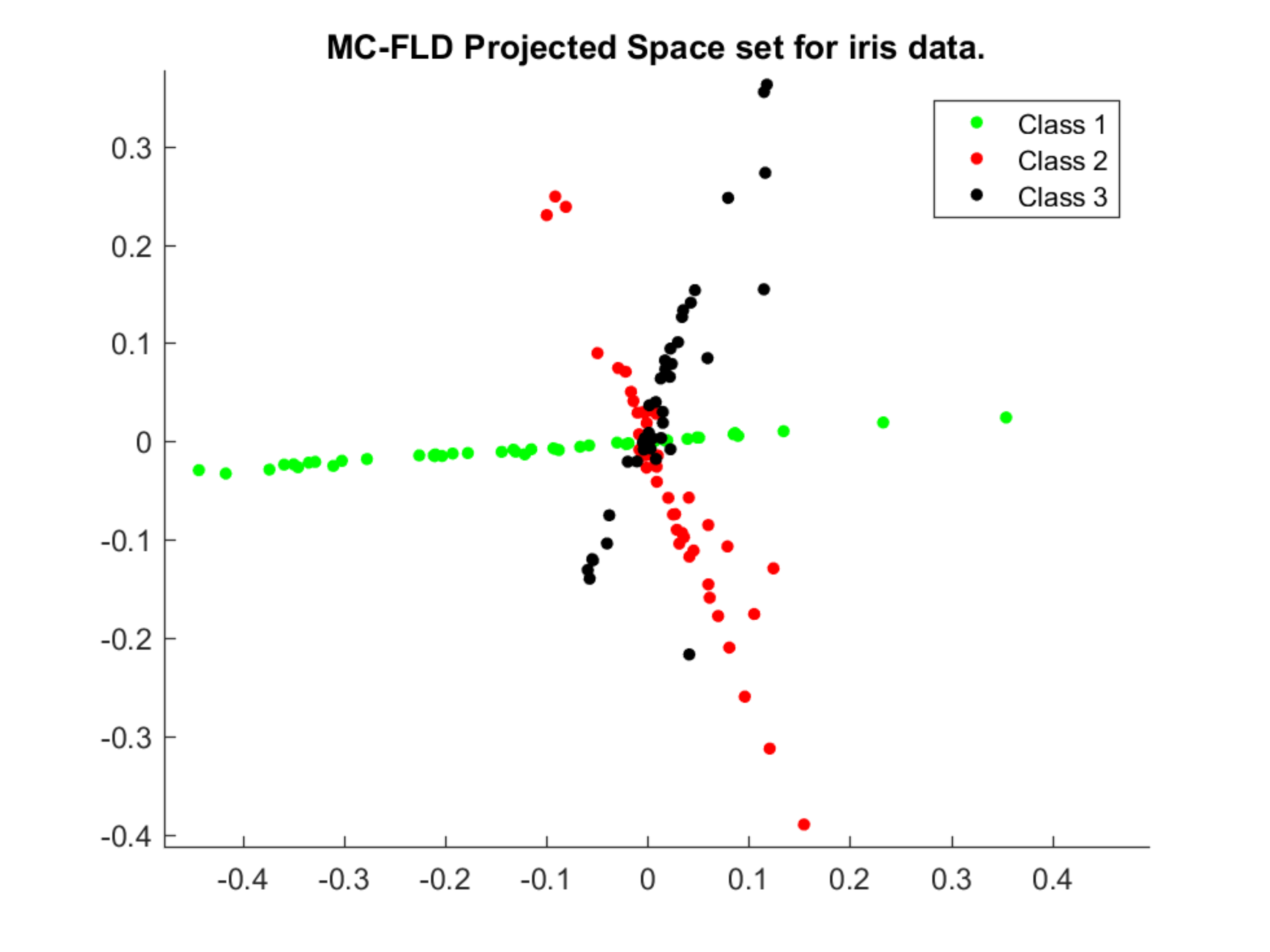}\tabularnewline
\hline 
\includegraphics[scale=0.15]{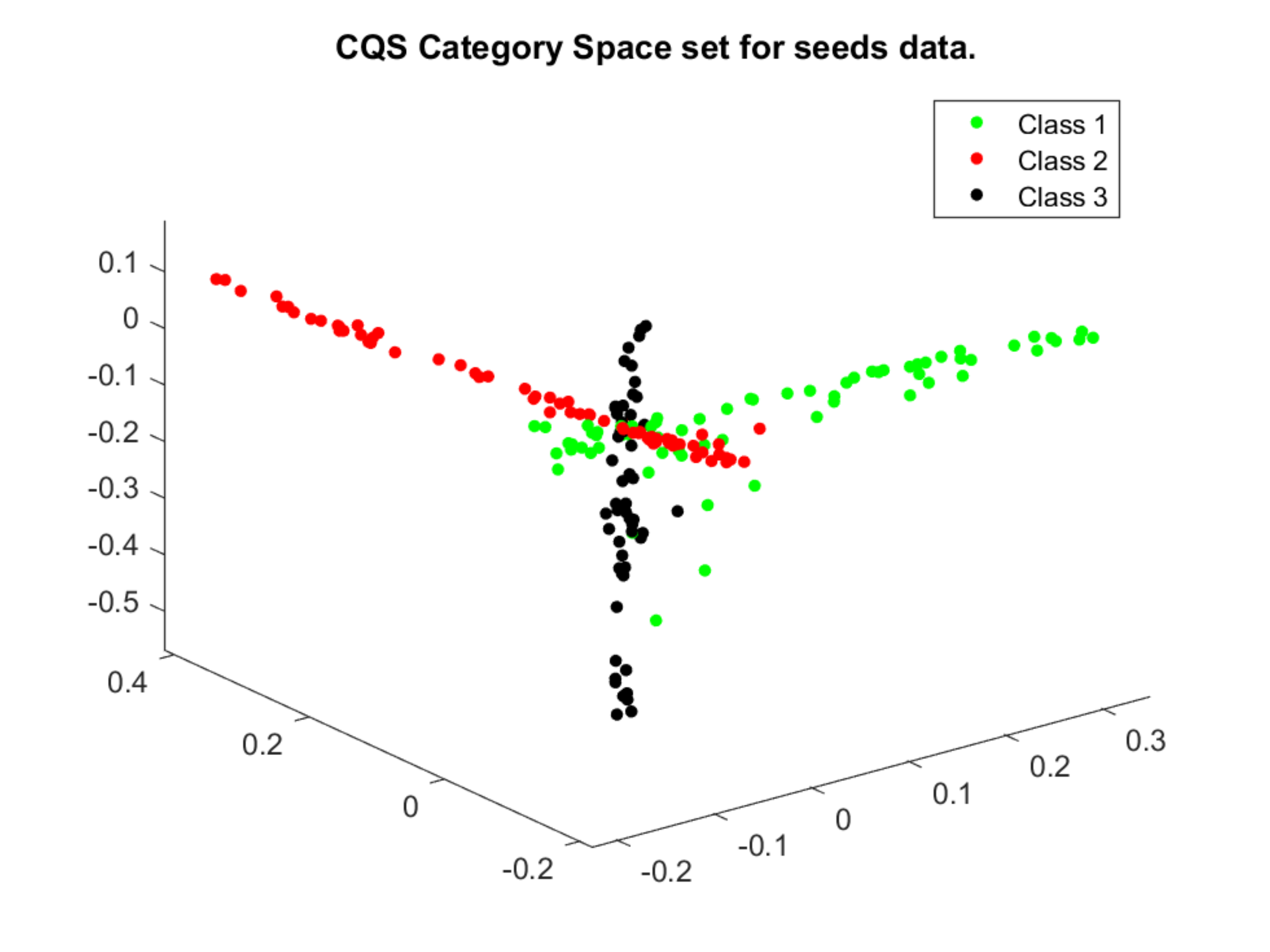} & \includegraphics[scale=0.15]{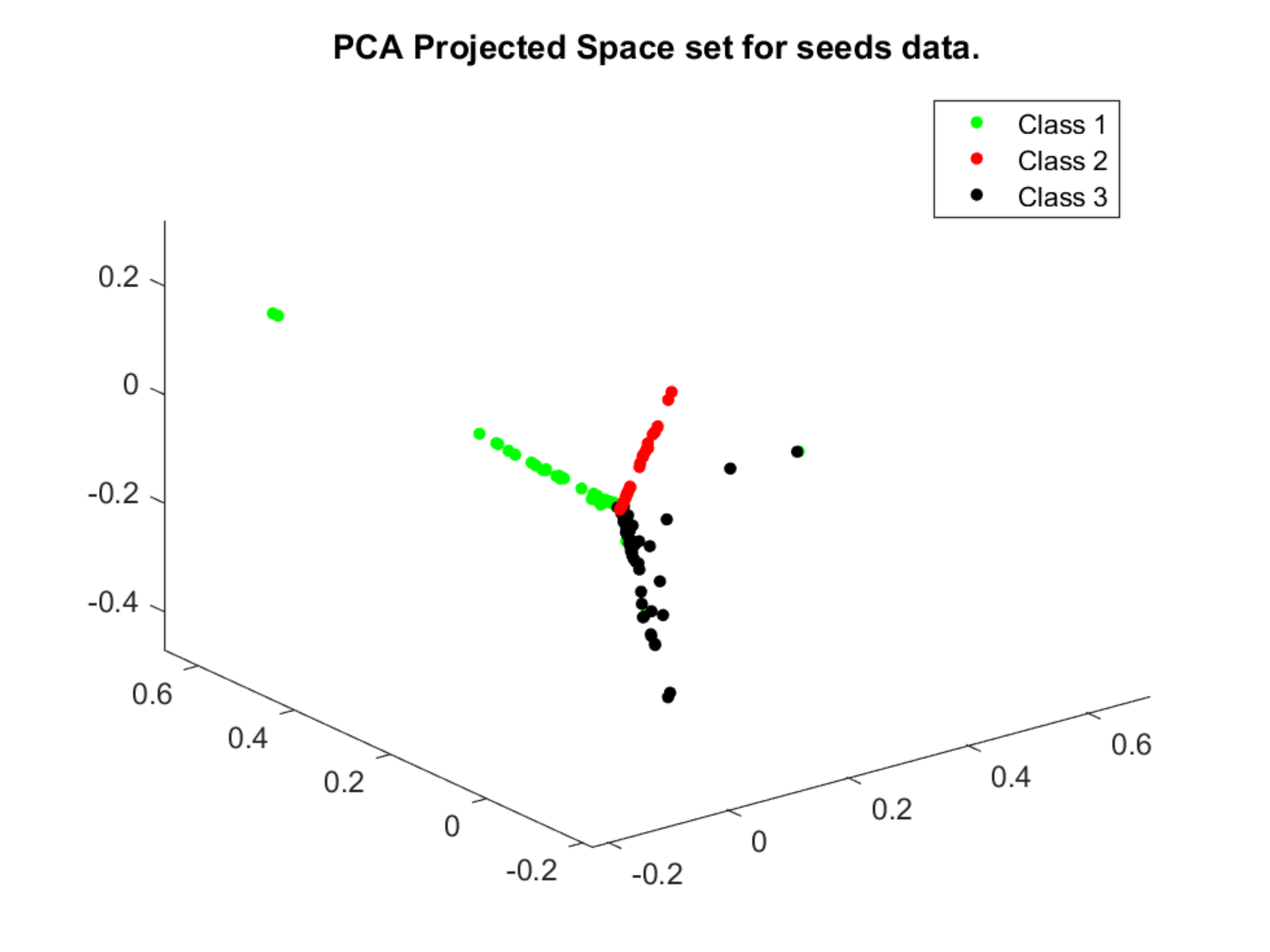} & \includegraphics[scale=0.15]{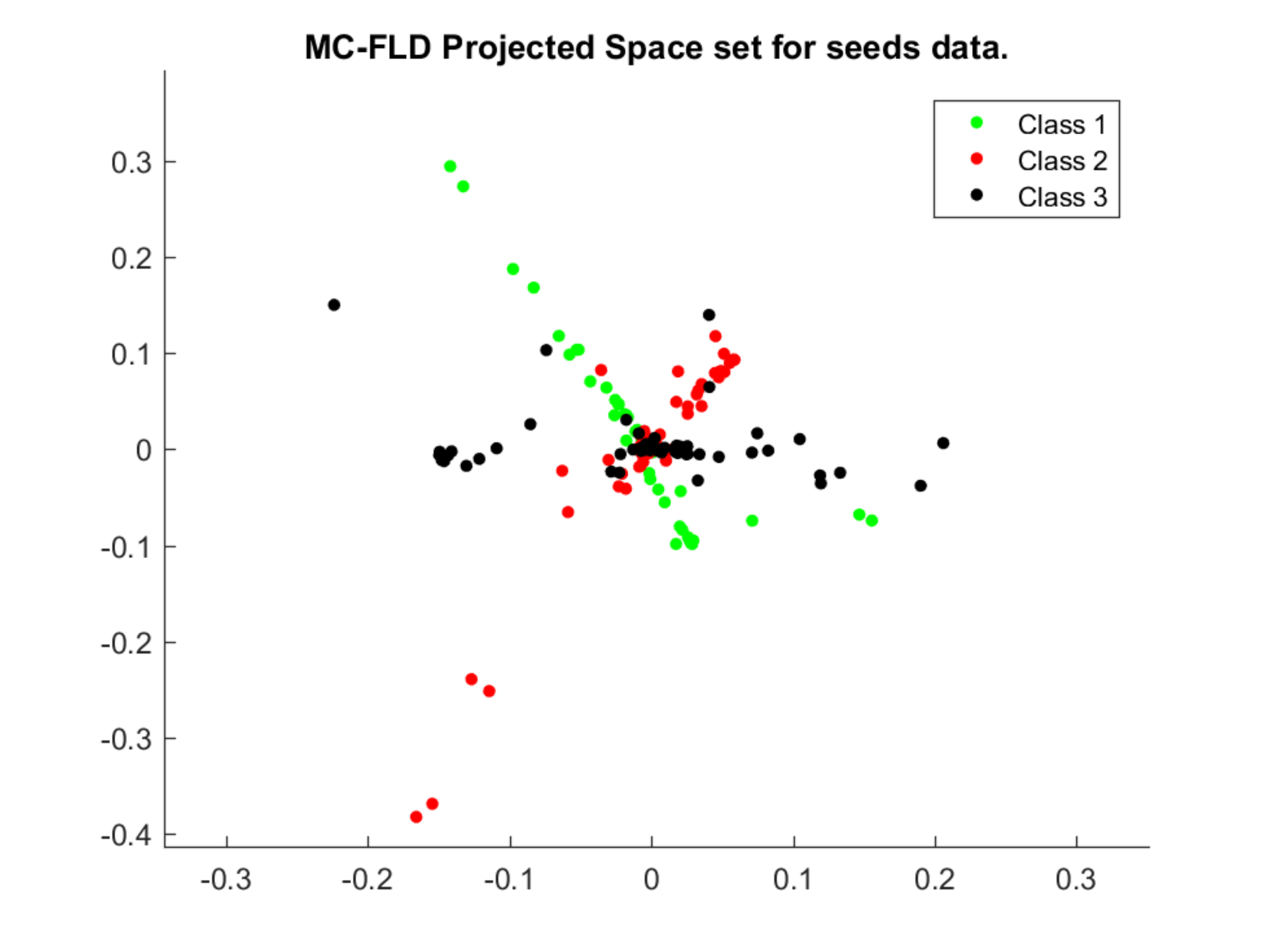}\tabularnewline
\hline 
\textbf{K-CQS} & \textbf{K-PCA} & \textbf{K-MC-FLD}\tabularnewline
\hline 
\end{tabular}
\par\end{centering}
\caption{Reduced dimensionality projection for a medium $\sigma$ value: From
top to bottom: Vertebral, Thyroid, Wine, Iris, Seeds.\label{fig:category-space-reduced}}
\end{figure}
\par\end{center}

\begin{center}
\begin{figure}[H]
\begin{centering}
\begin{tabular}{|c|c|c|}
\hline 
\includegraphics[scale=0.15]{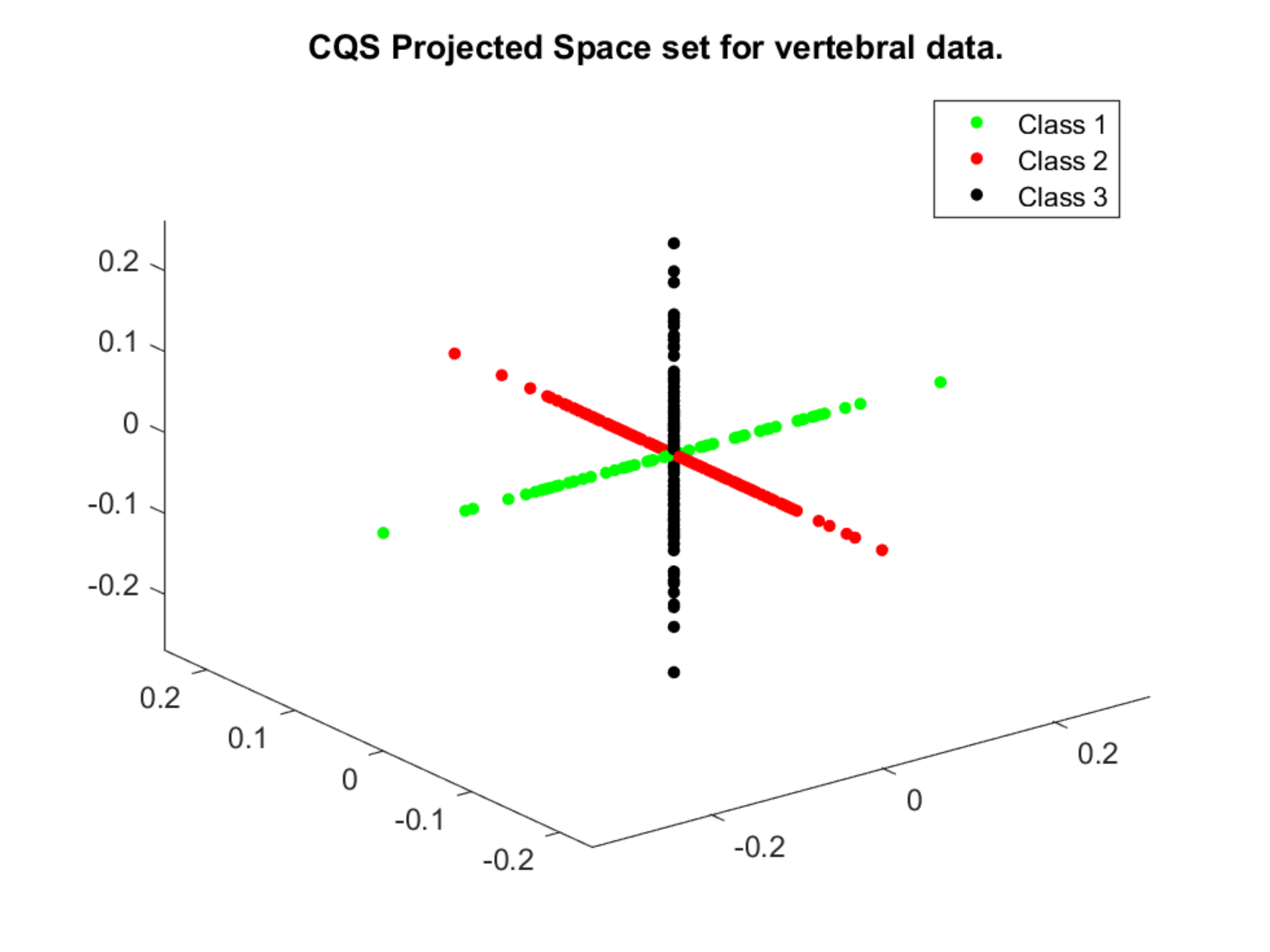} & \includegraphics[scale=0.15]{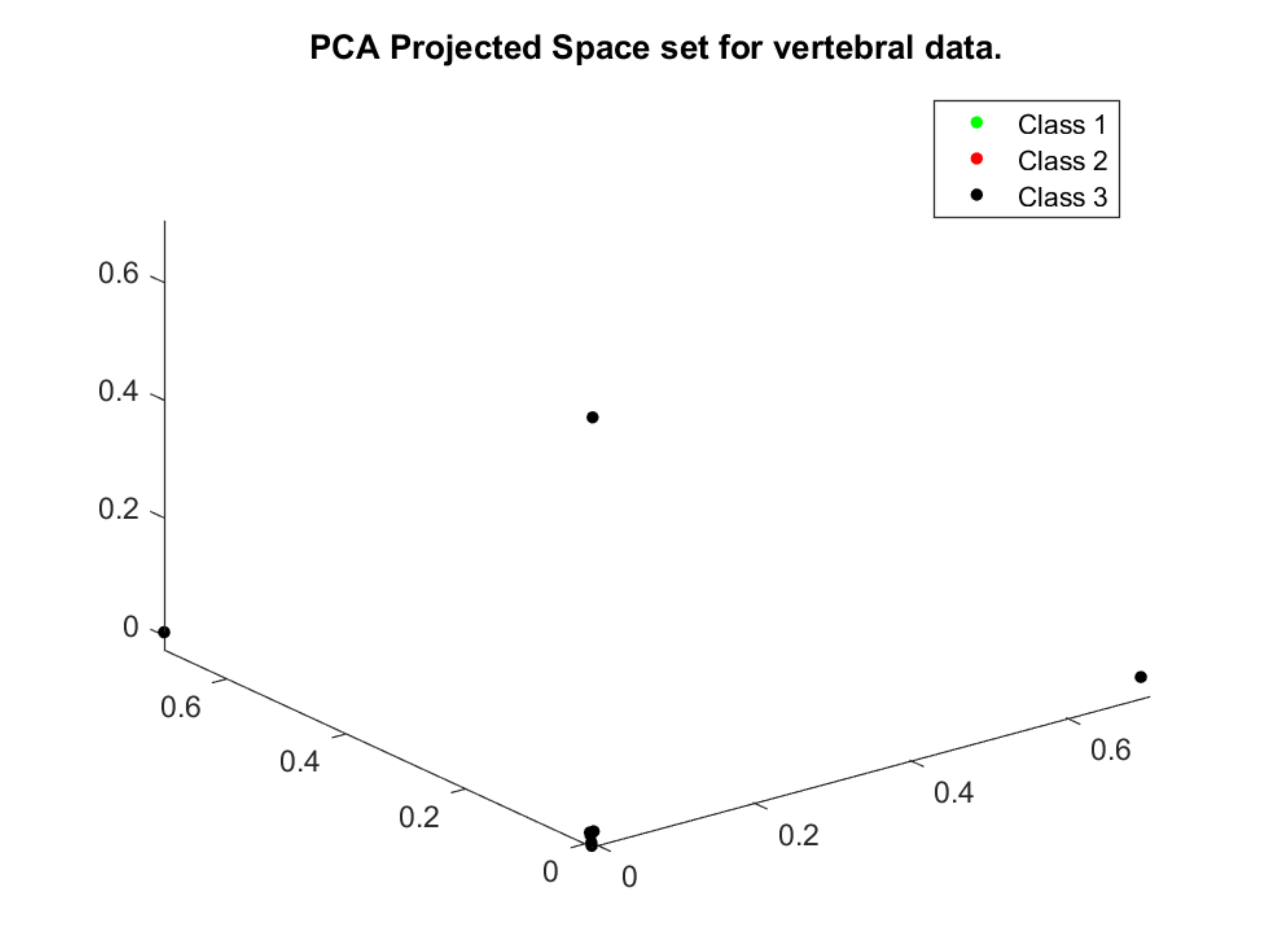} & \includegraphics[scale=0.15]{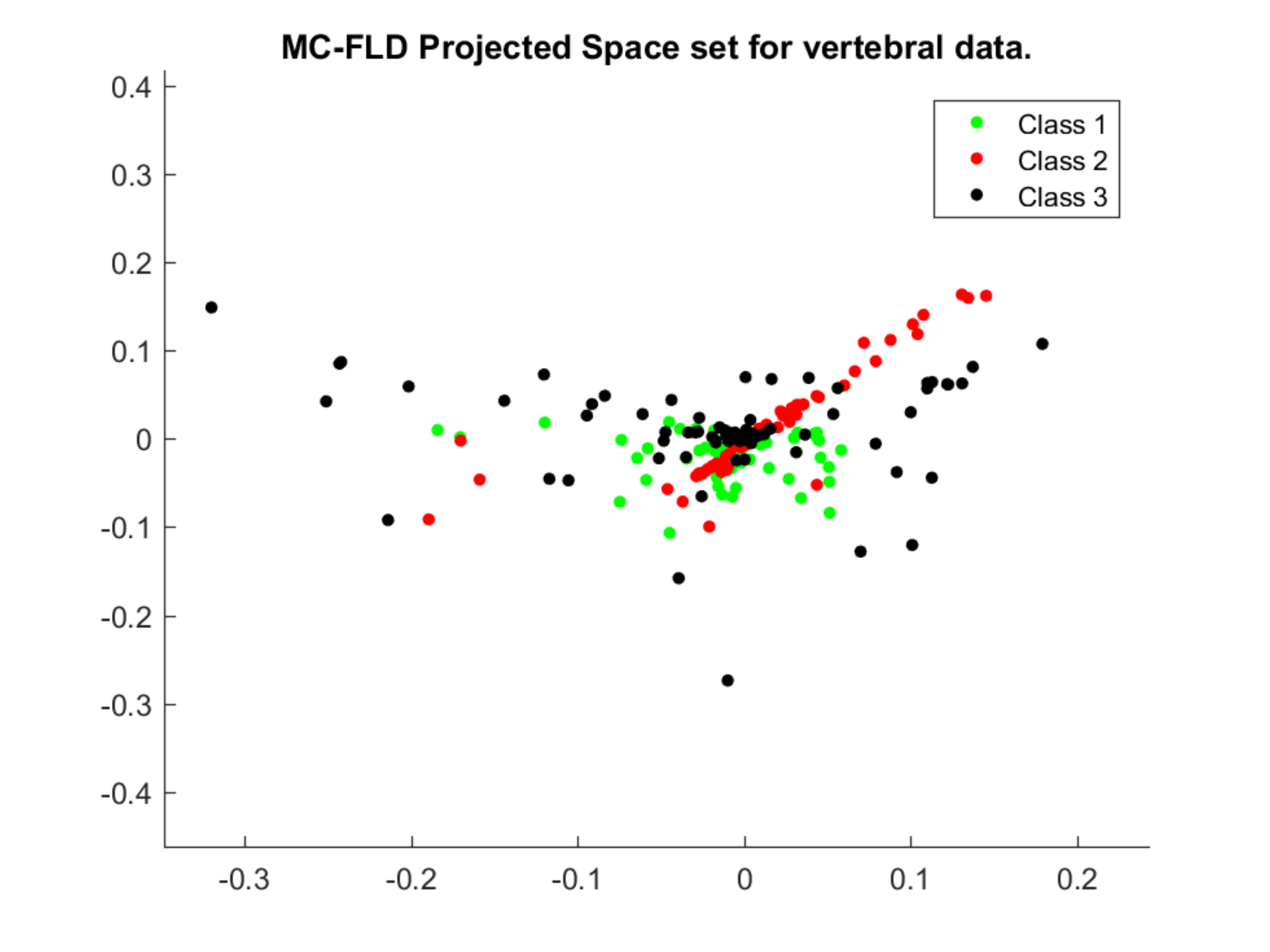}\tabularnewline
\hline 
\includegraphics[scale=0.15]{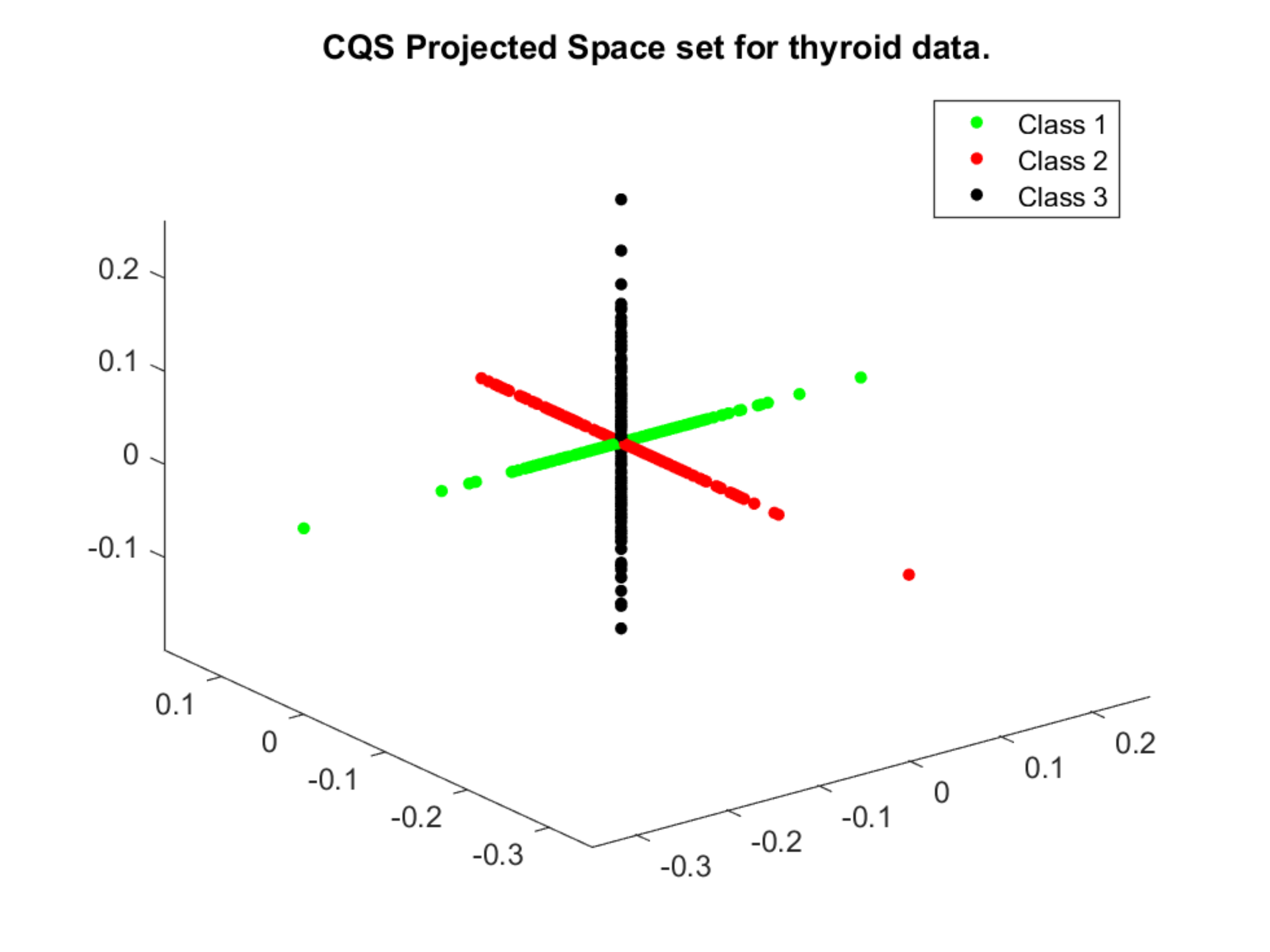} & \includegraphics[scale=0.15]{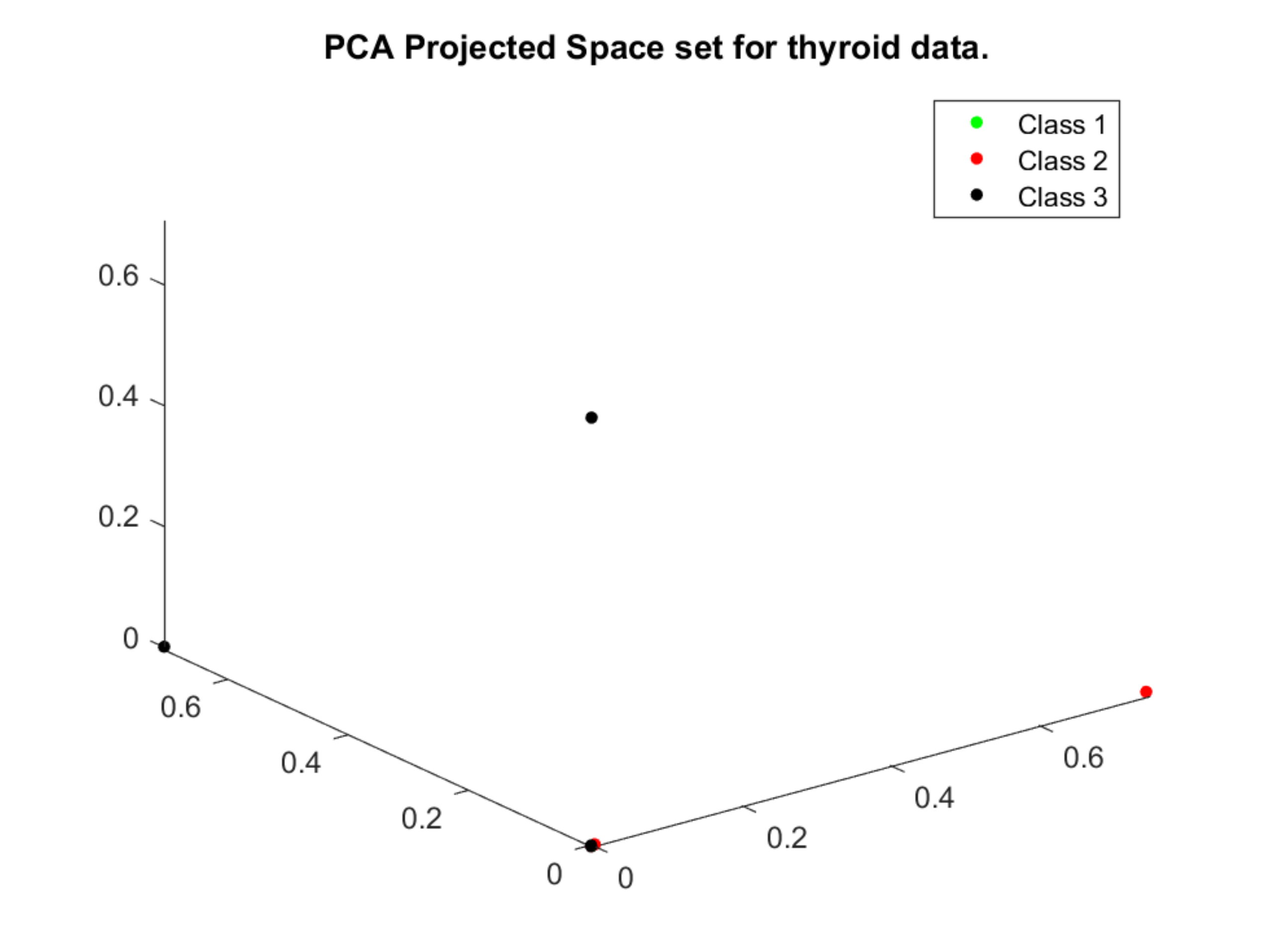} & \includegraphics[scale=0.15]{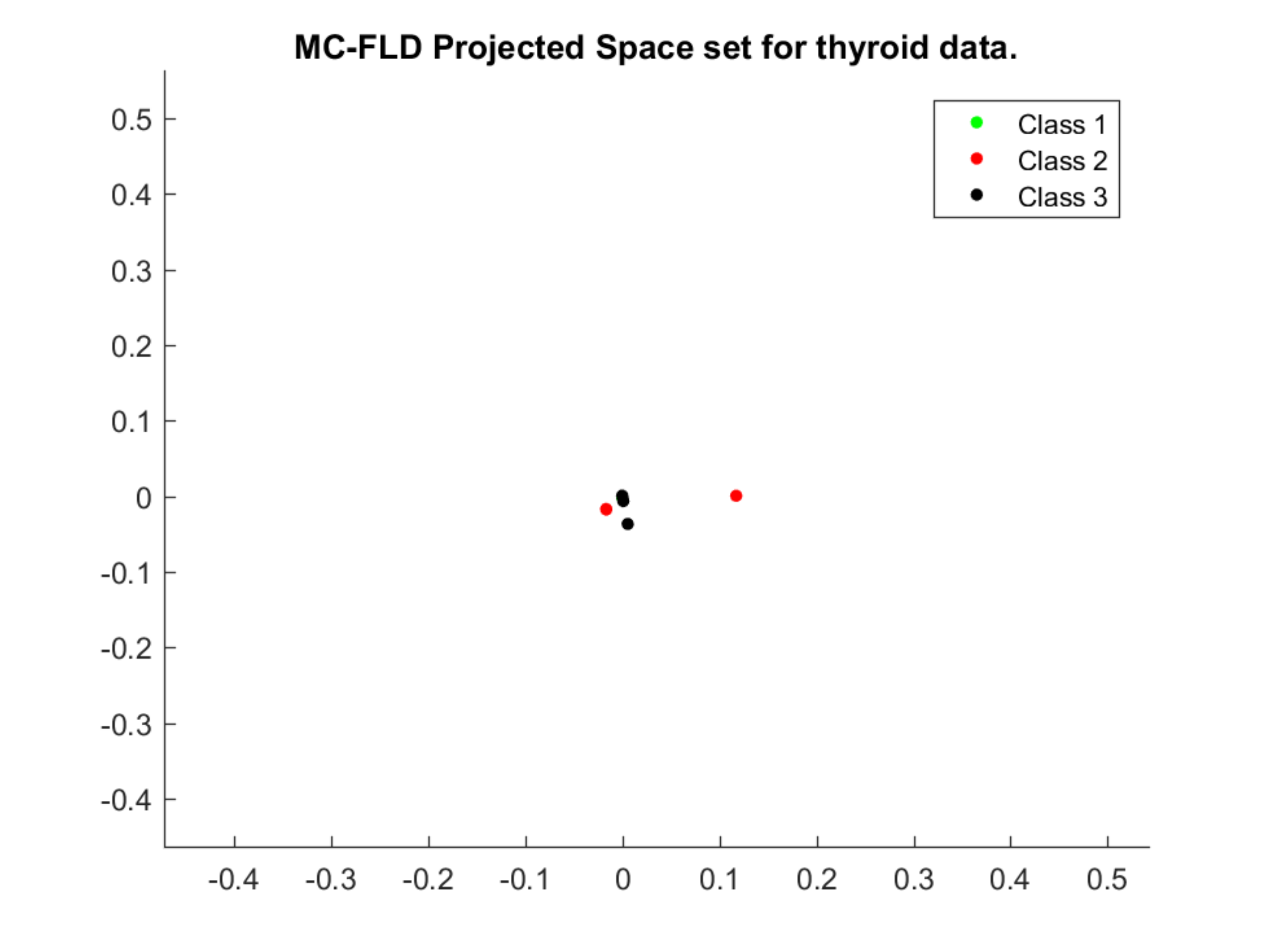}\tabularnewline
\hline 
\includegraphics[scale=0.15]{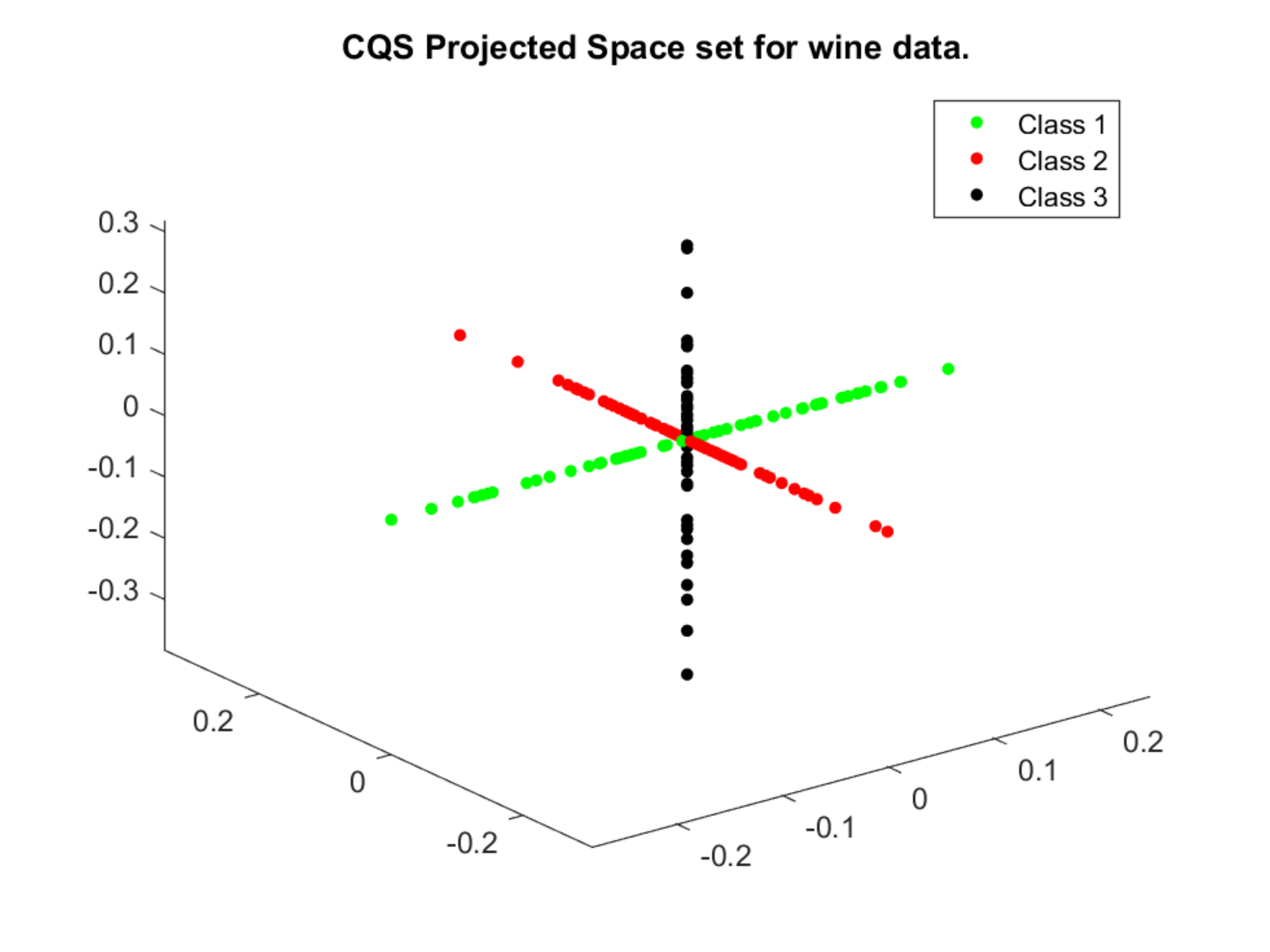} & \includegraphics[scale=0.15]{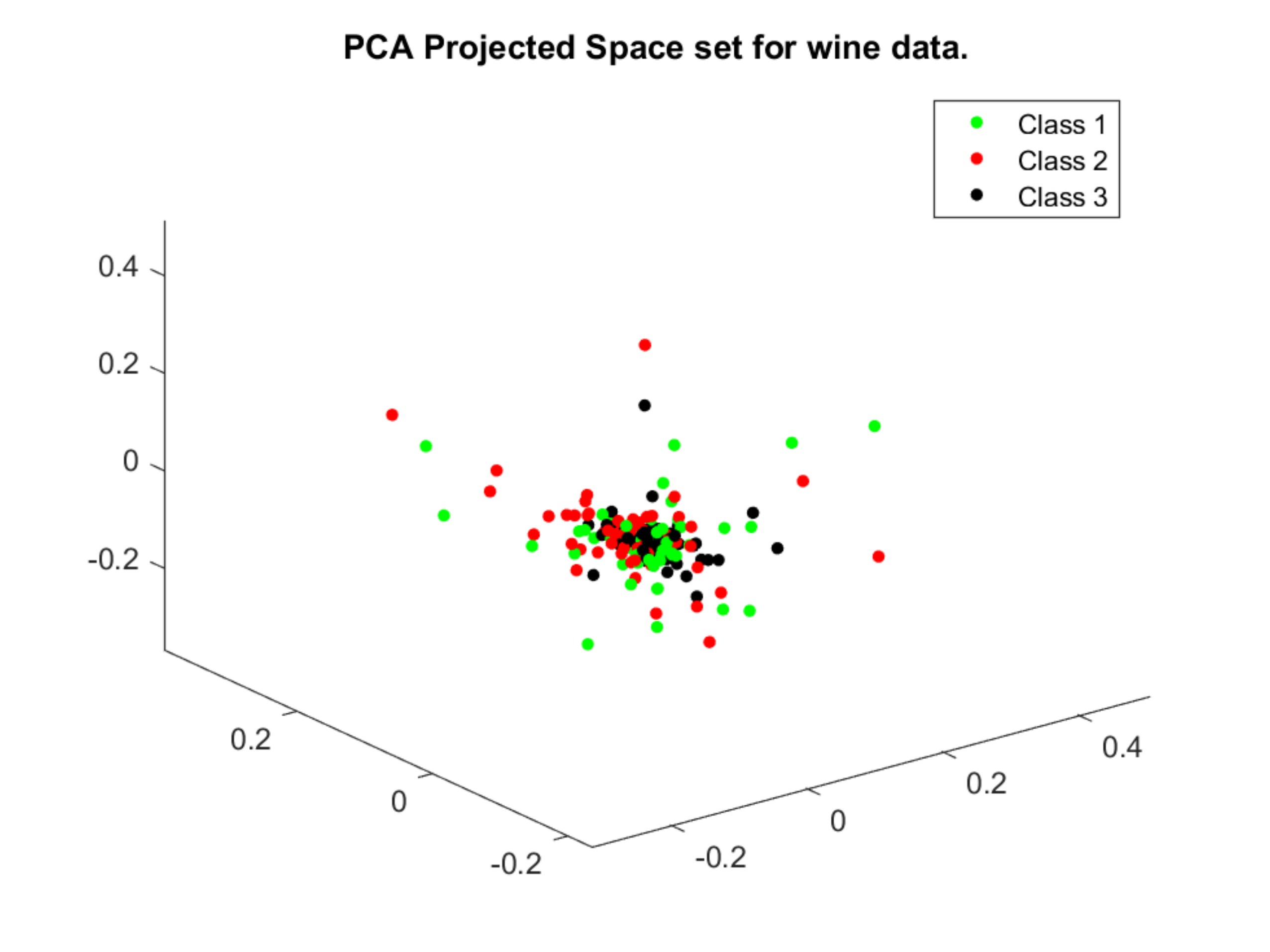} & \includegraphics[scale=0.15]{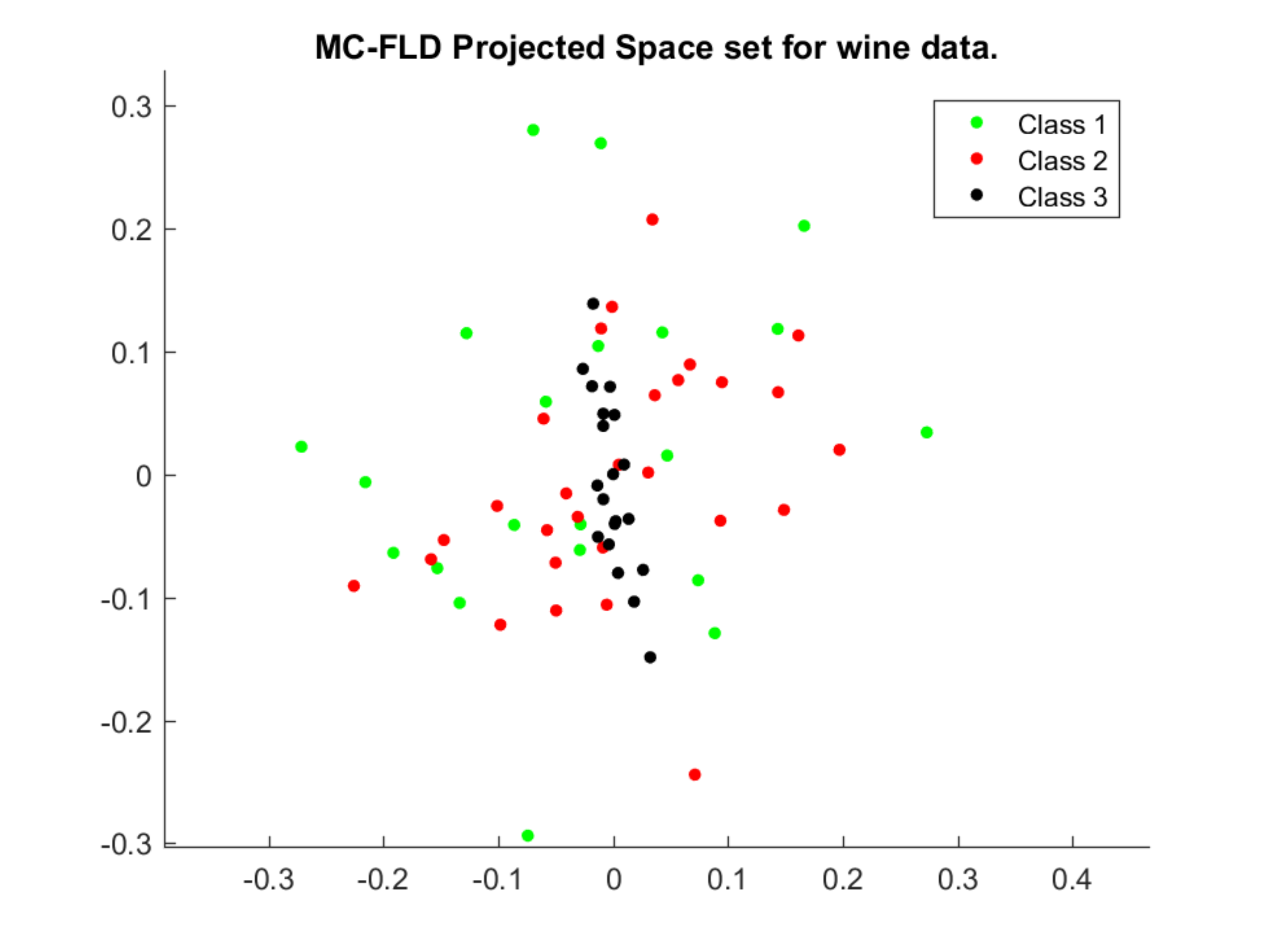}\tabularnewline
\hline 
\includegraphics[scale=0.15]{sm-images/all-wine} & \includegraphics[scale=0.15]{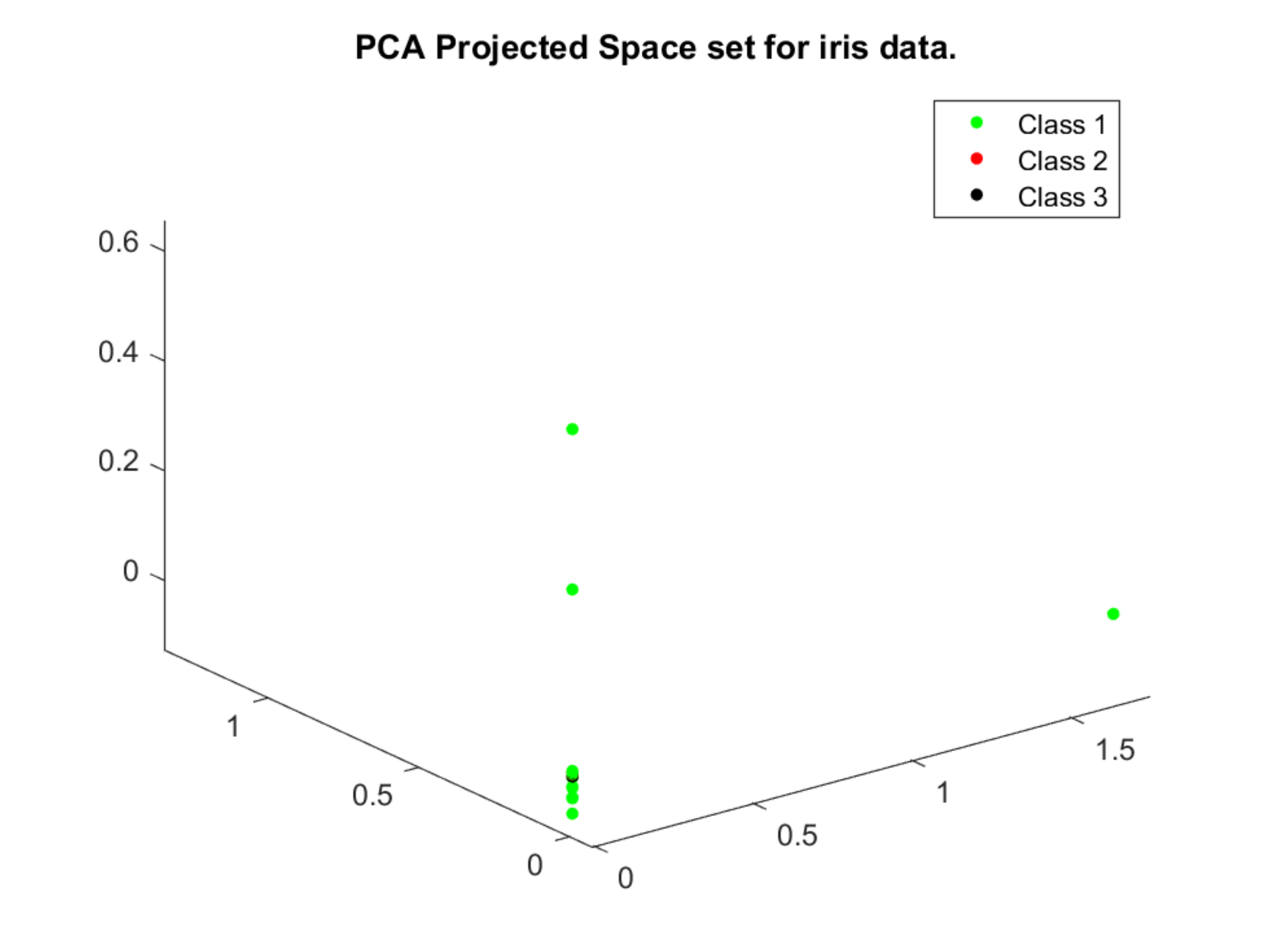} & \includegraphics[scale=0.15]{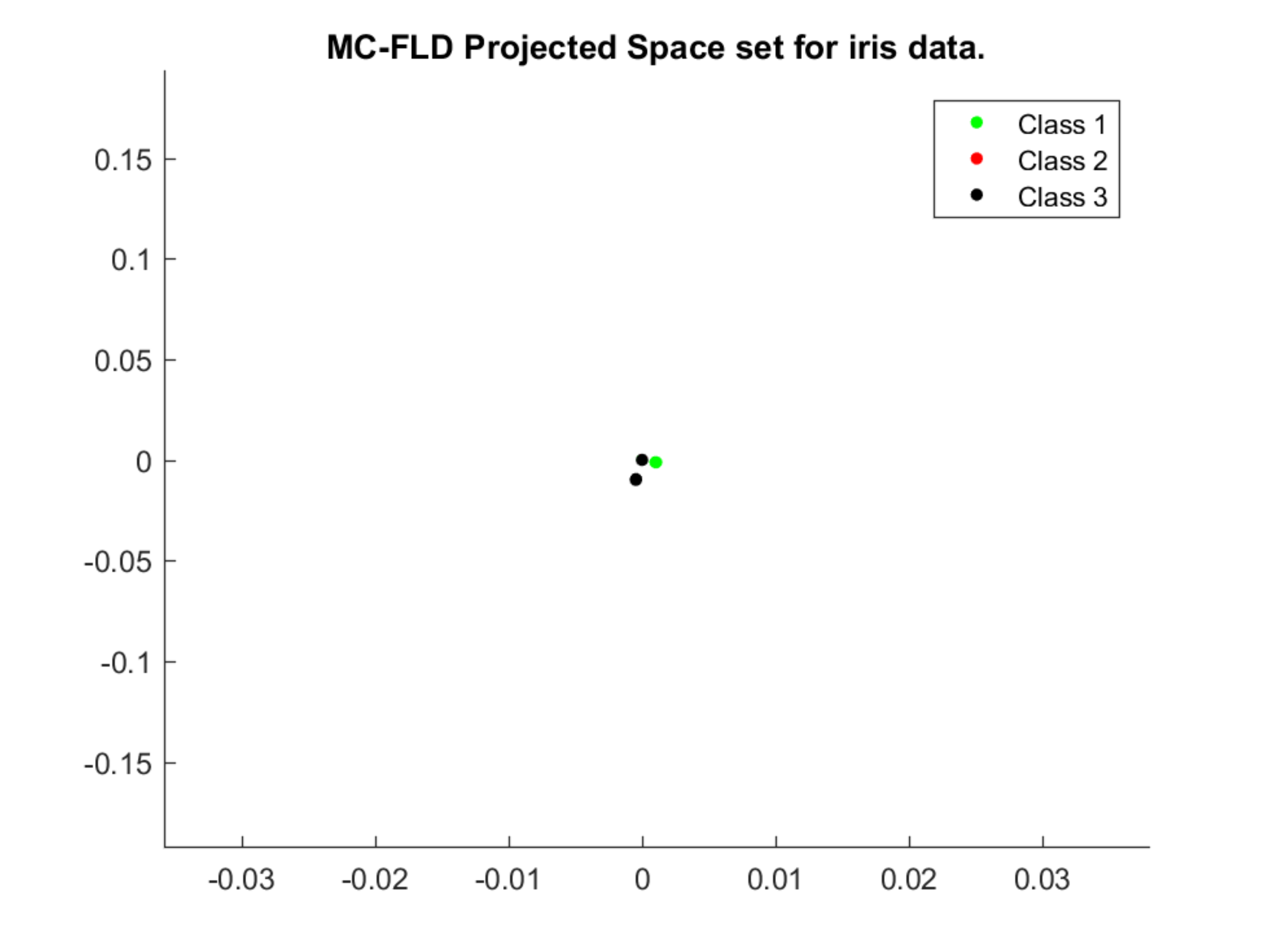}\tabularnewline
\hline 
\includegraphics[scale=0.15]{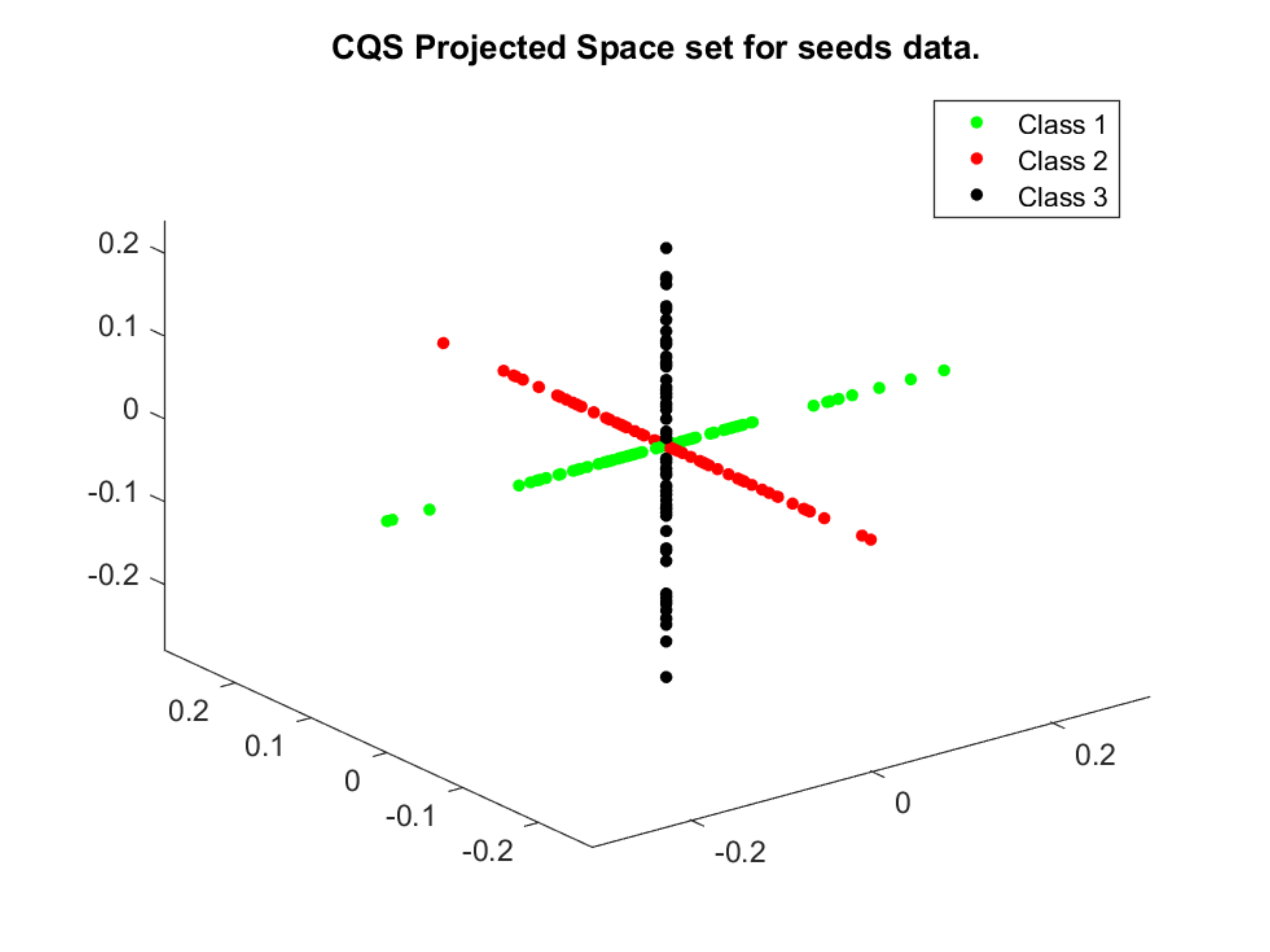} & \includegraphics[scale=0.15]{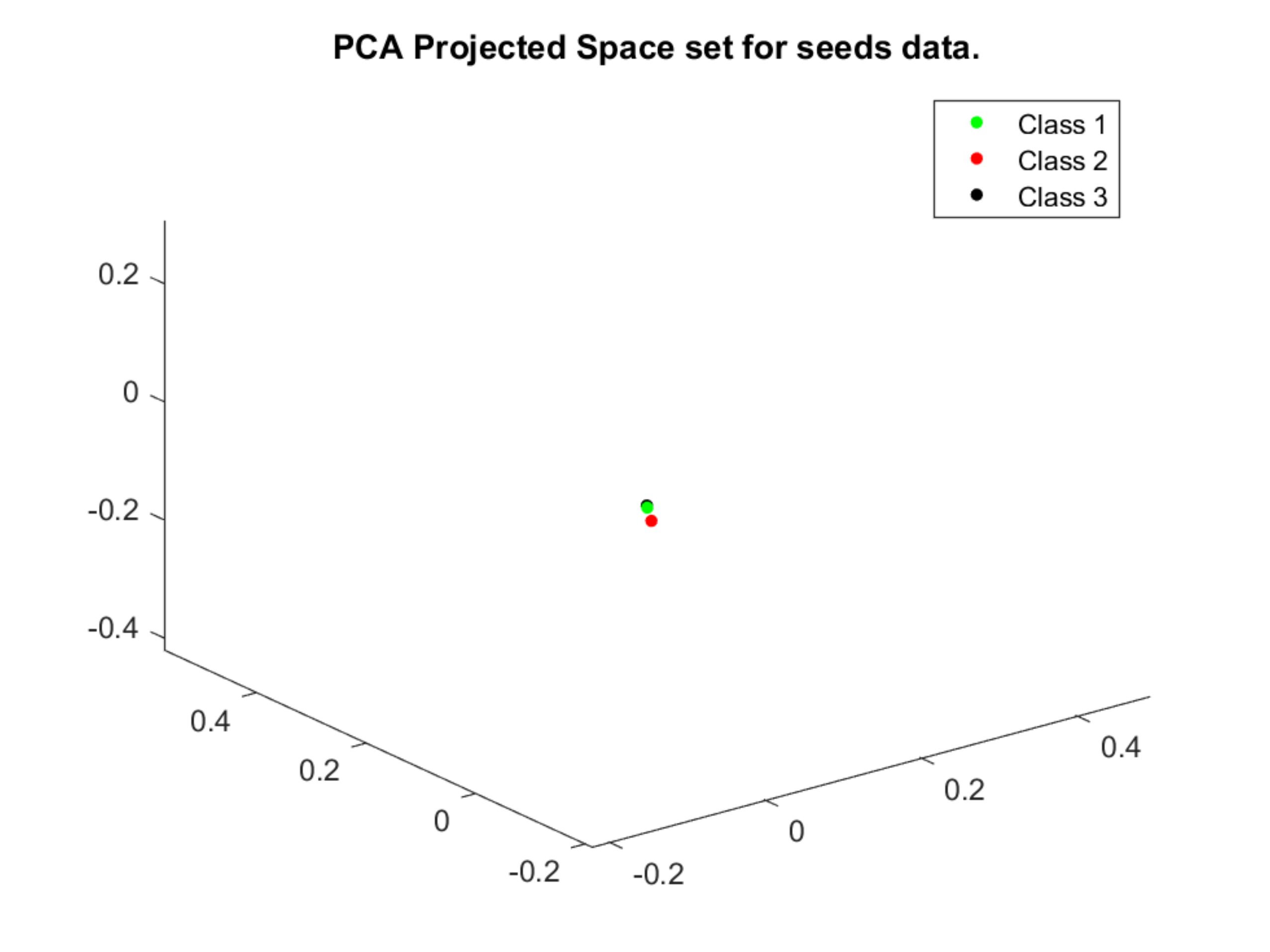} & \includegraphics[scale=0.15]{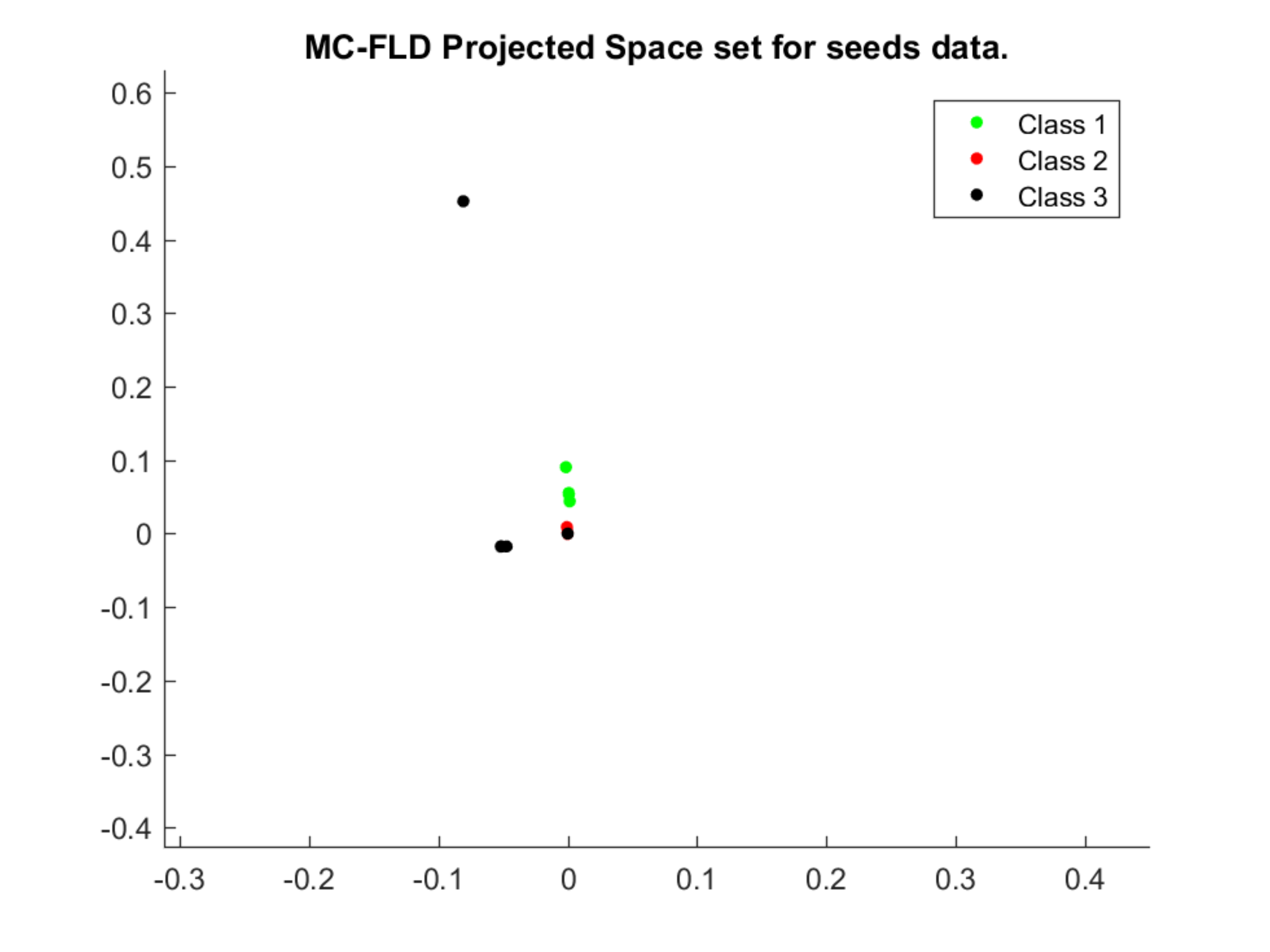}\tabularnewline
\hline 
\textbf{K-CQS} & \textbf{K-PCA} & \textbf{K-MC-FLD}\tabularnewline
\hline 
\end{tabular}
\par\end{centering}
\caption{Reduced dimensionality projection for a small $\sigma$ value. From
top to bottom: Vertebral, Thyroid, Wine, Iris, Seeds.\label{fig:category-space-reduced-md-sigma}}
\end{figure}
\par\end{center}

\begin{center}
\begin{figure}[H]
\begin{centering}
\begin{tabular}{|c|c|c|}
\hline 
\includegraphics[scale=0.15]{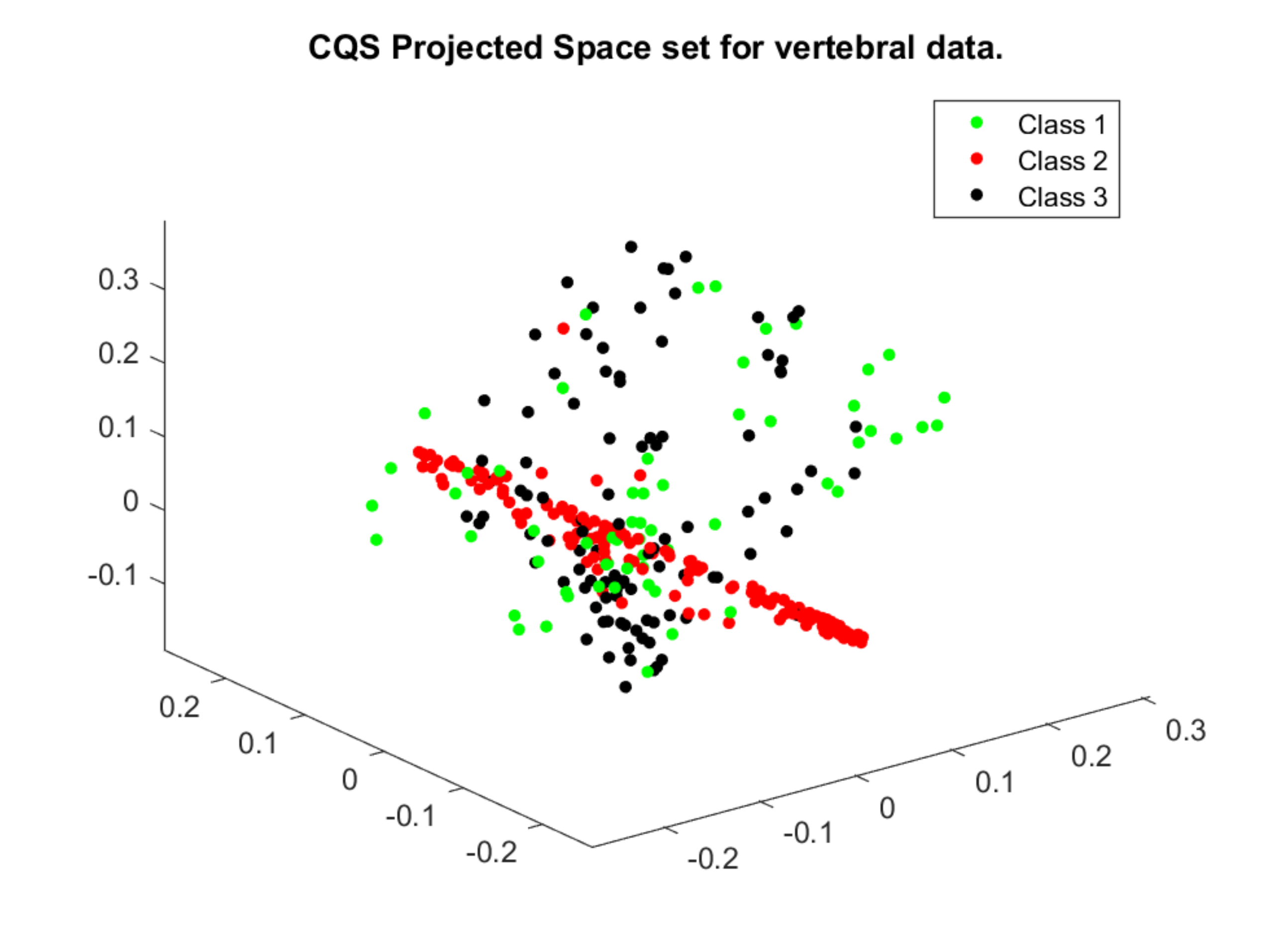} & \includegraphics[scale=0.15]{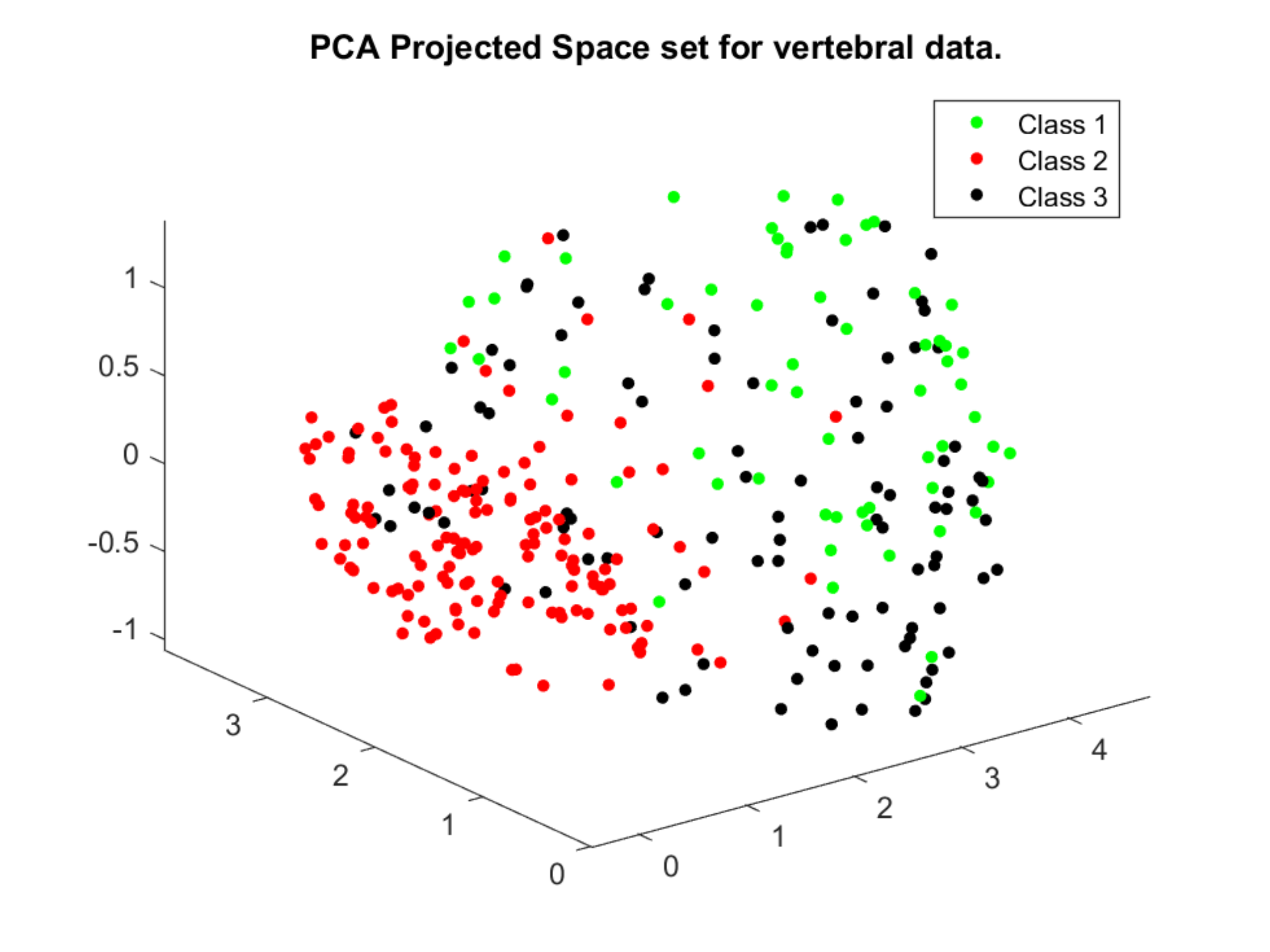} & \includegraphics[scale=0.15]{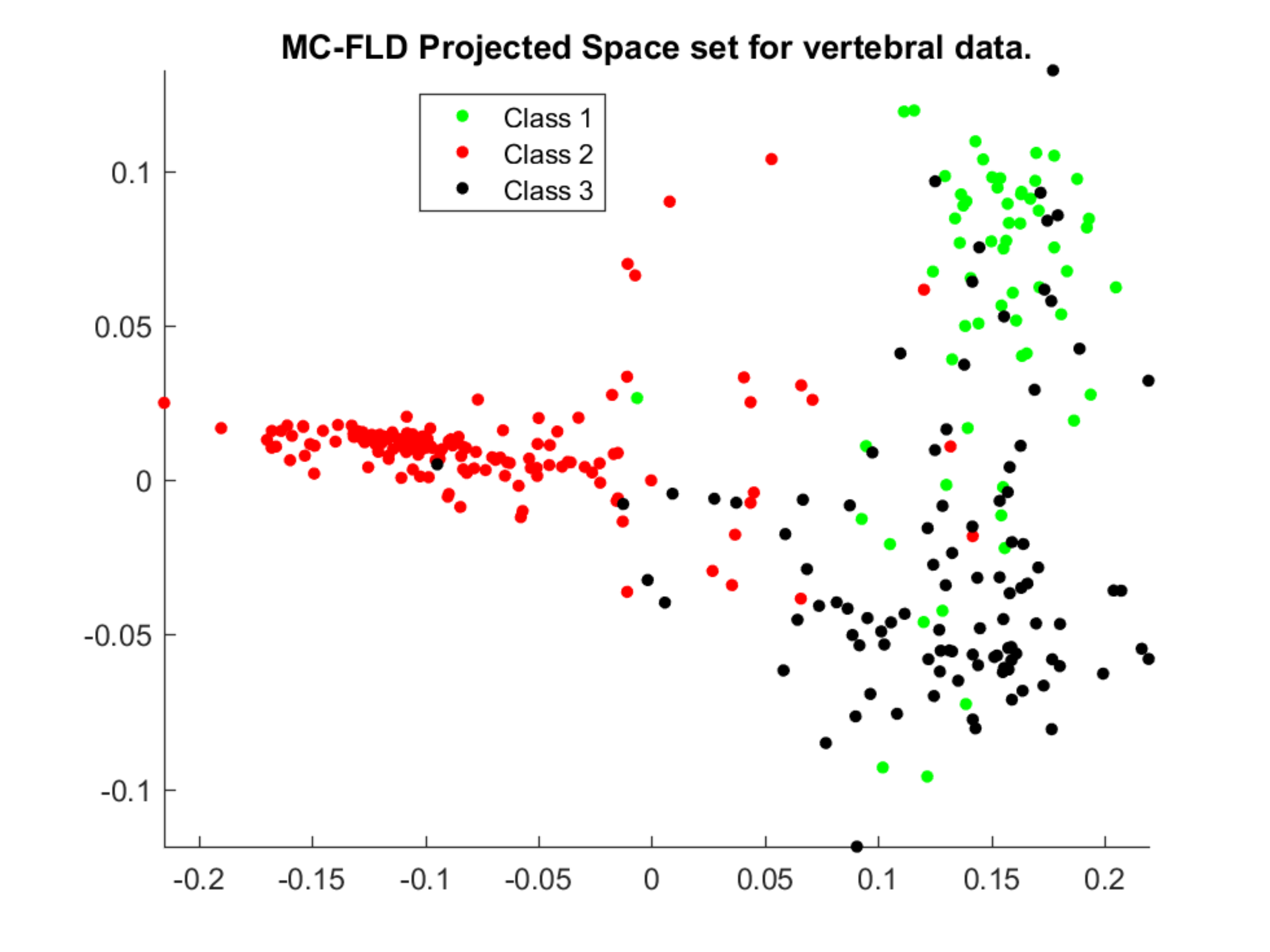}\tabularnewline
\hline 
\includegraphics[scale=0.15]{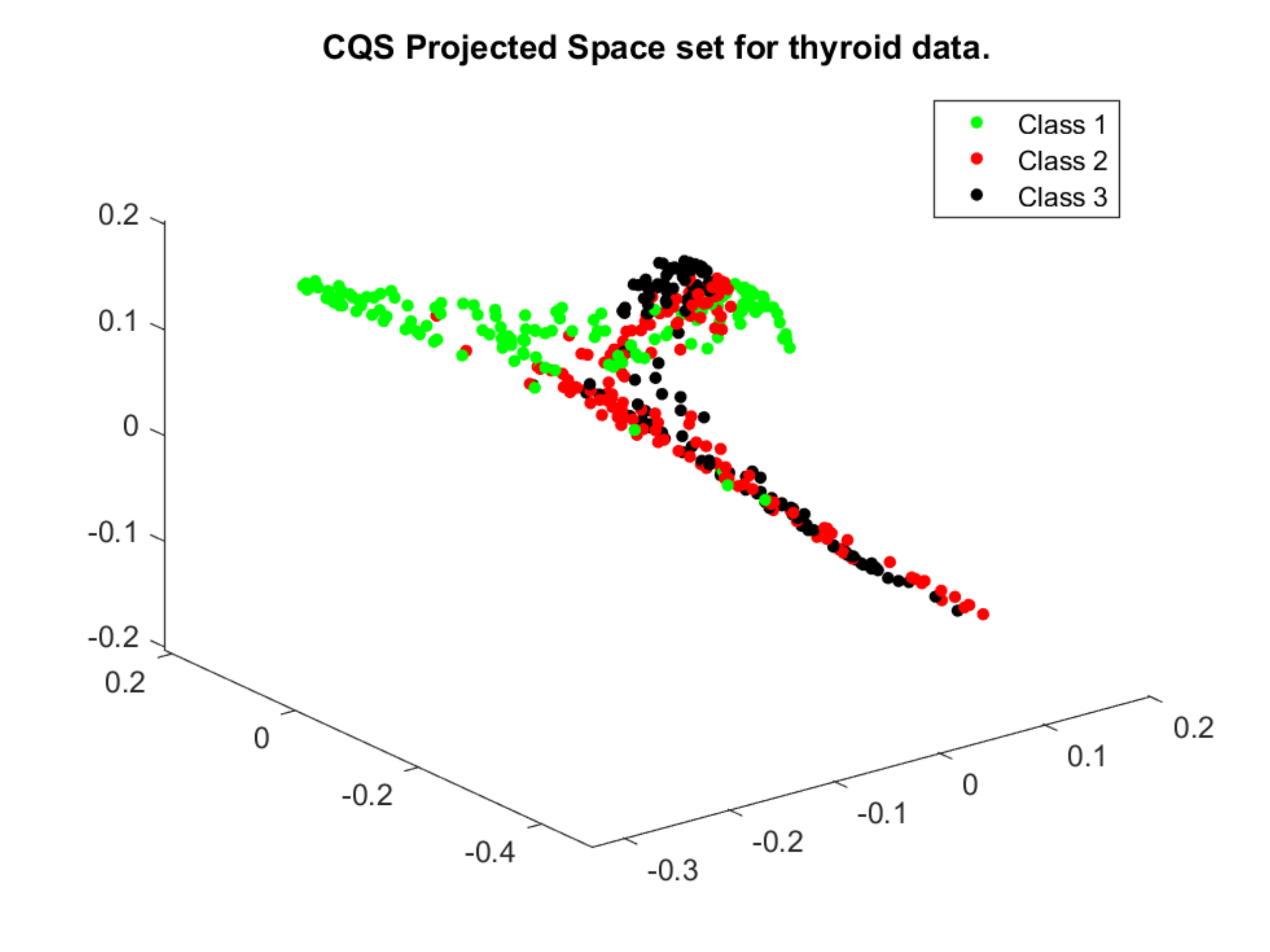} & \includegraphics[scale=0.15]{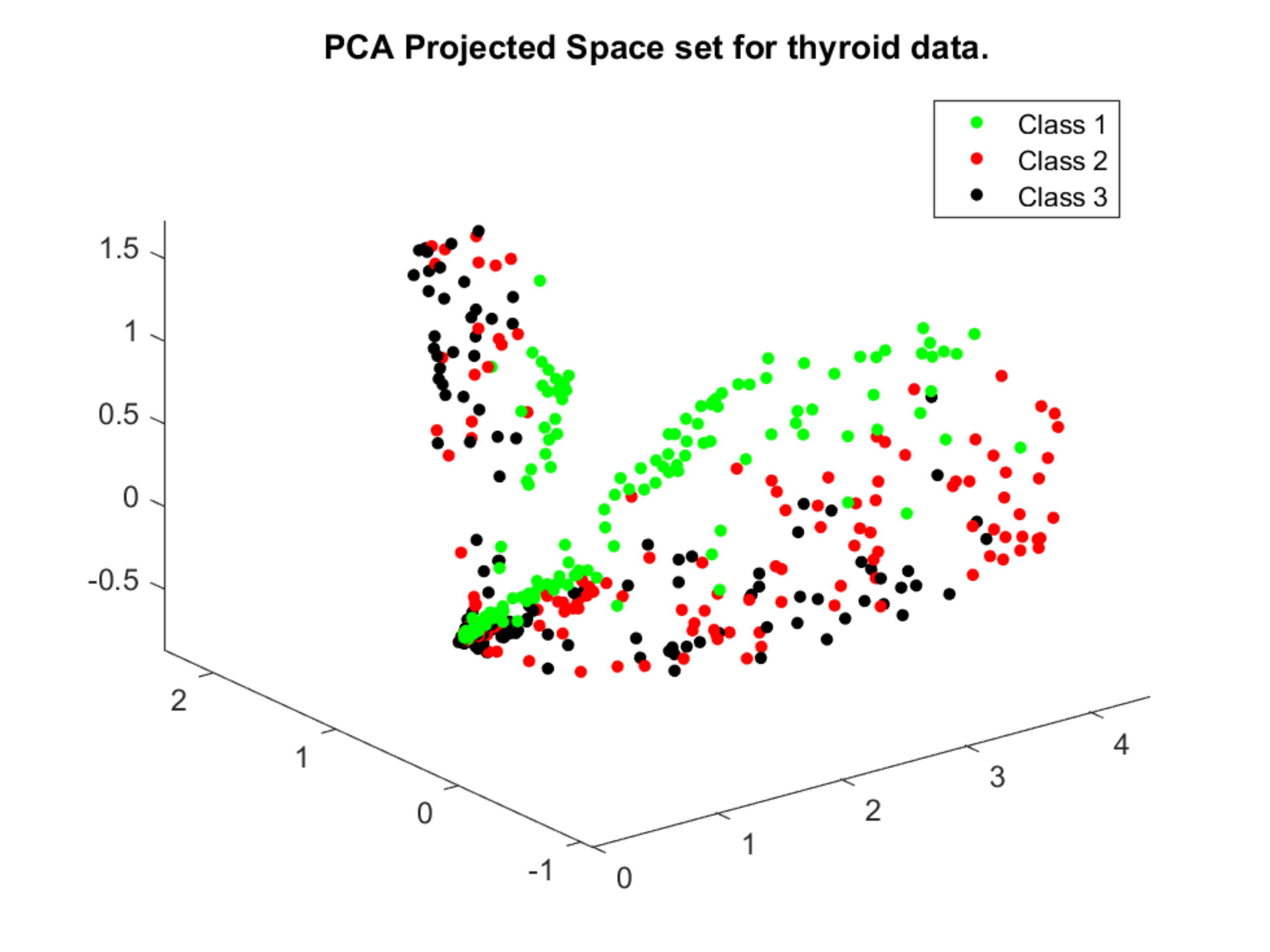} & \includegraphics[scale=0.15]{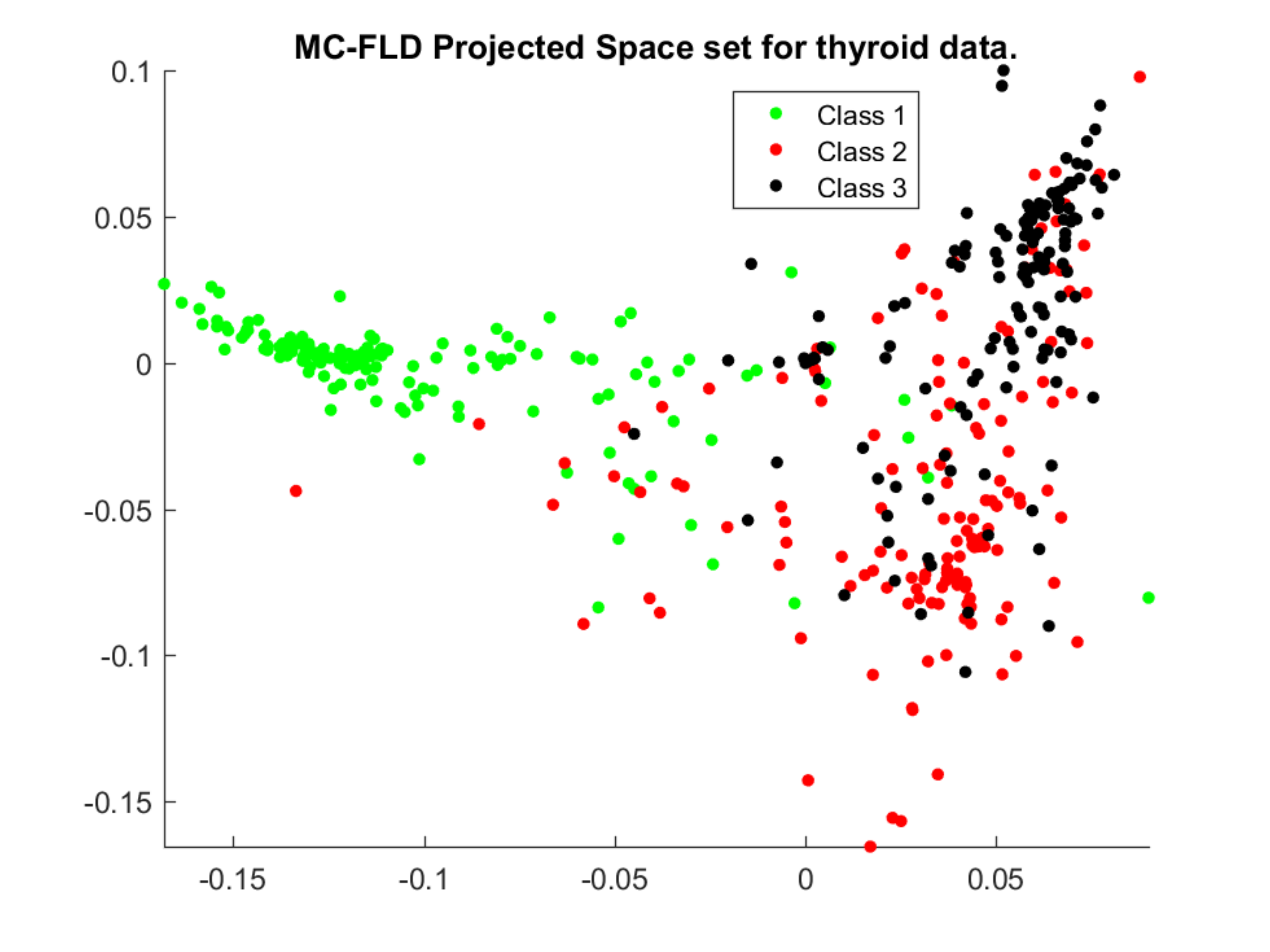}\tabularnewline
\hline 
\includegraphics[scale=0.15]{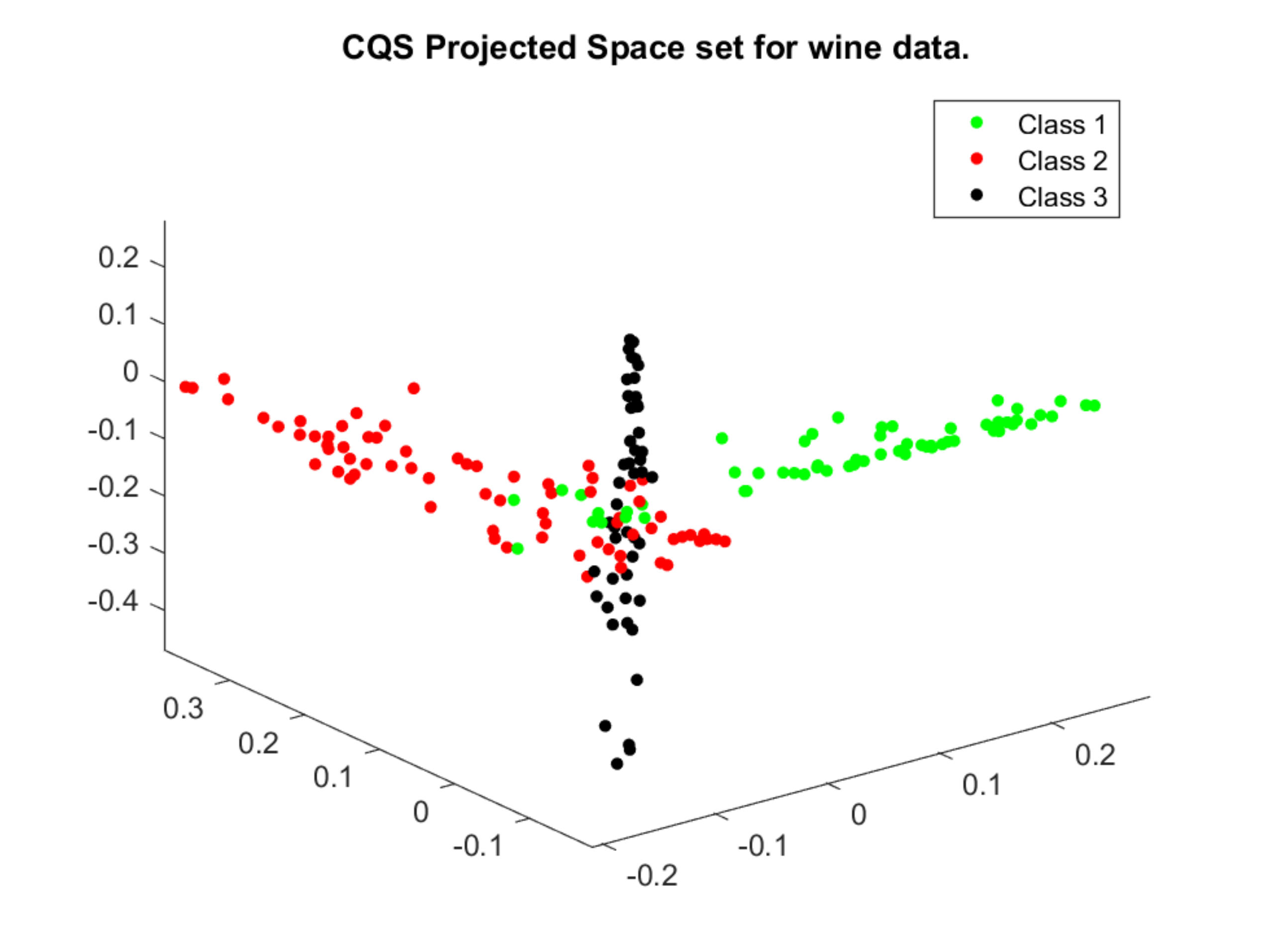} & \includegraphics[scale=0.15]{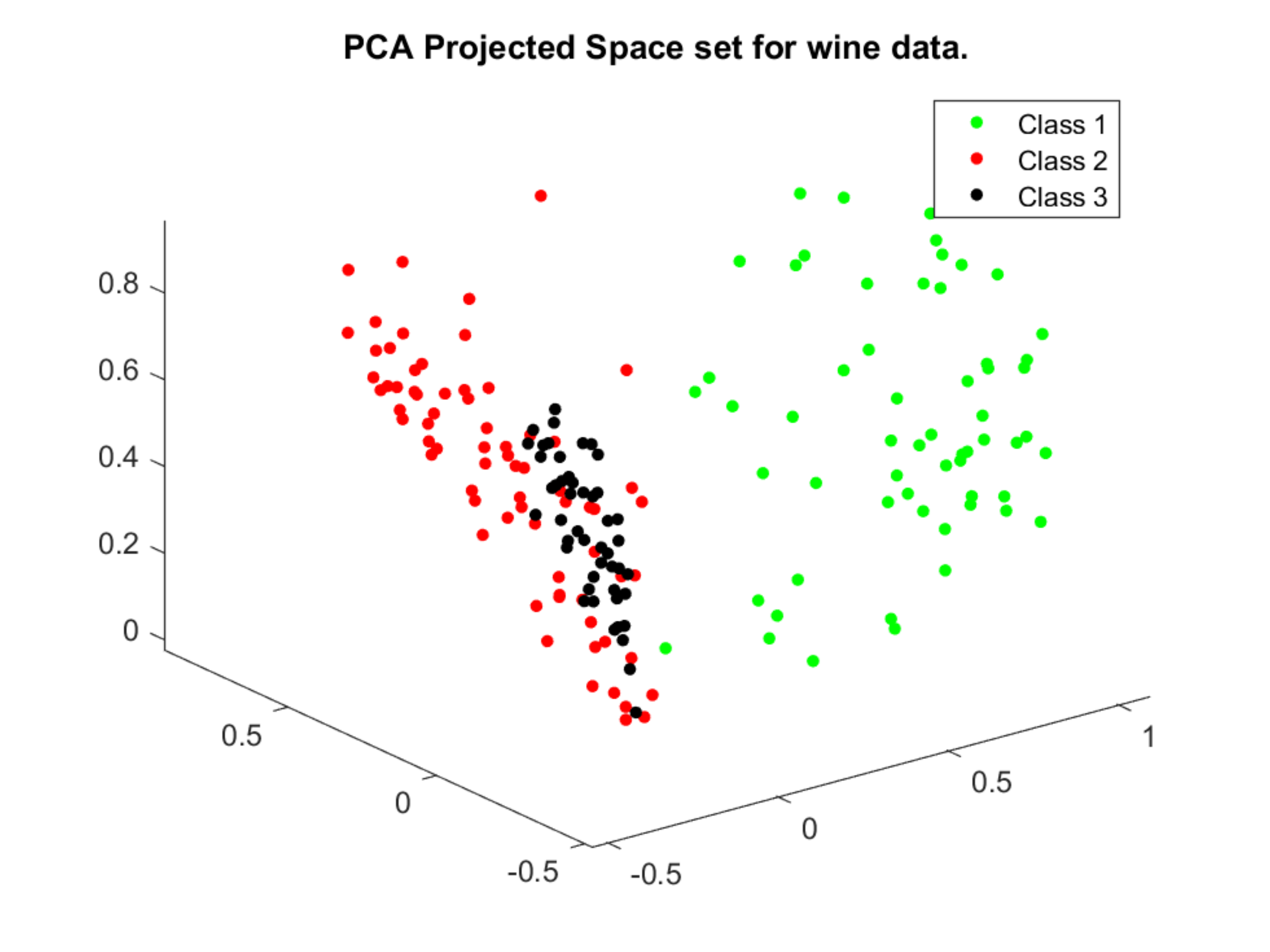} & \includegraphics[scale=0.15]{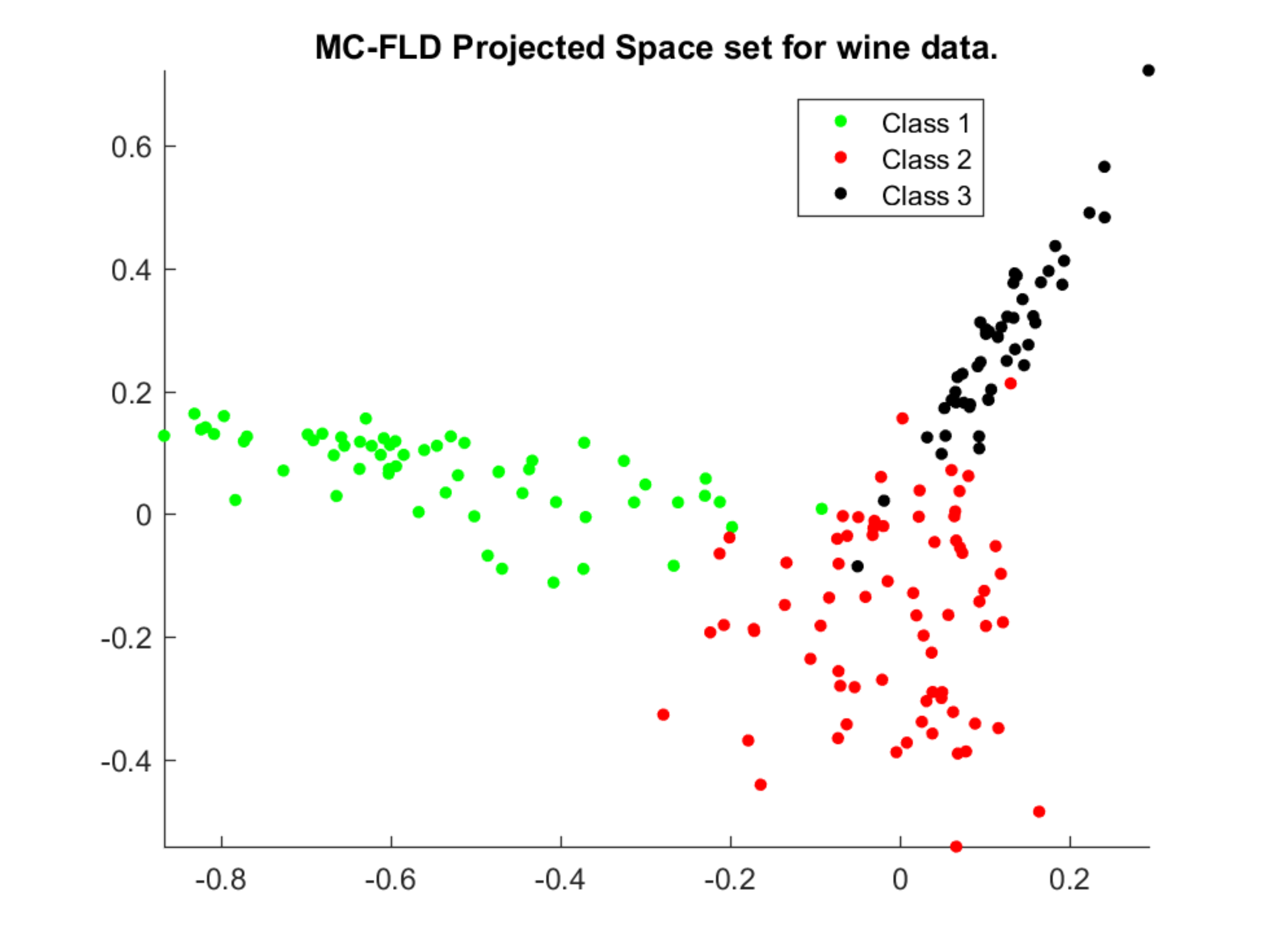}\tabularnewline
\hline 
\includegraphics[scale=0.15]{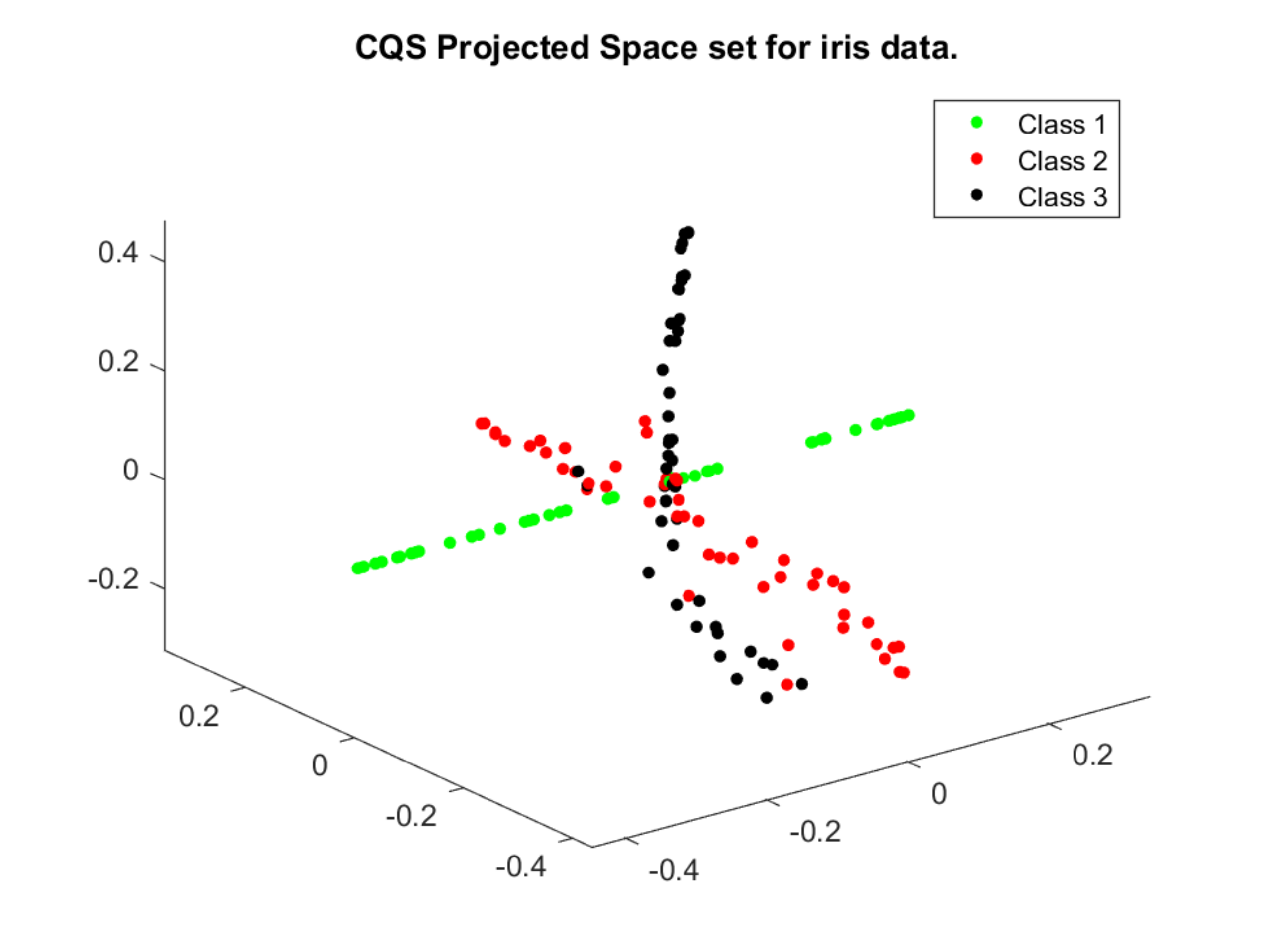} & \includegraphics[scale=0.15]{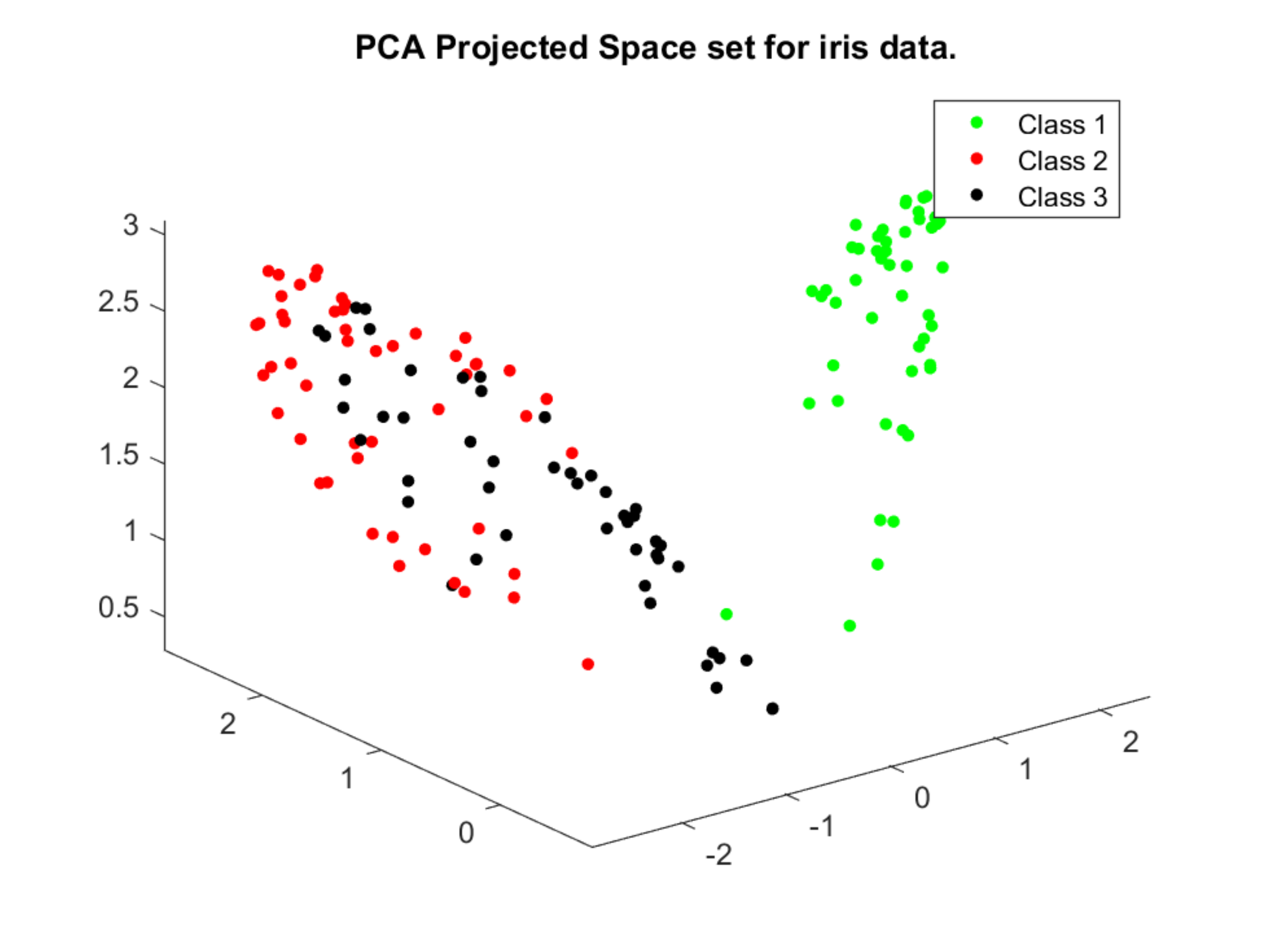} & \includegraphics[scale=0.15]{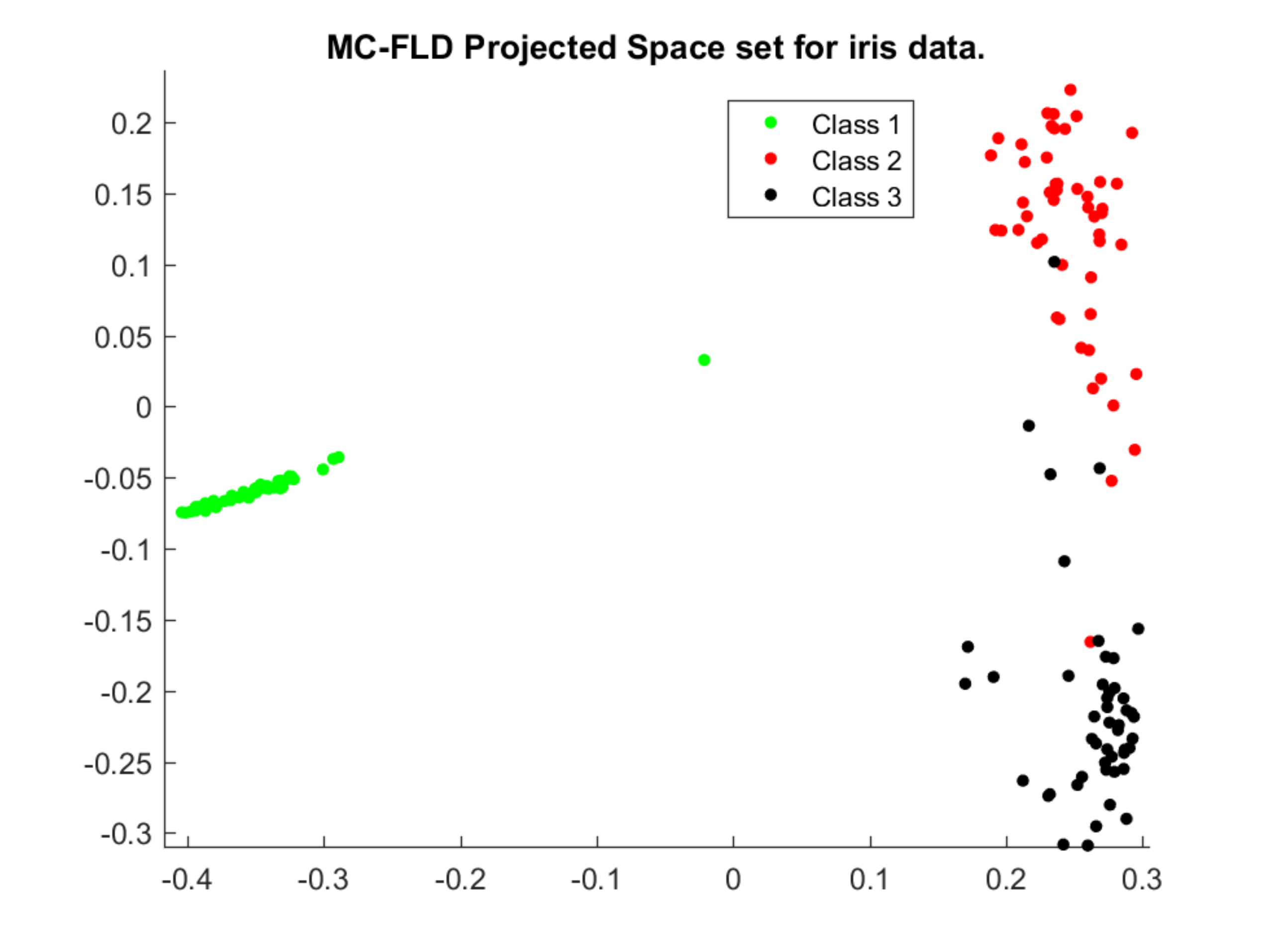}\tabularnewline
\hline 
\includegraphics[scale=0.15]{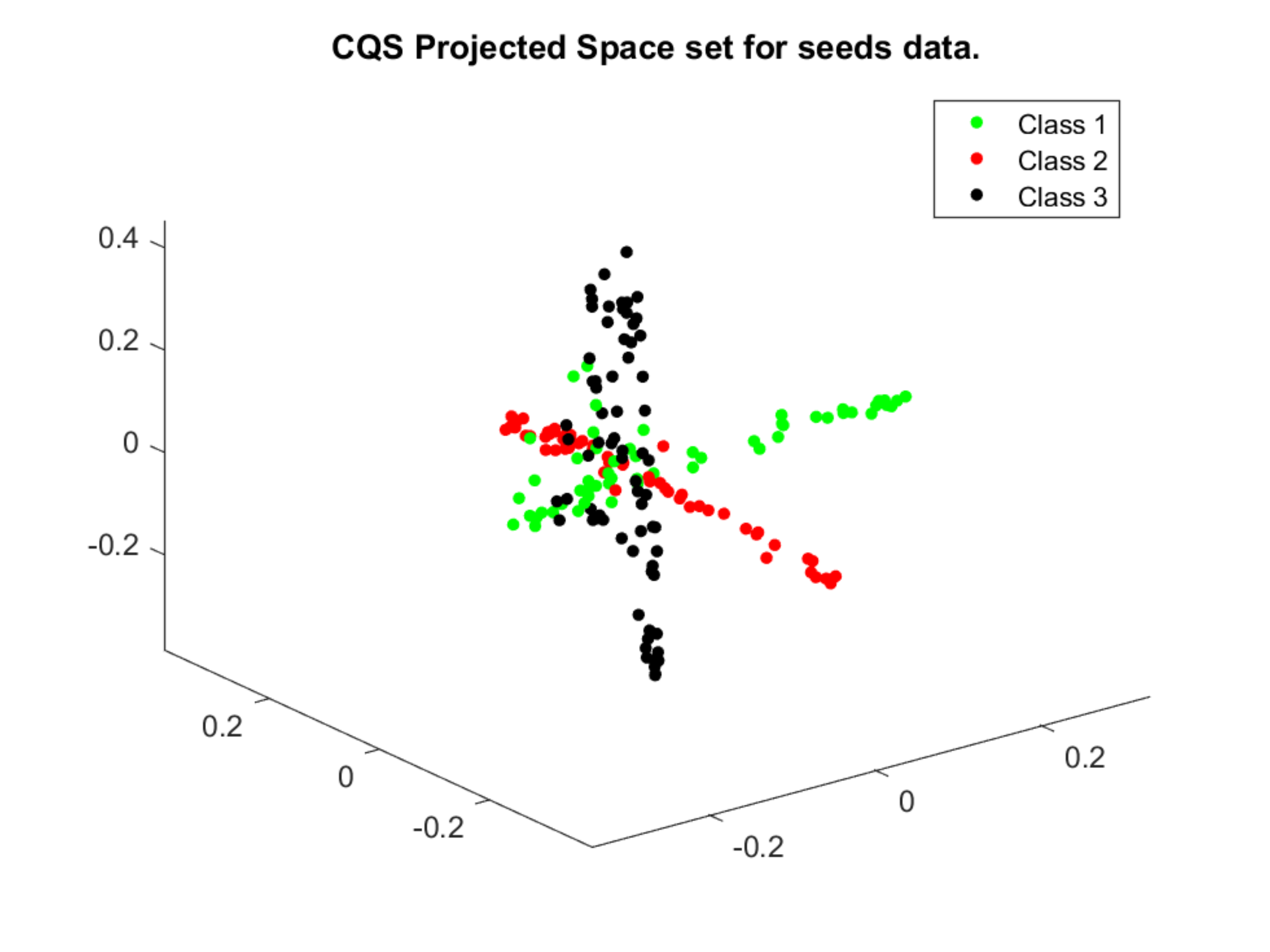} & \includegraphics[scale=0.15]{lg-images/all-iris-pca} & \includegraphics[scale=0.15]{lg-images/all-iris-fld}\tabularnewline
\hline 
\textbf{K-CQS} & \textbf{K-PCA} & \textbf{K-MC-FLD}\tabularnewline
\hline 
\end{tabular}
\par\end{centering}
\caption{Reduced dimensionality projection for a large $\sigma$ value. From
top to bottom: Vertebral, Thyroid, Wine, Iris, Seeds.\label{fig:category-space-reduced-lg-sigma}}
\end{figure}
\par\end{center}

\section{Conclusions}

In this work, we presented a new approach to supervised dimensionality
reduction\textemdash one that attempts to learn orthogonal category
axes during training. The motivation for this work stems from the
observation that the semantics of the multi-class Fisher linear discriminant
are unclear especially w.r.t. defining a space for the categories
(classes). Beginning with this observation, we designed an objective
function comprising sums of quadratic and absolute value functions
(aimed at maximizing the inner product between each training set pattern
and its class axes) with Stiefel manifold constraints (since the category
axes are orthonormal). It turns out that recent work has characterized
such problems and provided sufficient conditions for the detection
of global minima (despite the presence of non-convex constraints).
The availability of a straightforward Stiefel manifold optimization
algorithm tailored to this problem (which has no step size parameters
to estimate) is an attractive by-product of this formulation. The
extension to the kernel setting is entirely straightforward. Since
the kernel dimensionality reduction approach warps the patterns toward
orthogonal category axes, this raises the possibility of using the
angle between each pattern and the category axes as a classification
measure. We conducted experiments in the kernel setting and demonstrated
reasonable performance for the angle-based classifier suggesting a
new avenue for future research. Finally, visualization of dimensionality
reduction for three classes showcases the category space geometry
with clear semantic advantages over principal components and multi-class
Fisher. 

Several opportunities exist for future research. We notice clustering
of patterns near the origin of the category space, clearly calling
for an origin margin (as in SVM's). At the same time, we can also
remove the orthogonality assumption (in the linear case) while continuing
to pursue multi-class discrimination. Finally, extensions to the multi-label
case \cite{sun2013multi} are warranted and suggest interesting opportunities
for future work.

\bibliographystyle{abbrv}
\bibliography{mybibfile_revised}

\end{document}